\documentclass[letterpaper]{article} 
\usepackage{aaai25}  
\usepackage{times}  
\usepackage{helvet}  
\usepackage{courier}  
\usepackage[hyphens]{url}  
\usepackage{graphicx} 
\usepackage{natbib}  
\usepackage{caption} 
\usepackage{algorithm}
\usepackage{algorithmicx}
\usepackage[noend]{algpseudocode}
\usepackage{newfloat}
\usepackage{listings}
\DeclareCaptionStyle{ruled}{labelfont=normalfont,labelsep=colon,strut=off} 
\floatstyle{ruled}
\newfloat{listing}{tb}{lst}{}
\floatname{listing}{Listing}
\usepackage{graphicx}
\usepackage{booktabs}
\title{Exploring Explainable Multi-agent MCTS-minimax Hybrids in Board Game Using Process Mining}
\author {
	Yiyu Qian\textsuperscript{\rm 1}
 	Tim Miller\textsuperscript{\rm 2}
    Liyuan Zhao\textsuperscript{\rm 3}
}
\affiliations {
	\textsuperscript{\rm 1}yiyqian@student.unimelb.edu.au\\
    \textsuperscript{\rm 2}timothy.miller@uq.edu.au\\ 
	\textsuperscript{\rm 3}liyuanz6@uci.edu\\
}

\usepackage{bibentry}

\begin{document}

\maketitle

\begin{abstract}
Monte-Carlo Tree Search (MCTS) is a family of sampling-based search algorithms widely used for online planning in sequential decision-making domains and at the heart of many recent advances in artificial intelligence. Understanding the behavior of MCTS agents is difficult for developers and users due to the frequently large and complex search trees that result from the simulation of many possible futures, their evaluations, and their relationships. This paper presents our ongoing investigation into potential explanations for the decision-making and behavior of MCTS. A weakness of MCTS is that it constructs a highly selective tree and, as a result, can miss crucial moves and fall into tactical traps. Full-width minimax search constitutes the solution. We integrate shallow minimax search into the rollout phase of multi-agent MCTS and use process mining technique to explain agents' strategies in 3v3 checkers.
\end{abstract}

%

\section {Introduction}
Today's artificial intelligence systems have widely spread to applications in our everyday lives. They are powerful enough to outperform humans' efficiency on complex tasks, such as the decision-making process in recruitment organizations. Many companies rely on automated algorithms to reduce the effort of human sources in evaluating candidates' capacity and potential, but it often results in a typical scenario where candidates are rejected on resume selection without any reasons. These algorithms are usually called black-box machine learning models predicting a person's likelihood of being hired~\citep{miller2019explanation}. The black-box means that we only know a series of features will be fed into the model to generate output, but we will not know how the inner mechanism performs this decision-making process. In other words, the implementation of the algorithm is not transparent to users and those affected. Similar concerns arise in other high-stakes domains such as medical image analysis, where users must decide whether to trust a model's output~\citep{wang2024trustmis}, and the behavior of deep models remains difficult to evaluate reliably~\citep{fan2023rethinking}. To address this issue, researchers have started to explore the field of explainable artificial intelligence (XAI) since mid-1980s~\citep{buchanan1984rule}. The purpose of explainable AI is to show details of models' inner workings through providing reasons and evidence on how the output is generated ~\citep{miller2019but}. These models include commonly used machine learning models, such as support vector machine (SVM), deep learning (DL), and reinforcement learning (RL)~\citep{gunning2019xai}. 

In this paper, we explore the explainable RL, specifically focusing on explaining decision-making of MCTS-minimax hybrids in 3v3 checkers game. RL is an area of machine learning that creates optimal policies through agents learning from interactions with the environment ~\citep{sutton2018reinforcement}. Agents are planners who execute actions in problems of decision-making. RL makes decisions for agents in two ways: model-based and model-free. Agents in model-based RL rely on Markov Decision Process (MDP) to learn policies from the environment, and agents in model-free RL can directly learn policies from the environment ~\citep{puterman1990markov}. In terms of understanding behaviors of model-based RL agents and model-free RL agents, Wells points out that there are two major challenges in providing human-readable explanations~\citep{wells2021explainable}. One involves dealing with a large number of sequential decisions generated in a short period of time. Another is the fact that model-free RL agents are trained without training data. It is difficult to provide valid explanations without training data linking actions and observations in the environment. Recent work in model-free RL focuses on learning Intrinsically Interpretable policies ~\citep{milani2022survey}. Topin proposes the argumented MDP method to train agents to directly learn decision tree policies from the environment~\citep{topin2021iterative}. In terms of XRL, previous researchers attempted to generate explanations using natural language and saliency maps. Ehsan and Wang rationalize agents' behaviors by leveraging action-explanation pairs provided by humans~\citep{ehsan2018rationalization} ~\citep{wang2019verbal}. Greydanus constructs saliency maps to measure changes in the policy and highlight regions that are critical to agents' decisions~\citep{greydanus2018visualizing}. 

In prior research conducted by Baier and Winands, MCTS-minimax hybrids are developed by incorporating shallow minimax searches into the MCTS framework~\citep{baier2014mcts}. This is the initial step in combining the strategic strength of MCTS with the tactical strength of minimax. MCTS is a simulation-based search algorithm that builds a tree of possible moves and outcomes iteratively and selects the optimal move based on statistics gathered during the search. It is especially useful in games with uncertain outcomes and intricate decision spaces~\citep{browne2012survey}. Minimax is a deterministic search algorithm that exhaustively explores all possible moves and outcomes, assuming that the opponent will play optimally. It is highly effective in games with perfect information and deterministic outcomes, but may be impractical in games with large decision spaces~\citep{strong2011minimax}. In terms of explainable MCTS, Baier and Kaisers highlight the use of post-hoc models to explain MCTS decisions~\citep{baier2020towards}. They identify three fundamental types of post-hoc questions that occur naturally in numerous explanation contexts: 1. Why do you recommend this action? 2. What do you recommend in these possible futures? 3. Why don't you recommend this alternative action? However, they fail to explain three crucial questions of future research on explainable MCTS: 1. How should feature engineering work be applied on decision data collected from MCTS model? 2. What are recommended post-hoc models to employ? 3. How to specifically apply post-hoc model to process data obtained from MCTS model? These three inquiries are addressed in our research. 

To the best of our knowledge, no one has applied post-hoc interpretability to explain MCTS-minimax hybrids before. The process mining model is utilised as a post-hoc model. Process mining is a technique to discover processes observed in the real world using event logs~\citep{buijs2014quality}. We believe studying policy processes of agents in 3v3 checkers may bring humans ideas about how MCTS-minimax hybrids execute decisions, so we propose an approach to explain agents' behaviors using process mining techniques. We believe our process mining-based post-hoc model will provide explanations on three fundamental types of post-hoc questions. 

\section {Preliminaries} \label{Literature Review}

This section discusses preliminaries in process mining, process models' quality dimensions, MCTS and Minimax. 

\begin{figure}[ht]
	\centering
	\includegraphics[width=1\linewidth]{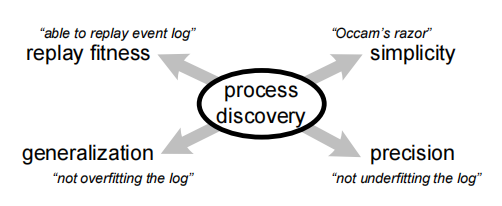}
	\caption{Four quality dimensions in process mining~\cite{buijs2014quality}}
	\label{fig:quality dimensions}
\end{figure}

\subsubsection{Process Mining} \label{Process Mining}
\begin{figure}[ht]
\centering
\includegraphics[width=0.4\linewidth]{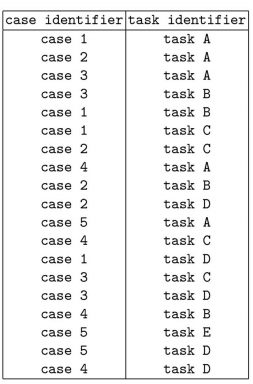}
\caption{Sample event log ~\cite{buijs2014quality}}
\label{fig:sample event log I}
\end{figure}

The purpose of process mining is to generate process models from records of an organization's operational processes ~\cite{buijs2014quality}. Such records, known as event logs, consist of cases and events. Considering the sample event log in figure \ref{fig:sample event log I} from Van’s research in process discovery, we can observe that there are five cases (1, 2, 3, 4, and 5) and four types of tasks (A, B, C, and D) ~\cite{van2004workflow}. For case 1, events ABCD are executed. For case 2, events ACBD are executed. For case 3, events ABCD are executed. For case 4, events ACBD are executed. For case 5, events AED are executed. A process model is used to describe processes' behaviors in the event log ~\cite{buijs2014quality}. Many commonly used process models are developed for process discovery purposes, such as Petri-net. 

Since nowadays black-box machine learning models raise a lot of issues throughout healthcare, business, and other domains, recent researchers start to apply process mining techniques to better understand the inner workings of black-box models. For example, Verenich proposes a “white-box” approach to make predictions on performance indicators of running events ~\citep{verenich2019predicting}. A performance indicator contains sequential cases and events of a running process instance. Information in performance indicators will be treated as event logs and converted to process models using process mining techniques. They then apply workflow analysis to process models and aim to help business analysts have a better understanding of process executions. Gerlach uses process mining to study behaviors of inferred multi-perspective likelihood graph-based event logs generated from the next event predictor’s (NEP) predictions ~\citep{gerlach2022inferring}. They evaluate event logs’ qualities of precision, fitness, and F1 score. Even though their approach can generate inferred likelihood graphs that show a great reflection on behaviors of predictions made by NEP, one prominent limitation is generating events and cases in NEP is quite time-intensive due to exhaustive search in all possible combinations of activities.

\subsubsection{Quality Dimensions} \label{Quality Dimensions}
Quality dimensions in Figure \ref{fig:quality dimensions} are used to measure process models' performance on describing behaviors observed ~\citep{buijs2014quality}. Replay fitness describes the amount of behaviors in an event log that can be demonstrated by a process model. A process model with high replay fitness can replay most cases and events in the event log. Simplicity describes how simple the process model is for humans to understand. This is inferred from Occam's razor: ``one should not increase, beyond what is necessary, the number of entities required to explain anything"~\citep{van2013mediating}. A process model with low simplicity has high complexity thus is it not appropriate to show behaviors in an event log. The precision describes the amount of behaviors showing up in the process model that are not observed in the event log. Van states that it is not enough for a good process model to just hold these qualities. It is possible that a process model replays all behaviors in an event log but shows significant amount of unnecessary behaviors that are not observed in the event log~\citep{van2013mediating}. Such process model may cause the underfitting problem that the model over-generates behaviors not in the event log~\citep{van2012replaying}. To solve this issue, the precision quality is needed to measure the amount of behaviors in a process model that are not observed in the event log. Another issue commonly showing up in a process model is overfitting such that the process model is able to describe behaviors in the current event log but it may not be able to predict behaviors in the next event log under the same system~\citep{van2012replaying}. The generalization describes the likelihood that a process model is able to describe behaviors in an unseen event log generated under the same system. 

In Gerlach's work, replay fitness, precision, and generalization are measured to evaluate the quality of resulting likelihood graph produced from event logs. To compute the replay fitness and precision between event logs generated from NEP (GCs) and ground truth event logs (GTCs), they apply the normalized Levenshtein Distance to compute distances between two cases. Here, L stands for an event log, c stands for a case in the event log L, and d stands for distance between two cases. They calculate F1-score to have a better understanding of model's comprehensive performance on precision and replay fitness. To measure the generalization ability, they first take a set of cases generated from NEP and remove cases that are present in the GCs. Then they look for cases that show up in both remaining cases and cases present in GTCs and divide the size of the set by the number of generated cases~\citep{gerlach2022inferring}. 

In Buijs' work, replay fitness, precision, simplicity and generalization are all measured between two process models, Petri-net and process trees, under each process discovery algorithm~\citep{buijs2014quality}. For example, considering the event log in Figure \ref{fig:sample event log II} and its Petri-net model and process tree model in Figure \ref{fig:Petri-net & Process Tree of sample event log II}, Buiji computes four qualities and results are displayed in Figure \ref{fig:sample event log II's Quality Dimensions}. The perfect score for each quality dimension's indicator is 1.000, and simplicity receives the highest score (s = 1.000) because each event only happens once in both process models. Even though the generalization's score is relatively lower compared to other three scores, it is still considered to be high since it is approximately 0.1 from the limit of perfect score. 




\subsubsection{Minimax} \label{Minimax search}
Minimax search algorithm (Algorithm \ref{Minimax Search algorithm}) is a decision-making strategy used in adversarial search problems, particularly in two-player, zero-sum games like chess, tic-tac-toe, and checkers. The algorithm finds the best course of action by first building a tree of all potential game states and then assessing the final game states. Under the presumption that their opponent is likewise taking the best possible judgements, this route depicts the sequence of actions that will result in the best possible result for the player.

\begin{algorithm}
	\caption{Minimax Search Algorithm}
	\label{Minimax Search algorithm}
	\begin{algorithmic}[1]
		\Function{Minimax}{$node$, $depth$, $isMaximizingPlayer$}
			\If{terminal state or $depth = 0$}
				\State \Return heuristic value of $node$
			\EndIf
			\If{$isMaximizingPlayer$}
				\State $value \gets -\infty$
				\For{each child $c$ of $node$}
					\State $value \gets \max(value, \Call{Minimax}{c, depth - 1, \mathrm{False}})$
				\EndFor
			\Else
				\State $value \gets \infty$
				\For{each child $c$ of $node$}
					\State $value \gets \min(value, \Call{Minimax}{c, depth - 1, \mathrm{True}})$
				\EndFor
			\EndIf
			\State \Return $value$
		\EndFunction
	\end{algorithmic}
\end{algorithm}

\subsubsection{MCTS} \label{MCTS}
Monte Carlo Tree Search (MCTS) rests on two fundamental concepts: that the true value of an action may be approximated using random simulation; and that these values may be used efficiently to adjust the policy towards a best-first strategy. The algorithm progressively builds a partial game tree, guided by the results of previous exploration of that tree. The tree is used to estimate the values of moves, with these estimates (particularly those for the most promising moves) becoming more accurate as the tree is built~\citep{browne2012survey}. 

The basic algorithm involves iteratively building a search tree until some predefined $computational$ $budget$, typically a time, memory or iteration constraint is reached, at which point the search is halted and the bestperforming root action returned. Each node in the search tree represents a state of the domain, and directed links to child nodes represent actions leading to subsequent states. Four steps are applied per search iteration~\citep{chaslot2008monte}. 

\begin{algorithm}
	\caption{General MCTS approach}
	\label{MCTS Search Algorithm}
	\begin{algorithmic}[1]
		\Function{MCTSSEARCH}{$s_0$}
			\State create root node $v_0$ with state $s_0$
			\While{within computational budget}
				\State $v_{l} \gets$ \Call{TREEPOLICY}{$v_0$}
				\State $\Delta \gets$ \Call{DEFAULTPOLICY}{$s(v_{l})$}
				\State \Call{BACKUP}{$v_{l}$, $\Delta$}
			\EndWhile
			\State \Return \Call{BESTCHILD}{$v_0$}
		\EndFunction
	\end{algorithmic}
\end{algorithm}

\begin{enumerate}
	\item Selection: Starting at the root node, a child selection policy is recursively applied to descend through the tree until the most urgent expandable node is reached. A node is expandable if it represents a nonterminal state and has unvisited (i.e. unexpanded) children.
	\item Expansion: One (or more) child nodes are added to expand the tree, according to the available actions.
	\item Simulation: A simulation is run from the new node(s) according to the default policy to produce an outcome.
	\item Backpropagation: The simulation result is ``backed up" (i.e. backpropagated) through the selected nodes to update their statistics.
\end{enumerate}




These steps are summarised in pseudocode in (Algorithm \ref{MCTS Search Algorithm}). Here $v_0$ is the root node corresponding to state $s_0$, $v_l$ is the last node reached during the tree policy stage and corresponds to state $s_l$, and $\Delta$ is the reward for the terminal state reached by running the default policy from state $s_l$. The result of the overall search $a$ (BESTCHILD($v_0$)) is the action $a$ that leads to the best child of the root node $v_0$, where the exact definition of ``best" is defined by the implementation.

\begin{figure}[!h]
	\centering
	\includegraphics[width=1\linewidth]{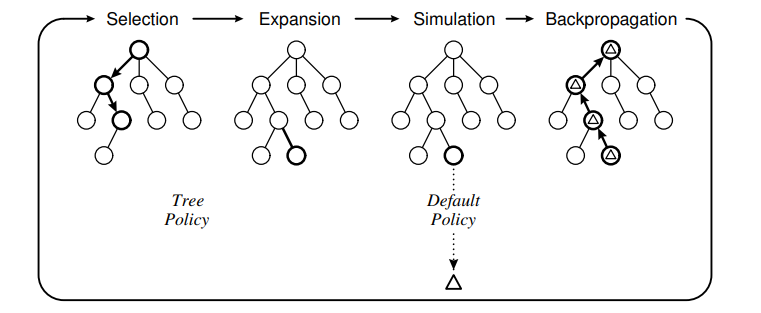}
	\caption{One iteration of the general MCTS approach ~\cite{browne2012survey}}
	\label{fig: One iteration of the general MCTS approach}
\end{figure}

Figure \ref{fig: One iteration of the general MCTS approach} shows one iteration of the basic MCTS algorithm. Starting at the root node $t_0$, child nodes are recursively selected according to some utility function until a node $t_n$ is reached that either describes a terminal state or is not fully expanded (note that this is not necessarily a leaf node of the tree). An unvisited action $a$ from this state $s$ is selected and a new leaf node $t_l$ is added to the tree, which describes the state s' reached from applying action $a$ to state $s$. This completes the tree policy component for this iteration.

A simulation is then run from the newly expanded leaf node $t_l$ to produce a reward value $\Delta$, which is then backpropagated up the sequence of nodes selected for this iteration to update the node statistics; each node's visit count is incremented and its average reward or $Q$ value updated according to $\Delta$. The reward value $\Delta$ may be a discrete (win/draw/loss) result or continuous reward value for simpler domains, or a vector of reward values relative to each agent $p$ for more complex multi-agent domains.



\section {Research Question}

In this paper, we propose a research question: can process mining provide both causal-relationship based explanations and distal explanations for MCTS-minimax hybrids agent's decision-making process? In relation to fundamental post-hoc questions, question 1 (Why do you recommend this action?) and question 3 (Why don't you recommend this alternative action?) belong to causal-relationship based explanations, and question 2 (What do you recommend in these possible futures?) belong to distal explanations. 

To answer this question, we analyze process models that show behaviors of MCTS-minimax hybrids agent. Specifically, in a process model, for each player agent, we identify whether there is a causal relationship between each pair of states and whether agent' next action can be identified based on a series of previous actions and states.  

As far as we know, process mining has never been applied on XRL in previous work. The most recent explainability-focused research in process mining has studied inner workings of the black-box NEP~\cite{gerlach2022inferring}. They show that NEP can generate convincing cases and events through analyzing how events related to each other in the likelihood graph. Our work is quite similar because we treat model-free RL agents' decision making process as a black-box and analyze how states and actions related to each other. 

\section {Methodology}
This section provides an overview our proposed methodology for the research question. 

\subsection{Hypothesis} \label{Hypothesis}

Our hypothesis consists of two parts: 

\begin{enumerate}
	\item The process mining technique can provide players with all decisions considered by the MCTS-minimax hybrids agent in every turn of the 3v3 checkers game.
	\item Based on the process model, both causal-relationship based explanations and distal explanations can be given to interpret agent's actions. 
\end{enumerate}

For causal-relationship based explanations, question 1 (Why do you recommend this action?) and question 3 (Why don't you recommend this alternative action?) can be answered according to all decisions considered by MCTS-minimax hybrids agent from one game turn to multiple future game turns. For distal explanations, question 2 (What do you recommend in these possible futures?) can be answered based on agent' previous decisions shown in process model. 

\subsection{General Approach}
\begin{figure}[!h]
	\centering
	\includegraphics[width=0.75\linewidth]{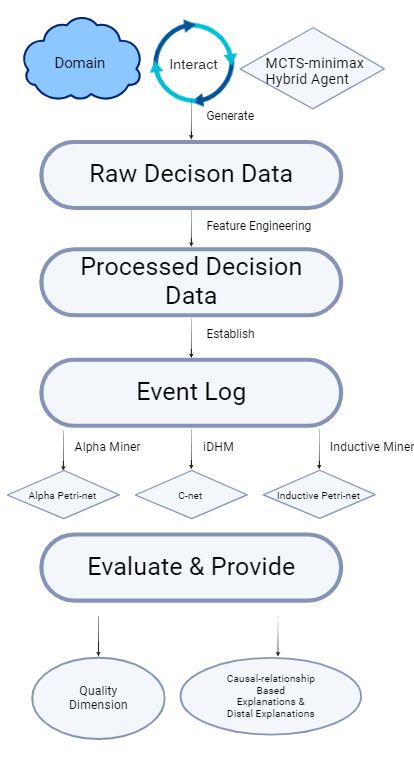}
	\caption{Overview of methodology}
	\label{fig:Overview of methodology}
\end{figure}

Figure \ref{fig:Overview of methodology} shows the general approach of our methodology. We aim to find process models that can help us understand the relationship between decisions made by MCTS-minimax hybrids agent and predict agent's consecutive future decisions. Our proposed general approach consists of four parts:

\begin{enumerate}
	\item Select a domain that has distinct size (state features/number of actions). 
	\item Determine the number of episodes that needs to be executed. For every episode, perform feature engineering work on decisions made by MCTS-minimax hybrids agent and record all decisions. 
	\item Establish an event log based on all episodes executed by the agent. Generate process models based on the event log using various process mining algorithm and use quality dimensions to evaluate model's performance. 
	\item Apply process models to provide causal-relationship based explanations on question 1 (Why do you recommend this action?), question 3 (Why don't you recommend this alternative action?) and distal explanations on question 2 (What do you recommend in these possible futures?). 
\end{enumerate}

In our study, we apply the general approach to discover the environment of a typical board, checkers. Two players sit on opposing sides of the board and compete against each other in the non-cooperative game of checkers (Figure~\ref{fig:Checkers}). One player has dark pieces, while the other has bright ones. Players alternate turns and are not permitted to move an opponent's piece. A move consists of moving a piece diagonally to a neighboring vacant square; if the square directly behind it is empty and the subsequent square contains an opponent's piece, the opponent's piece may be captured (and removed from play) by hopping over it.  Only the dark squares on the board are used for placement of pieces. Only a diagonal entry is permitted for a piece to enter an empty square. When leapt, capturing is necessary. A player is declared the loser when all of their pieces are gone or when they have used up all of their authorised moves. A player's piece is said to be ``crowned" (or ``kinged") if it advances into the kings row on the opposite player's side of the board, becoming a ``king" and obtaining the capacity to move both forward and backward~\citep{samuel1959some}.


\begin{enumerate}
	\item Use an open-sourced checkers domain on Github repository committed by Tim Ruscica (\url{https://github.com/techwithtim/Python-Checkers-AI}). This domain has distinct state features and number of actions. The game board consists of 8 rows and 8 columns, so there are 64 coordinates. State features are represented by coordinates on the game board. Number of actions are represented by each piece's all possible movements. 
	\item Execute 100 episodes of games. In each turn of every episode, the agent makes decisions on selecting which piece to move. Feature engineering work is applied on decisions made by the agent. All decisions are recorded. 
	\item Establish an event log through merging all episodes executed by the agent. Use ({ProM} (\url{https://promtools.org/})), a widely used framework in process mining research, to generate three process models based on the event log using alpha process discovery algorithm, interactive discovery of hybrid models (iDHM) , and inductive miner algorithm. Evaluate each model's performance using conformance checking and performance analysis provided on ({ProM} (\url{https://promtools.org/})).
	\item Apply process models to provide causal-relationship based explanations on question 1 (Why does the MCTS-minimax hybrids agent select this movement?), question 3 (Why not MCTS-minimax hybrids select other movements?) and distal explanations on question 2 (According to the process model, what will be MCTS-minimax hybrids agent's next movement in the future? ). 
\end{enumerate}

\section {Experiment}
This section discusses details of each approach in the methodology and how the experiment is conducted in our research. Our experiment aims to evaluate the replay fitness for each process model generated by alpha discovery algorithm, iDHM, and inductive miner algorithm. We aim to use a process model to provide both causal-relationship based explanations and distal explanations on MCTS-minimax hybrids agent's behaviors. 

\subsection{Domain Setup}

We develop a MCTS-minimax hybrids agent using Python object-oriented programming~\citep{wegner1990concepts}. Specifically, we treat agent as a Python class object ($MCTS$-$minimax$). Steps ($selection$, $expansion$, $simulation$, $backpropagation$) and minimax search algorithm are all treated as a function inside class object.




The game board is a two-dimensional array, as well as an attribute belonging to $Board$ class object. Each element in the array is an instantiated $Piece$ class object. Each $Piece$ has four major attributes: 

\begin{enumerate}
	\item COLOR (data type: tuple)
	\item PIECE ID (data type: integer)
	\item X COORDINATE (data type: integer)
	\item Y COORDINATE (data type: integer)
\end{enumerate}

We use RGB (255, 0, 0) to represent red color and RGB (255, 255, 255) to represent white color. Our study uses a simple reward mechanism: 

\begin{enumerate}
	\item Capture an enemy piece: + 7 points
	\item Become a crown: + 7 points
\end{enumerate}

The original checkers game has 12 white pieces and 12 red white pieces. We reduce the number of pieces from 12 to 3. Our study is conducted under a 3v3 checkers domain. 

\subsection{Minimax and MCTS Integration}

In this study, we perform the strategy of $MCTS$-$MR$ for $MCTS$ with $Minimax$ $Rollouts$ through integrating shallow minimax search into the rollout (simulation) stage of multi-agent MCTS. 

Preliminaries states that the minimax search algorithm works by constructing a game tree (Figure~\ref{fig: Game Tree}), which consists of nodes representing game states and edges representing moves. A game tree is very similar to a binary tree (Figure~\ref{fig: Binary Tree}). 

Details on how minimax search algorithm is integrated into MCTS are specified in Appendix A: Integrate Minimax into MCTS.

Our MCTS-minimax hybrids algorithm model is open-sourced. It can be freely accessed on \url{https://github.com/qyy752457002/Explainable-AI/}.

\subsection{Data Collection and Feature Engineering}

Each episode is recorded as a CSV file. We generate 100 CSV files for red player ($red$ $episode$) and 100 CSV files for white player ($white$ $episode$). In each episode, when the current turn belongs to red player, we establish red agent through instantiating $MCTS$-$minimax$ class object; when the current turn belongs to white player, we establish white agent through instantiating $MCTS$-$minimax$ class object. The agent outputs the most optimal action for both red player and white player. Red player's actions are stored in a $red$ $episode$ and white player's actions are stored in a $white$ $episode$. We generate an event log for red player ($red$ $eventlog$) through merging all 100 $red$ $episodes$ and an event log for white player ($white$ $eventlog$) through merging all 100 $white$ $episodes$. Each episode's id is considered as a case and each action in that episode is considered as an event. 

Listing~\ref{lst: Get moves} shows the function $get$ $moves$ inside our $MCTS$-$minimax$ class object. Listing~\ref{lst: Main part 1} and ~\ref{lst: Main part 2} show our $main$ function used to run the game. The $get$ $moves$ function is used to retrieve a list of movements based on the game state in the current tree node and perform feature engineering work on the movement data retrieved inside function. 

Feature engineering is applied again in the $main$ function to further process the movement data returned by MCTS-minimax hybrids agent.  Feature engineering is the process of transforming raw data into a format that fits places and transitions in process models. It involves selecting, creating, and transforming features in our movement data in order to produce high-quality event logs. The goal of feature engineering is to extract relevant information from the data and present it in a way that is more meaningful and effective for the specific task at hand. 

Our feature engineering work consists of four parts: 

\begin{enumerate}
	\item Convert movements of each piece from exact coordinates on game board (ex. (2, 4) $\rightarrow$ (1, 6)) to abstract representation (ex. (`left', `up')).
	\item Select features to construct MCTS-minimax hybrids agent's movement data tuple. 
	\item Create enemy piece id in the last turn and enemy movement in the last turn. 
	\item Select features to construct event log's transition data tuple. 
\end{enumerate}

Since our study is conducted under 3v3 checkers domain, we label each red piece from 1 to 3 and each white piece from 1 to 3.

\begin{table*}[!h]
	\centering
	\begin{tabular}{|c|c|c|c|c|c|}
		\hline
		\textbf{last\_turn\_id} & \textbf{last\_turn\_movement} & \textbf{piece\_id} & \textbf{move} & \textbf{captured} & \textbf{reward} \\
		\hline
		-1 & () & 2 & (`left', `down') & [] & 0 \\ \hline 
		1 & (`right', `down') & 3 & (`left', `up') & [] & 0 \\ \hline 
		2 & (`right', `down') & 3 & (`left', `down') & [] & 0 \\ \hline 
		1 & (`right', `up') & 3 & (`left', `down') & [] & 0 \\ \hline 
		3 & (`right', `up') & 3 & (`left', `down') & [2] & 14 \\ \hline 
		1 & (`right', `down') & 3 & (`right', `up') & [] & 0 \\ \hline 
		1 & (`right', `up') & 3 & (`left', `down') & [] & 7 \\ \hline 
		3 & (`right', `down') & 3 & (`right', `down') & [] & 0 \\ \hline
		1 & (`right', `down') & 3 & (`left', `up') & [] & 7 \\
		\hline
	\end{tabular}
	\caption{Partial red episode}
	\label{tab:partial_red_episode}
\end{table*}


\begin{table}[!h]
	\centering
	\small
	\begin{tabular}{|c|l|}
		\hline
		\textbf{task\_id} & \textbf{transition} \\
		\hline
		1 & ((-1, `()'), (2, ``(`left', `up')"), 0) \\ \hline 
		1 & ((3, ``(`right', `up')"), (3, ``(`left', `up')"), 0) \\ \hline 
		1 & ((3, ``(`right', `down')"), (2, ``(`left', `down')"), 0) \\ \hline 
		1 & ((3, ``(`right', `down')"), (2, ``(`left', `down')"), 0) \\ \hline 
		1 & ((3, ``(`right', `up')"), (2, ``(`left', `up')"), 0) \\ \hline 
		1 & ((1, ``(`right', `up')"), (3, ``(`left', `up')"), 7) \\ \hline 
		2 & ((-1, `()'), (2, ``(`left', `down')"), 0) \\ \hline 
		2 & ((1, ``(`right', `down')"), (3, ``(`left', `up')"), 0) \\ \hline 
		2 & ((2, ``(`right', `down')"), (3, ``(`left', `down')"), 0) \\ \hline 
		2 & ((1, ``(`right', `up')"), (3, ``(`left', `down')"), 0) \\ \hline 
		2 & ((3, ``(`right', `up')"), (3, ``(`left', `down')"), 14) \\ \hline 
		2 & ((1, ``(`right', `down')"), (3, ``(`right', `up')"), 0) \\ \hline 
		3 & ((-1, `()'), (2, ``(`left', `down')"), 0) \\ \hline 
		3 & ((2, ``(`right', `up')"), (2, ``(`left', `up')"), 0) \\ \hline 
		3 & ((1, ``(`right', `down')"), (3, ``(`left', `up')"), 0) \\ \hline 
		3 & ((2, ``(`right', `down')"), (1, ``(`left', `up')"), 0) \\ \hline 
		3 & ((3, ``(`right', `down')"), (3, ``(`left', `up')"), 0) \\ \hline
		3 & ((2, ``(`right', `up')"), (1, ``(`left', `down')"), 0) \\
		\hline
	\end{tabular}
	\caption{Simplified red event log}
	\label{tab:red_event_log}
\end{table}


\subsection{Trial}
We believe that different iterations, simulation depth, and minimax search depth may affect the quality of process models. We execute three trials:

\begin{enumerate}
	\item $Trial$ $1$: We keep simulation depth 30 and minimax search depth 3. We set iteration times 1000, 2000, and 3000. 
	\item $Trial$ $2$: We keep iteration times 3000 and minimax search depth 3. We set simulation depth 10, 20, and 30. 
	\item $Trial$ $3$: We keep iteration times 3000 and simulation depth 30. We set minimax search depth 1, 2, and 3. 
\end{enumerate}

In each trial, our assumption is that setting fixed parameters maximum results in the most ideal action executed by the MCTS-minimax hybrids agent. For every variable parameter, we perform a test. For example, since $Trial$ $1$ has three variable parameters on 1000, 2000, and 3000, we run three tests parallel on a multi-core operating system using multi-processing technique. We import Python's multiprocessing module and treat every test as a process. For the test in each process, we use the agent to play 100 episodes of games. We do not consider the usage of multi-threading due to the impact of Python's Global Interpreter Lock~\cite{beazley2010understanding}, where a mutex allows only one thread to hold the control of the Python interpreter, which means if we treat a test as a thread, only one test can be in a state of execution at any point in time. Thus, using multi-threading in Python cannot achieve tasks' parallelism.

\subsection{Process Model Evaluation}

In this section, we discuss our methodology of evaluating process models' quality. In each trial, we generate process models for red agent and process models for white agent using alpha process discovery algorithm, iDHM, and inductive miner algorithm. For each process model, we also establish corresponding replay log using conformance analysis plug-in provided in ProM. According to the study conducted by Van Der Aalst et al, the conformance analysis compares a process model to an event log of the same process to show where the real process deviates from the modeled process ~\citep{van2012replaying}. The replay log shows all global statistics for a process model, including model's replay fitness property. A model with good fitness allows for most behavior seen in the event log. A model has a perfect fitness if all traces in the log can be replayed by the model from beginning to end. Unfortunately, ProM does not support conformance analysis on C-net, the process model generated by iDHM. We only discover replay logs for Petri-nets generated by alpha process discovery algorithm and inductive miner algorithm. For every trial, process models are applied to test our hypothesis. For every test in each trial, we are interested in three values among global statistics shown in the replay log of a process model: trace fitness, move-log fitness, and move-model fitness. A process model is considered as a $fitting$ model if all of these three values are perfect  (equal to 1). Such process model can show most of behavior seen in the event log. Otherwise, if  anyone of those values is less than 1, the process model is $non-fitting$. Such process model can show minority of behavior seen in the event log~\citep{ghawi2016process}. 

Appendix C: Evaluate Process Models Based on Trial 1, Trial 2, and Trial 3 states how process models are evaluated in Trial 1, Trial 2, and Trial 3.

\begin{figure}
	\includegraphics[width=1\linewidth]{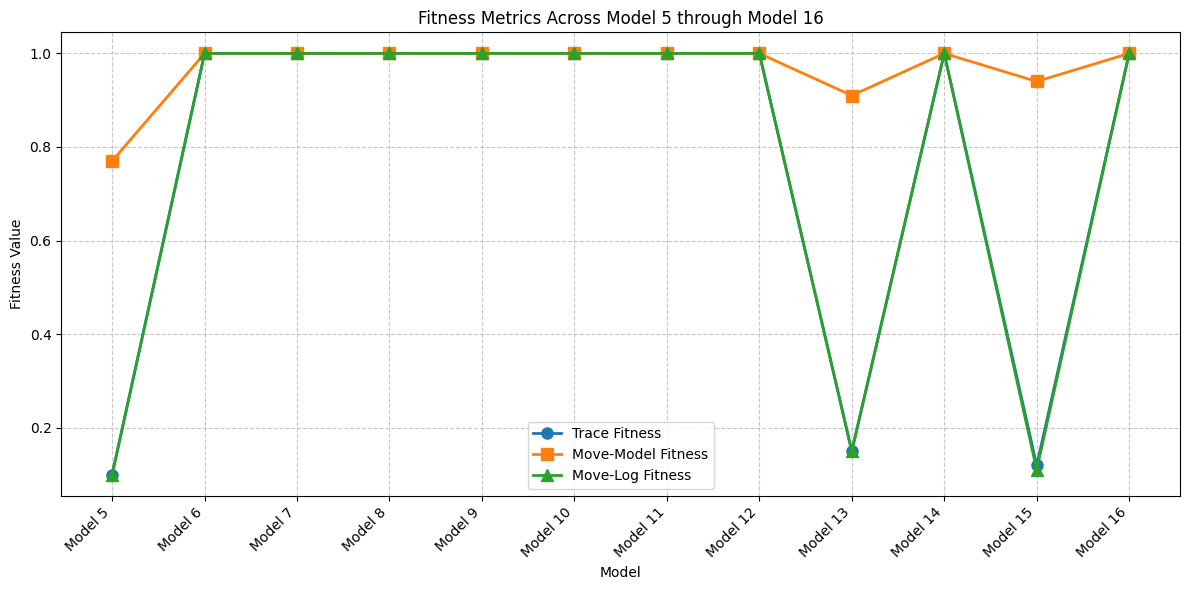}
	\caption{Visual plot shows fitness metrics across Model 5 through Model 16}
	\label{fig:Visual plot shows fitness scores across Tables 5 to 16}
\end{figure}

\begin{table*}[!h]
	\centering
	\begin{tabular}{|l|c|c|c|c|}
		\hline
		Metric & R.A-A.D 1000 & R.A-I.M 1000 & W.A-A.D 1000 & W.A-I.M 1000 \\
		\hline
		Calc. Time (ms) & 15.74 & 3243.23 & 4.33 & 3738.53 \\
		Num. States & 981.20 & 200008.10 & 1.00 & 200019.90 \\
		Trace Fitness & 0.10 & 1.00 & 1.00 & 1.00 \\
		Raw Fitness Cost & 60.00 & 0.00 & 0.00 & 0.00 \\
		Move-Model Fitness & 0.77 & 1.00 & 1.00 & 1.00 \\
		Pre-process time (ms) & 0.40 & 0.34 & 0.33 & 0.59 \\
		Move-Log Fitness & 0.10 & 1.00 & 1.00 & 1.00 \\
		Trace Length & 65.90 & 65.90 & 65.20 & 65.20 \\
		Approx. mem. used (kb) & 105.30 & 12007.10 & 62.60 & 11603.90 \\
		\hline
	\end{tabular}
	
	\vspace{0.5cm}
	
	\begin{tabular}{|l|c|c|c|c|}
		\hline
		Metric & R.A-A.D 2000 & R.A-I.M 2000 & W.A-A.D 2000 & W.A-I.M 2000 \\
		\hline
		Calc. Time (ms) & 4.34 & 2365.69 & 4.50 & 2988.04 \\
		Num. States & 1.00 & 200004.50 & 1.00 & 200006.17 \\
		Trace Fitness & 1.00 & 1.00 & 1.00 & 1.00 \\
		Raw Fitness Cost & 0.00 & 0.00 & 0.00 & 0.00 \\
		Move-Model Fitness & 1.00 & 1.00 & 1.00 & 1.00 \\
		Pre-process time (ms) & 0.51 & 0.46 & 0.30 & 0.37 \\
		Move-Log Fitness & 1.00 & 1.00 & 1.00 & 1.00 \\
		Trace Length & 108.33 & 108.33 & 108.33 & 107.83 \\
		Approx. mem. used (kb) & 68.50 & 11026.00 & 11026.00 & 10127.00 \\
		\hline
	\end{tabular}
	
	\vspace{0.5cm}
	
	\begin{tabular}{|l|c|c|c|c|}
		\hline
		Metric & R.A-A.D 3000 & R.A-I.M 3000 & W.A-A.D 3000 & W.A-I.M 3000 \\
		\hline
		Calc. Time (ms) & 15.12 & 2906.28 & 12.54 & 3750.52 \\
		Num. States & 1189.50 & 200006.70 & 816.90 & 200008.30 \\
		Trace Fitness & 0.15 & 1.00 & 0.12 & 1.00 \\
		Raw Fitness Cost & 72.80 & 0.00 & 75.80 & 0.00 \\
		Move-Model Fitness & 0.91 & 1.00 & 0.94 & 1.00 \\
		Pre-process time (ms) & 0.52 & 0.48 & 0.43 & 0.39 \\
		Move-Log Fitness & 0.15 & 1.00 & 0.11 & 1.00 \\
		Trace Length & 85.30 & 85.30 & 84.90 & 84.90 \\
		Approx. mem. used (kb) & 130.90 & 11703.60 & 107.90 & 12266.50 \\
		\hline
	\end{tabular}
	
	\caption{Global statistics: Petri-net generated by various algorithms (W.A white agent, R.A red agent, I.M inductive miner, A.D alpha discovery) (Fixed Simulation Depth, Fixed Minimax Search Depth, Iteration Times = n)}
	\label{tab:petri-net-stats}
\end{table*}

\subsection {Discovery} \label{discovery}

In order to test hypothesis and give solution to our research questions, we have to ensure the process model used is $fitting$ so that the model can reflect all behaviors seen in the event log. 

Process models generated by inductive miner algorithm where iteration times = 3000 are chosen here for demonstration because both red agent's Petri-net and white agent's Petri-net have perfect fitness. For better illustration, we use each agent's Petri-net based on the simplified event log consisting of 10 events.  

Analysis of discovery on white agent and red agent can be referred to Appendix D: Discovery Analysis.

\section {Limitation}
Our methodology has several limitations that require improvements:

\begin{itemize}
	\item Lack of evaluations on variable quality dimensions: In this study, we evaluate process models' qualities using replay fitness, where we see how many traces in the event log can be replayed by the model from beginning to end. However, there are other three metrics to evaluate a process model: 1. precision: describes the amount of behaviors showing up in the process model that are not observed in the event log; 2. simplicity: describes how simple the process model is for humans to understand; 3. generalization: describes the likelihood that a process model is able to describe behaviors in an unseen event log generated under the same system. Unfortunately, due to limitations of ProM framework, these three metrics are unable to be computed directly. Except for simplicity metric, precision and generalization are unable to be merely perceived through observing structures of places and transitions in process models. 
	\item Difficulty of balancing between simplicity and interpretability: To generate an event log for red player or white player, we perform feature engineering work on the corresponding episodes and merge episodes together. Because the MCTS-minimax hybrids agent plays 100 episodes, the event log has 100 cases. This yields an extremely complex process model with low simplicity, and it takes a huge amount of time for ProM to complete conformance analysis. A process model with low simplicity implies the model is difficult for humans to understand the decision-making process of MCTS-minimax hybrids agent. Even though we have data to support our hypotheses: 1. process mining technique is able to provide players with all decisions considered by the MCTS-minimax hybrids agent in every turn of the 3v3 checkers domain; 2. based on the process model, both causal-relationship based explanations can be given to interpret agent's actions, our findings are only based on a process model that is created by an event log consisting of 10 events. Such process model has high simplicity and it is appropriate for humans to understand structures of places and transitions, but it may not be representative enough to interpret decision-making strategies of MCTS-minimax hybrids agent. In a process model with high simplicity, decisions of MCTS-minimax hybrids agent in every state of the 3v3 checkers domain cannot be fully covered. 
\end{itemize}

More limitations can be referred to Appendix E: Limitation Extended. 

\section {Future Work}

While results indicate that under the 3v3 checkers domain, using process mining can provide players with all possible decisions considered by the MCTS-minimax hybrids agent, as well as both causal-relationship based explanations on question 1 (Why does the MCTS-minimax hybrids agent select this movement?), question 3 (Why not MCTS-minimax hybrids select other movements?) and distal explanations on question 2 (According to the process model, what will be MCTS-minimax hybrids agent's next movement in the future?), further research is needed to understand the impact of this technique under a complex domain. Thus, our future work involves using a more complex checkers domain, such as 6v6 checkers or complete real-world 12v12 checkers to explore the explainability of MCTS-minimax hybrids agent's decision making strategy.

Using a complex checkers domain may result in more possible choices of actions in every game state. Thus, we plan to implement proposed pruning operation stated in Appendix G: MCTS Pruning Operation. 

In addition, future work relevant to feature engineering and human study can be referred to Appendix F: Future Work on Feature Engineering and Human Study.

\begin{table}[!h]
	\centering
	\begin{tabular}{|c|c|}
		\hline
		\textbf{Reward} & \textbf{Actions} \\
		\hline
		10 & A, B, C \\
		\hline
		6 & D, E \\
		\hline
		4 & F, G, H \\
		\hline
		0 & I, J, K, L \\
		\hline
	\end{tabular}
	\caption{Reward hashtable}
	\label{Reward Hashtable}
\end{table}

\section{Conclusion}

Our research shows that compared to process models generated by alpha process discovery algorithm and iDHM, the process model Petri-net generated by inductive miner algorithm can provide us with more insights to analyze MCTS-minimax hybrids agent's decision-making strategy due to the fact that Petri-net generated by inductive miner algorithm always shows a distinct starting place and a distinct ending place. In addition, all transitions always converge to one point in the Petri-net, which meets our expectations that the outcome of one game episode is either red agent wins the game or the white agent wins the game. Current discovery is only based on $Trial$ $1$: $Fixed$ $Simulation$ $Depth$, $Fixed$ $Minimax$ $Search$ $Depth$, $Variable$ $Iteration$ $Times$ of 3v3 checkers domain, and three fundamental types of post-hoc questions proposed in our research can only be explained in this domain. For future work, further exploration needs to be applied on $Trial$ $2$: $Fixed$ $Iteration$ $Times$, $Fixed$ $Minimax$ $Search$ $Depth$, $Variable$ $Simulation$ $Depth$ and $Trial$ $3$: $Fixed$ $Simulation$ $Depth$, $Fixed$ $Iteration$ $Times$, $Variable$ $Minimax$ $Search$ $Depth$, and further research needs to be conducted on different sequential decision-making domains with distinct sizes using MCTS-minimax hybrids. 

\section{Acknowledgments}
We thank any anonymous reviewers for their valuable suggestions. We also thank University of Melbourne Associate Professor Artem Polyvyanyy for supervision in the research. 


\clearpage

\bibliography{aaai25}

\newpage

\appendix
\clearpage

\section{Appendix A: Integrate Minimax into MCTS} \label{Appendix A: Integrate Minimax in MCTS}

According to Kural's study~\cite{kural2005tree}, there are three ways to explore a binary tree using depth-first search algorithm: 

\begin{enumerate}
	\item Preorder traversal: We explore root node first, and then we explore left subtree, and then we explore right subtree.
	\item Inorder traversal: We explore left subtree first, and then we explore root node, and then we explore the right subtree.
	\item Postorder traversal: We explore left subtree first, and then we explore right subtree, and then we explore root node.
\end{enumerate}

For every node in a binary tree, we can also find corresponding spots for preorder traversal, inorder traversal, and post traversal. In our opinion, three questions need to be considered when traversing a binary tree:

\begin{enumerate}
	\item In the spot of preorder traversal, what actions need to be taken before reaching a node? 
	\item In the spot of inorder traversal, what actions need to be taken after completely traversing the left subtree and before starting to traverse the right subtree?
	\item In the spot of postorder traversal, what actions need to be taken before leaving a node? 
\end{enumerate}

Listing~\ref{lst: Preorder}, Listing~\ref{lst: Inorder}, and Listing~\ref{lst: Postorder} show how preorder traversal, inorder traversal, and postorder traversal are implemented in a binary tree both iteratively and recursively. The recursion way to execute these three traversal algorithm is bit easy to understand. The recursion stops if the current node is None (either the root node is None or one level below the leaf node has been reached). For preorder traversal, current node's value is pushed to the $ans$ array at preorder position. For inorder traversal, current node's value is pushed to the $ans$ array at inorder position. For postorder traversal, current node's value is pushed to the $ans$ array at postorder position. The iteration way to execute these three algorithm requires a stack, a last-in-first-out data structure, to store nodes in the binary tree.

Since a game state may come with multiple moves, a node in a game tree may have multiple edges. Thus, a game tree can be a N-ary tree, where the root node may have multiple subtrees. There are only two ways to explore the N-ary tree: 

\begin{enumerate}
	\item Preorder traversal: We explore the root node first, and then we explore every subtree from left to right.
	\item Postorder traversal: We explore every subtree from left to right, and then we explore the root node.
\end{enumerate}

Inorder traversal is unable to be executed in a N-ary tree due to the fact that in a binary tree, each node will only switch the left subtree to the right subtree once, but a node in a N-ary tree may have many children and it needs to switch the subtree several times to traverse. Thus, there does not exist the spot of inorder traversal in a N-ary tree, and only question 1 and question 3 are considered when traversing a N-ary game tree. 

The minimax search algorithm (Listing~\ref{lst: Minimax}) is written as a function inside $MCTS$-$minimax$ class object. The function has four arguments: 

\begin{enumerate}
	\item Self: It points to the instance object of $MCTS$-$minimax$ (Listing~\ref{lst: MCTS_Minimax_agent}) class object.
	\item Node: It is an instantiated $TreeNode$ (Listing~\ref{lst: TreeNode}) class object.
	\item Depth: It is an integer that represents the maximum depth of the game tree.
	\item Max player: It is a boolean value where $True$ stands for maximizing player and $False$ stands for minimizing player.
\end{enumerate}

The minimax search algorithm is implemented in depth-first manner recursively. The algorithm returns two variables: $evaluation$ $score$ and $best$ $movement$, where the $evaluation$ $score$ is calculated based the number of red pieces and white pieces remaining on the two-dimensional game board belonging to $Board$ class object, and $best$ $movement$ is a tuple consisting of instantiated $Board$ class object, $reward$ earned in the current movement, $next$ $turn$ (either red (255, 0, 0) or white (255, 255, 255)), $terminate$ ($True$ means current node is ending game state, and $False$ means current node is not ending game state), and $movement$ $info$. The recursion stops when we reach the maximum depth of the game tree or a winner has come out in the current game board, and then $evaluation$ $score$ and instantiated $Board$ class object in the current tree node are returned. The time complexity of running minimax search algorithm is $O(n)$ where $n$ stands for the number of nodes traversed in a game tree, and each node is visited once. Each recursive call adds the function of minimax search algorithm to the stack memory, and stack memory keeps it until the call is completed. Since the number of recursive call is equivalent to the maximum depth of the game tree $h$, the space complexity of running minimax search algorithm is $O(h)$. 

Considering the case that $max$ $player$ is $True$ and it is our player's turn, in the spot of preorder traversal (what actions need to be taken before reaching a node?), we initialize $evaluation$ $score$ as $-\infty$ and $best$ $movement$ as $None$. We iterate over all possible movements. For each movement, we take out instantiated $Board$ class object, $next$ $turn$, and $terminate$; we create a new node for the next game state through passing these three variables and current tree node to the constructor of $TreeNode$ class object; we recursively call the function of minimax search algorithm to receive the $evaluation$ $score$ for each movement; we find the maximum $evaluation$ $score$ and the $best$ $movement$. In the spot of postorder traversal (what actions need to be taken before leaving a node?), we return maximum $evaluation$ $score$ and the $best$ $movement$ to one level up the game tree. 

Considering the case that $max$ $player$ is $False$ and it is enemy player's turn, in the spot of preorder traversal (what actions need to be taken before reaching a node?), we initialize both $evaluation$ $score$ as $+\infty$ and $best$ $movement$ as $None$. We iterate over all possible movements. For each movement, we take out instantiated $Board$ class object, $next$ $turn$, and $terminate$; we create a new node for the next game state through passing these three variables and current tree node to the constructor of $TreeNode$ class object; we recursively call the function of minimax search algorithm to receive the $evaluation$ $score$ for each movement; we find the minimum $evaluation$ $score$ and the $best$ $movement$. In the spot of postorder traversal (what actions need to be taken before leaving a node?), we return minimum $evaluation$ $score$ and the $best$ $movement$ to one level up the game tree. 

In the simulation stage of traditional MCTS algorithm, the action is chosen randomly. Through embedding minimax search algorithm in the simulation stage of MCTS, the randomness is eliminated and the most optimal action is guaranteed. Listing~\ref{lst: Simulation} shows how the simulation actually functions in our $MCTS$-$minimax$ class object. Since it is 2-player MCTS, the reward is an array of length 2, where the first element stands for $reward$ for white player and the second element stands for $reward$ for red player. Due to the ``Out of memory" (OOM) problem in real-life programs, we set the simulation depth to a specific number. A while loop is utilized to execute the simulation. In every round of simulation, the function of minimax search algorithm is called to reterive the most optimal action. 

\section{Appendix B: White Episode and White Event Log} \label{Appendix B: White Episode and White Event Log}

\begin{table}[!h]
	\centering
	\begin{tabular}{|c|c|c|c|c|c|}
		\hline
		\textbf{last\_turn\_id} & \textbf{last\_turn\_movement} & \textbf{piece\_id} & \textbf{move} & \textbf{captured} & \textbf{reward} \\
		\hline
		2 & (`left', `down') & 1 & (`right', `down') & [] & 0 \\ \hline 
		3 & (`left', `up') & 2 & (`right', `down') & []  & 0 \\ \hline 
		3 & (`left', `down') & 1 & (`right', `up') & [] & 0 \\ \hline 
		3 & (`left', `down') & 3 & (`right', `up') & [] & 0 \\ \hline 
		3 & (`left', `down') & 1 & (`right', `down') & [] & 0 \\ \hline 
		3 & (`right', `up') & 1 & (`right', `up') & [2] & 7 \\ \hline 
		3 & (`left', `down') & 3 & (`right', `down') & [] & 0 \\ \hline 
		3 & (`right', `down') & 1 & (`right', `down') & [] & 0 \\ \hline
		3 & (`left', `up') & 3 & (`right', `down') & [] & 0 \\
		\hline
	\end{tabular}
	\caption{Partial white episode}
	\label{tab:partial_white_episode}
\end{table}

\begin{table}[!h]
	\centering
	\small
	\begin{tabular}{|c|l|}
		\hline
		\textbf{task\_id} & \textbf{transition} \\
		\hline
		1 & ((2, ``(`left', `up')"), (3, ``(`right', `up')"), 0) \\ \hline 
		1 & ((3, ``(`left', `up')"), (3, ``(`right', `down')"), 0) \\ \hline 
		1 & ((2, ``(`left', `down')"), (3, ``(`right', `up')"), 0) \\ \hline 
		1 & ((2, ``(`left', `up')"), (1, ``(`right', `up')"), 0) \\ \hline 
		1 & ((3, ``(`left', `up')"), (1, ``(`right', `down')"), 0) \\ \hline 
		1 & ((1, ``(`left', `up')"), (2, ``(`right', `up')"), 0) \\ \hline 
		2 & ((2, ``(`left', `down')"), (1, ``(`right', `down')"), 0) \\ \hline 
		2 & ((3, ``(`left', `up')"), (2, ``(`right', `down')"), 0) \\ \hline 
		2 & ((3, ``(`left', `down')"), (1, ``(`right', `up')"), 0) \\ \hline 
		2 & ((3, ``(`left', `down')"), (3, ``(`right', `up')"), 0) \\ \hline 
		2 & ((3, ``(`left', `down')"), (1, ``(`right', `down')"), 0) \\ \hline 
		2 & ((3, ``(`right', `up')"), (1, ``(`right', `up')"), 7) \\ \hline 
		3 & ((2, ``(`left', `down')"), (2, ``(`right', `up')"), 0) \\ \hline 
		3 & ((2, ``(`left', `up')"), (1, ``(`right', `down')"), 0) \\ \hline 
		3 & ((3, ``(`left', `up')"), (2, ``(`right', `down')"), 0) \\ \hline 
		3 & ((1, ``(`left', `up')"), (3, ``(`right', `down')"), 0) \\ \hline 
		3 & ((3, ``(`left', `up')"), (2, ``(`right', `up')"), 0) \\ \hline
		3 & ((1, ``(`left', `down')"), (2, ``(`right', `up')"), 0) \\
		\hline
	\end{tabular}
	\caption{Simplified white event log}
	\label{tab:white_event_log}
\end{table}

\clearpage

\section{Appendix C: Evaluate Process Models Based on Trial 1, Trial 2, Trial 3} \label{Appendix C: Evaluate Process Models Based on Trial 1, Trial 2, Trial 3}

\subsubsection {Trial 1: Fixed Simulation Depth, Fixed Minimax Search Depth, Variable Iteration Times} \label{Trial}

\begin{figure}[!h]
	\centering
	\includegraphics[width=\linewidth]{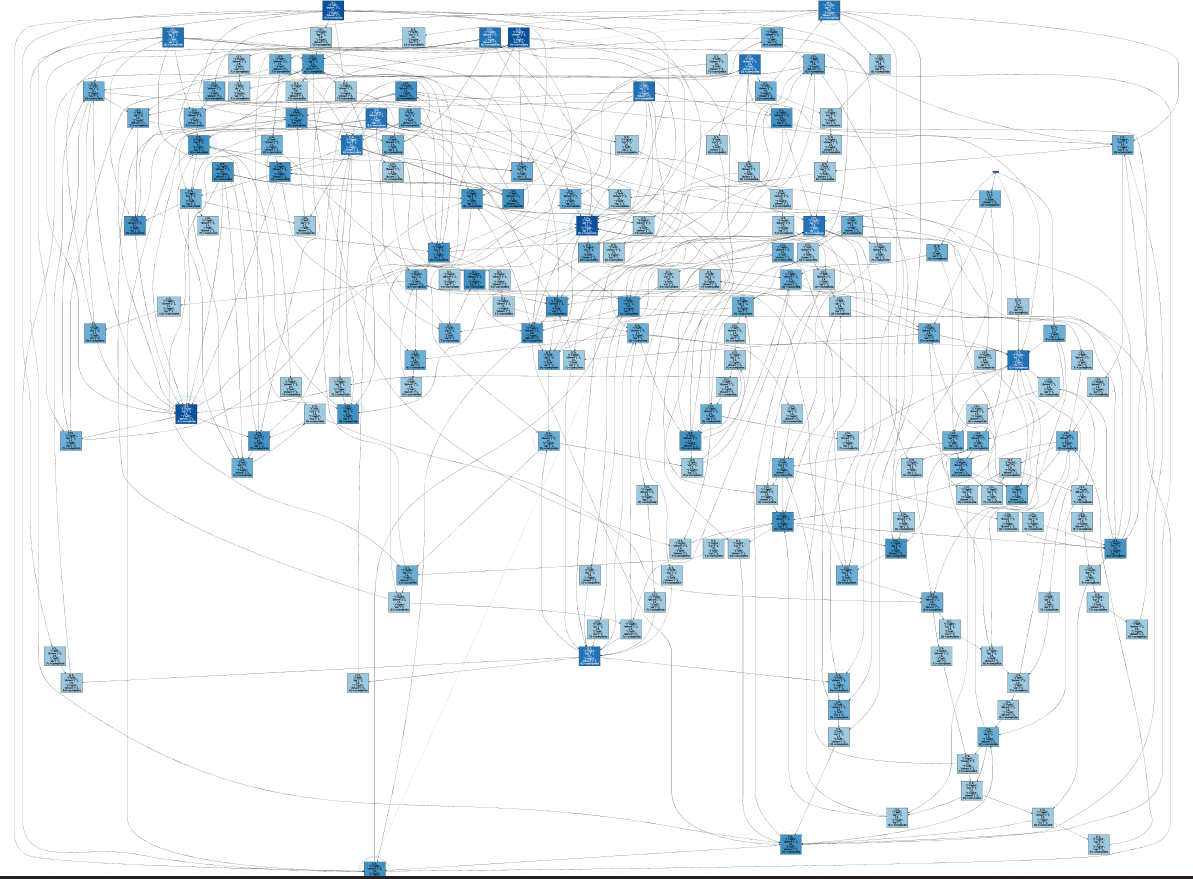}
	\caption{Red agent C-net generated by iDHM (Fixed Simulation Depth, Fixed Minimax Search Depth, Iteration Times = 1000)}
	\label{fig: red_iteration1000_iDHM}
\end{figure}

\begin{figure}[!h]
	\centering
	\includegraphics[width=\linewidth]{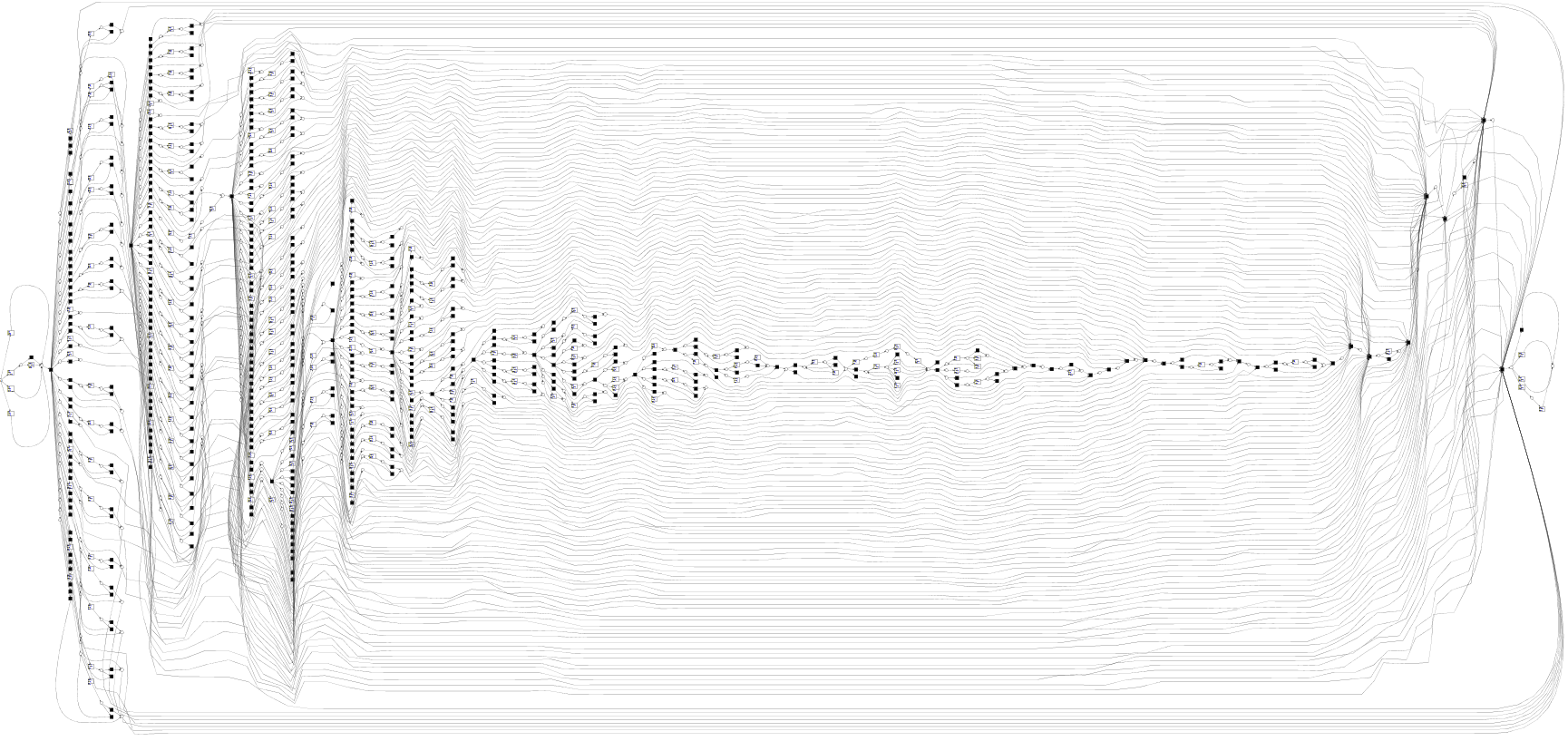}
	\caption{Red agent Petri-net generated by inductive miner algorithm (Fixed Simulation Depth, Fixed Minimax Search Depth, Iteration Times = 1000)}
	\label{fig: red_iteration1000_inductive}
\end{figure}

\begin{figure}[!h]
	\centering
	\includegraphics[width=\linewidth]{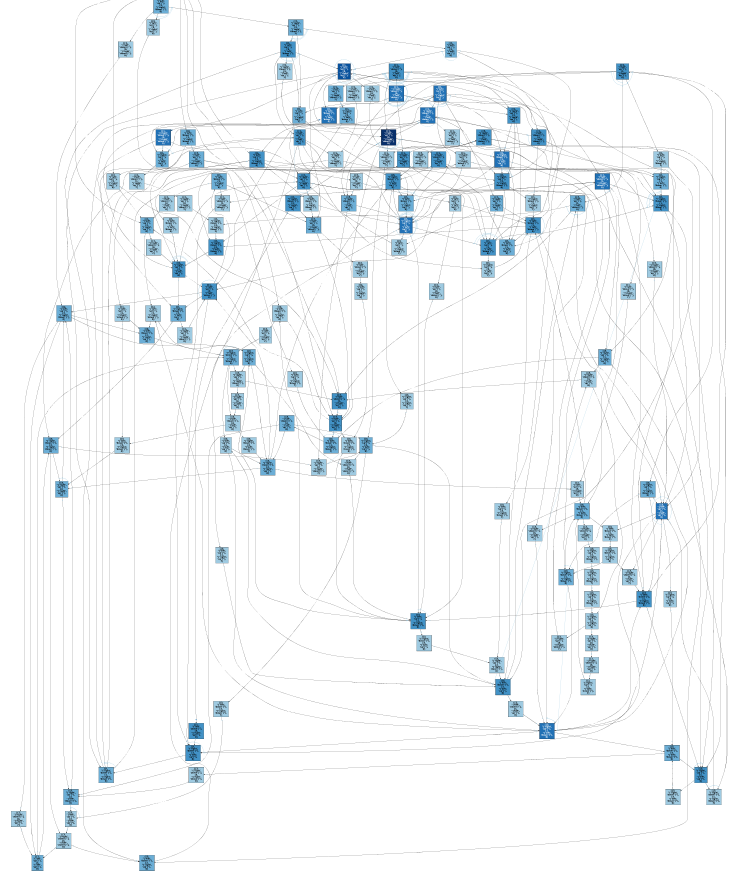}
	\caption{White agent C-net generated by iDHM (Fixed Simulation Depth, Fixed Minimax Search Depth, Iteration Times = 1000)}
	\label{fig: white_iteration_1000_iDHM}
\end{figure}

\begin{figure}[!h]
	\centering
	\includegraphics[width=1\linewidth]{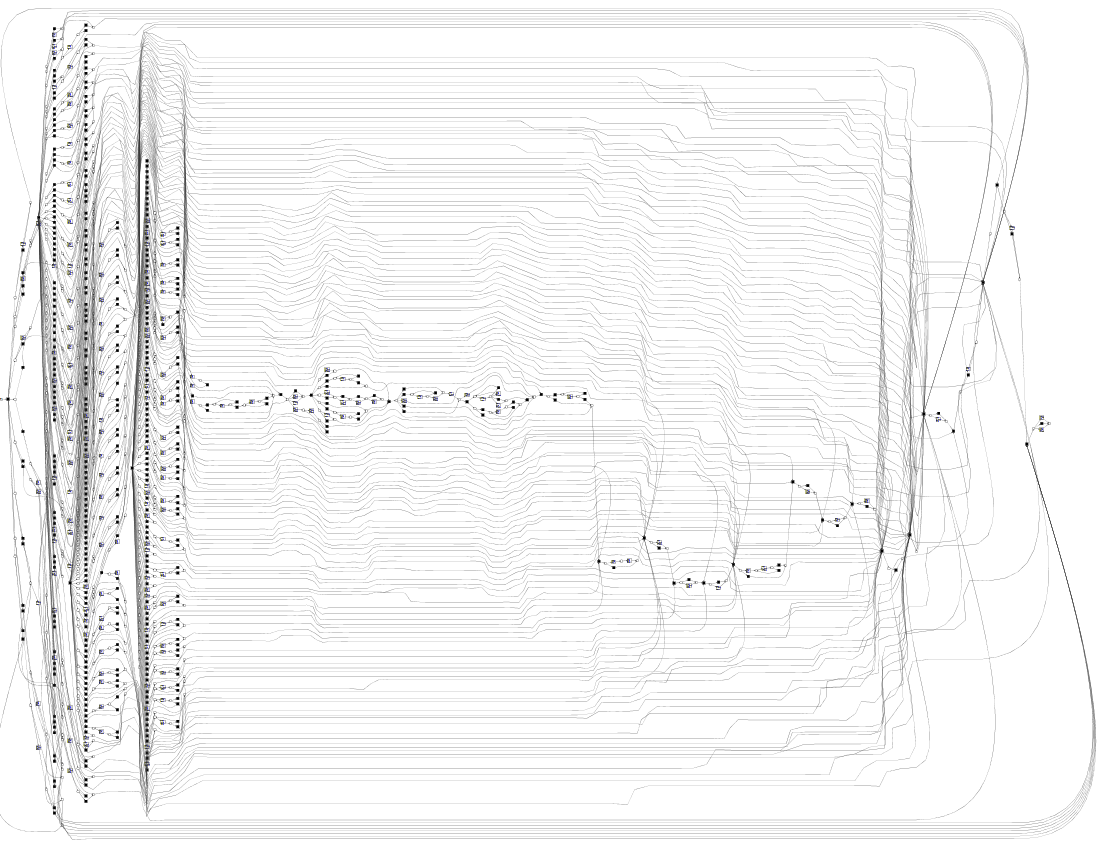}
	\caption{White agent Petri-net generated by inductive miner algorithm (Fixed Simulation Depth, Fixed Minimax Search Depth, Iteration Times = 1000)}
	\label{fig: white_iteration_1000_inductive}
\end{figure}

\begin{figure}[!h]
	\centering
	\includegraphics[width=\linewidth]{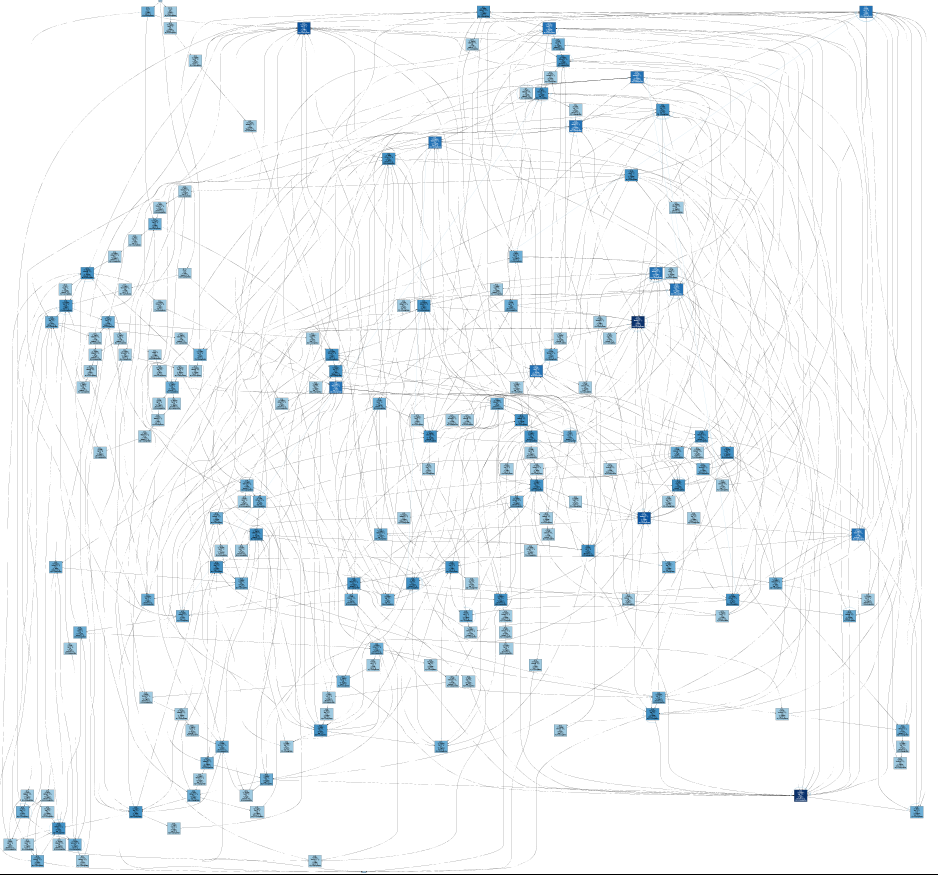}
	\caption{Red agent C-net generated by iDHM (Fixed Simulation Depth, Fixed Minimax Search Depth, Iteration Times = 2000)}
	\label{fig: red_iteration_2000_iDHM}
\end{figure}

\begin{figure}[!h]
	\centering
	\includegraphics[width=\linewidth]{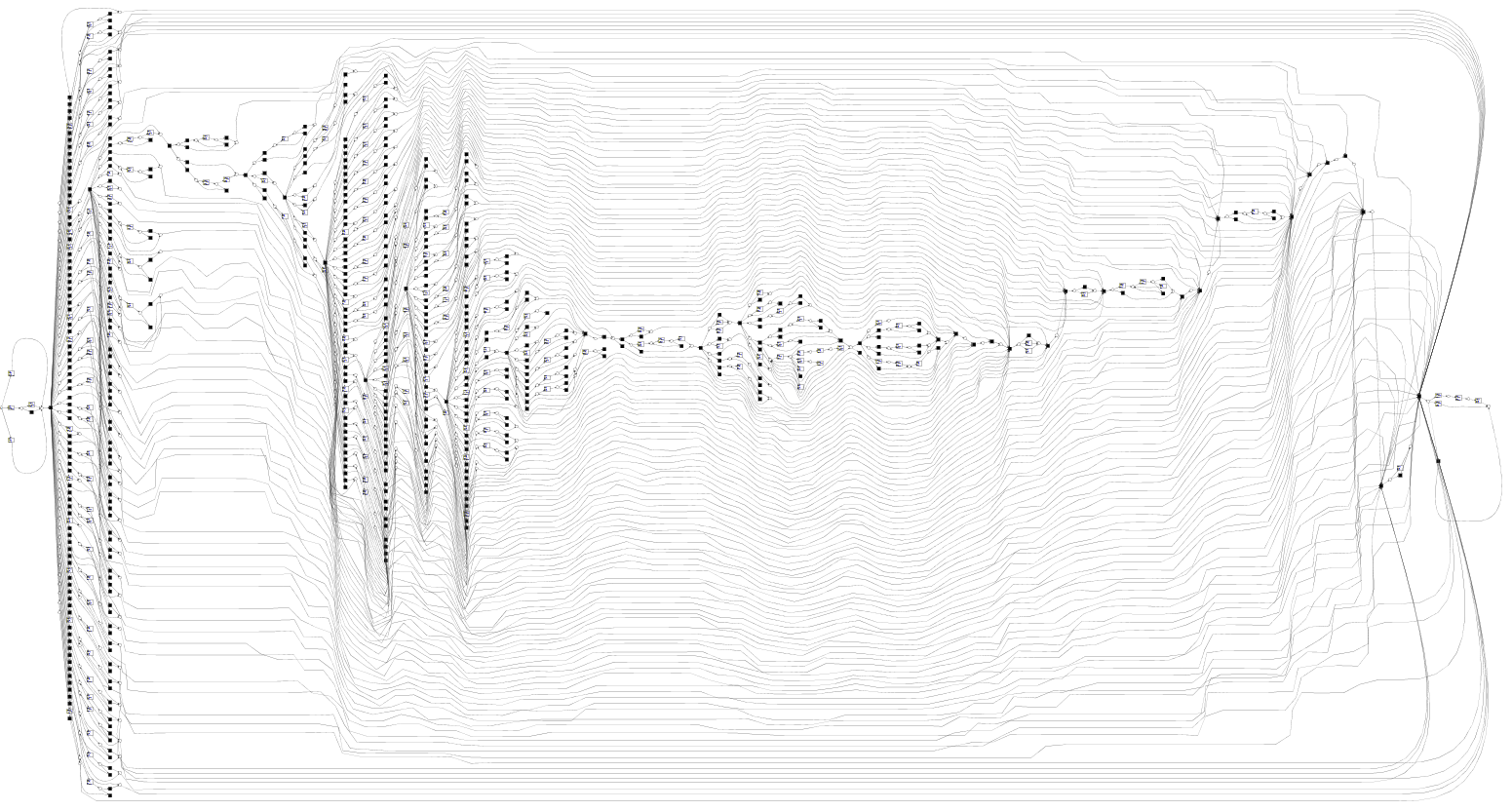}
	\caption{Red agent Petri-net generated by inductive miner algorithm (Fixed Simulation Depth, Fixed Minimax Search Depth, Iteration Times = 2000)}
	\label{fig: red_iteration_2000_inductive}
\end{figure}

\begin{figure}[!h]
	\includegraphics[width=\linewidth]{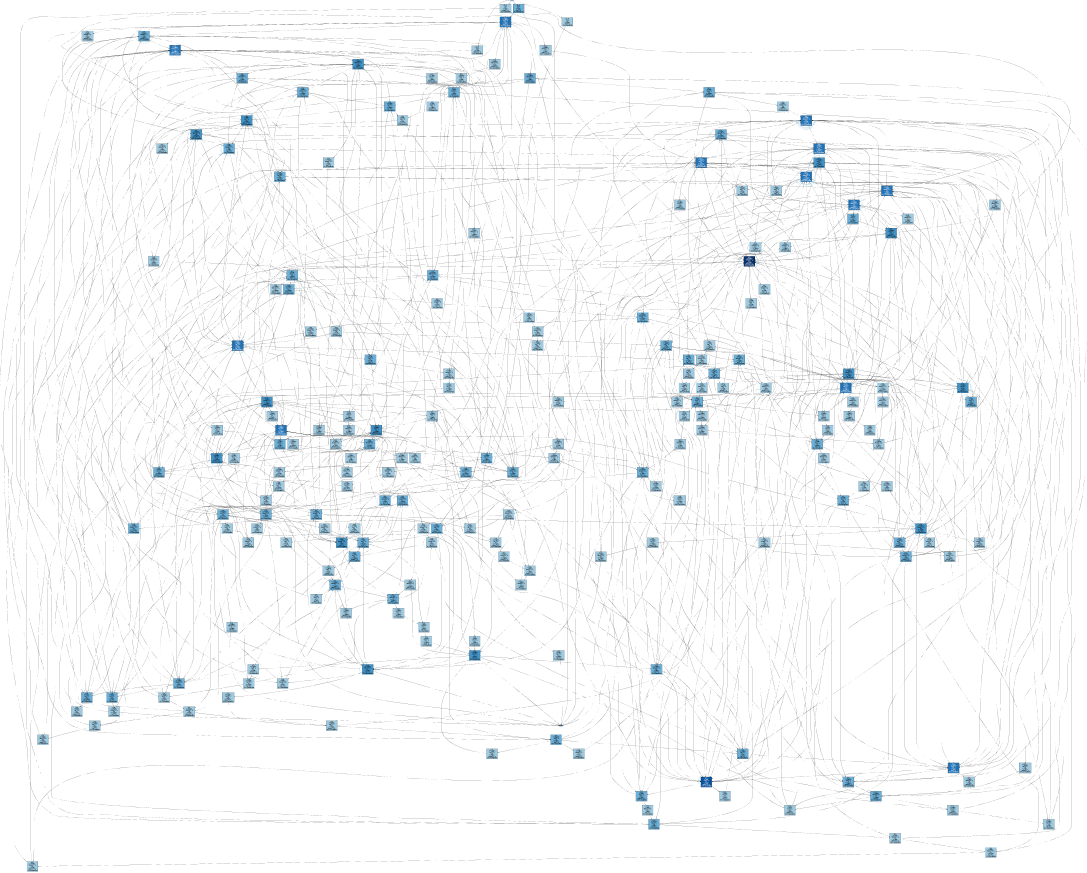}
	\caption{Red agent C-net generated by iDHM (Fixed Simulation Depth, Fixed Minimax Search Depth, Iteration Times = 3000)}
	\label{fig: red_iteration_3000_iDHM}
\end{figure}

\begin{figure}[!h]
	\centering
	\includegraphics[width=\linewidth]{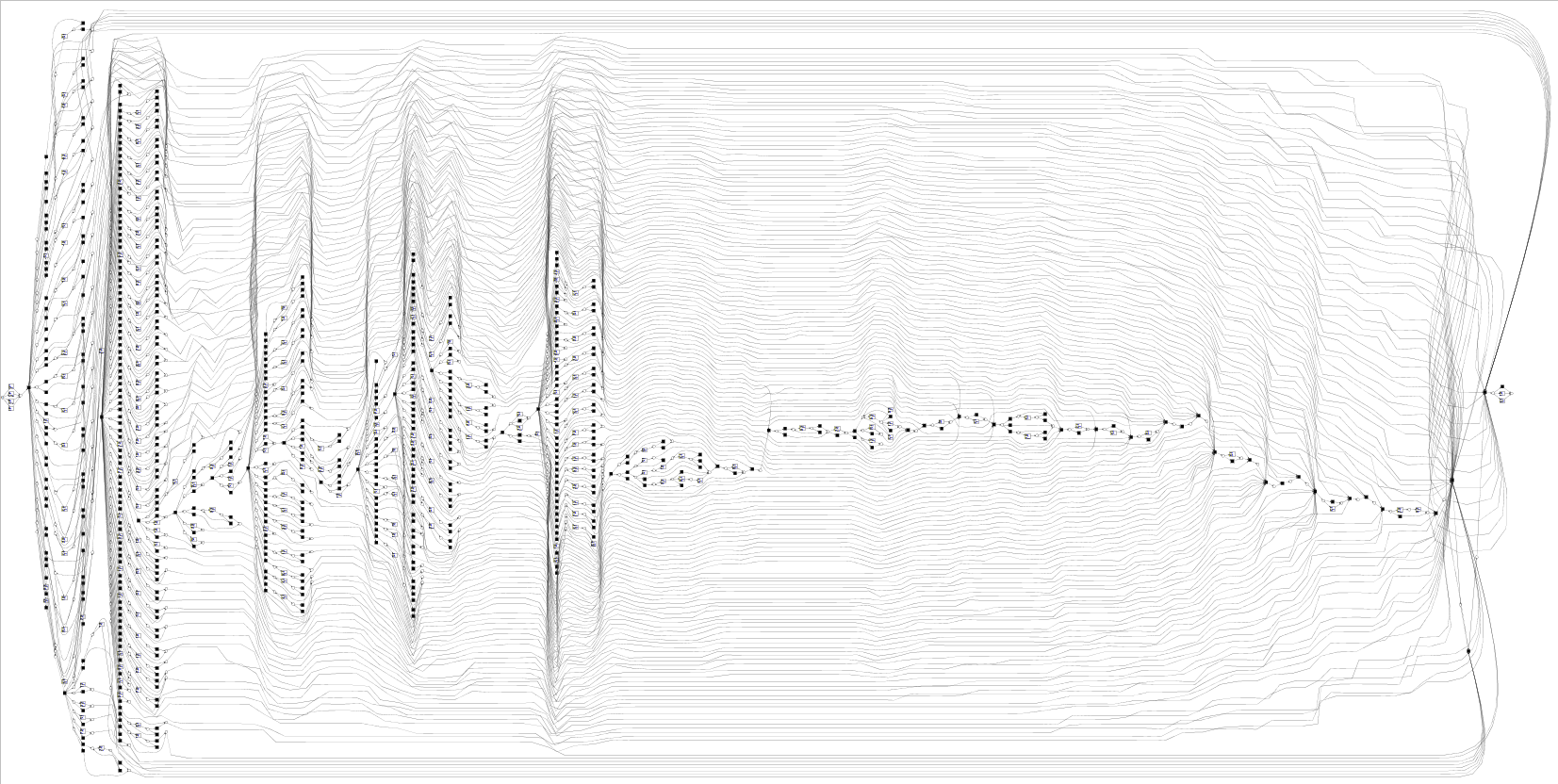}
	\caption{Red agent Petri-net generated by inductive miner algorithm (Fixed Simulation Depth, Fixed Minimax Search Depth, Iteration Times = 3000)}
	\label{fig: red_iteration_3000_inductive}
\end{figure}

\begin{figure}[!h]
	\centering
	\includegraphics[width=\linewidth]{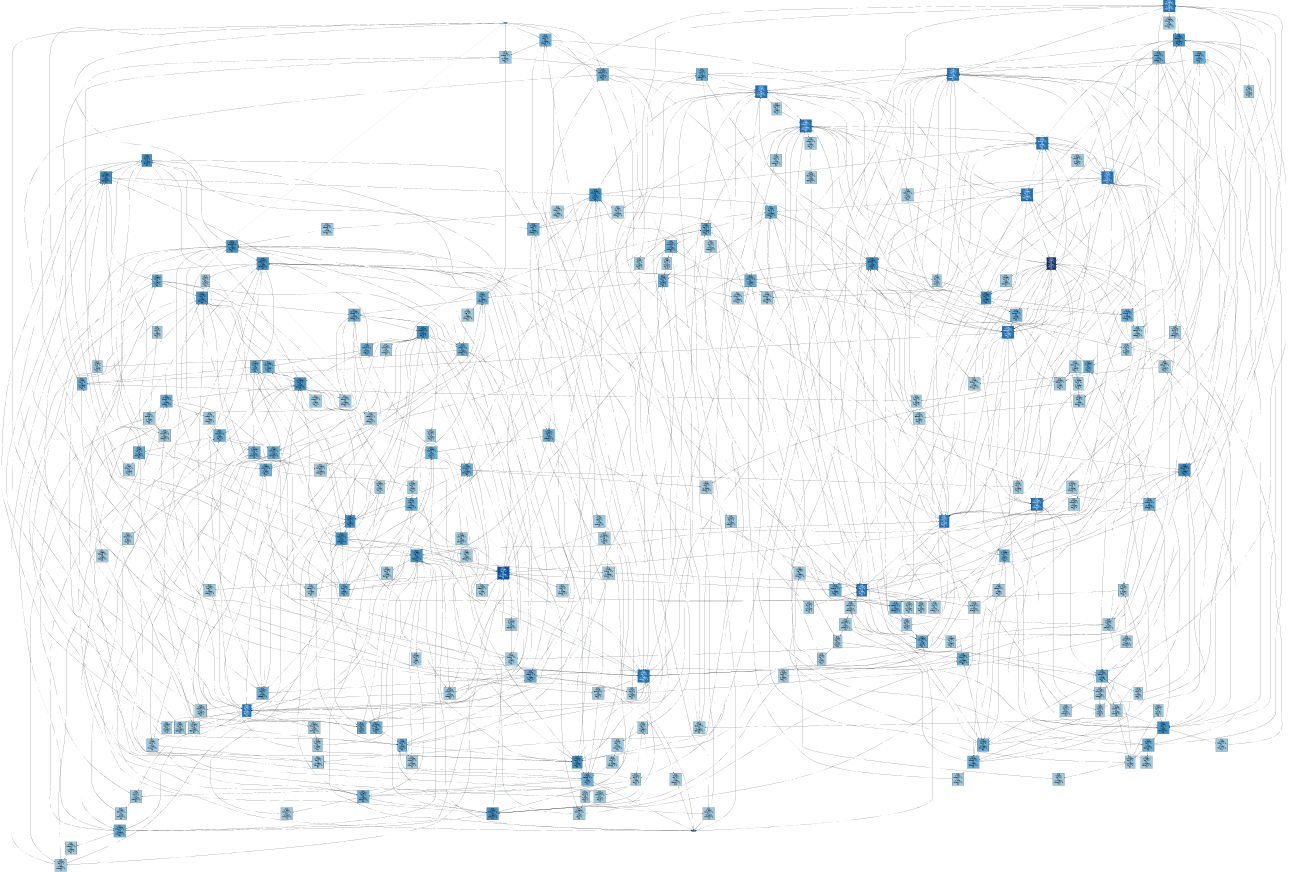}
	\caption{White Agent C-net generated by iDHM (Fixed Simulation Depth, Fixed Minimax Search Depth, Iteration Times = 3000)}
	\label{fig: white_iteration_3000_iDHM}
\end{figure}

\begin{figure}[!h]
	\centering
	\includegraphics[width=\linewidth]{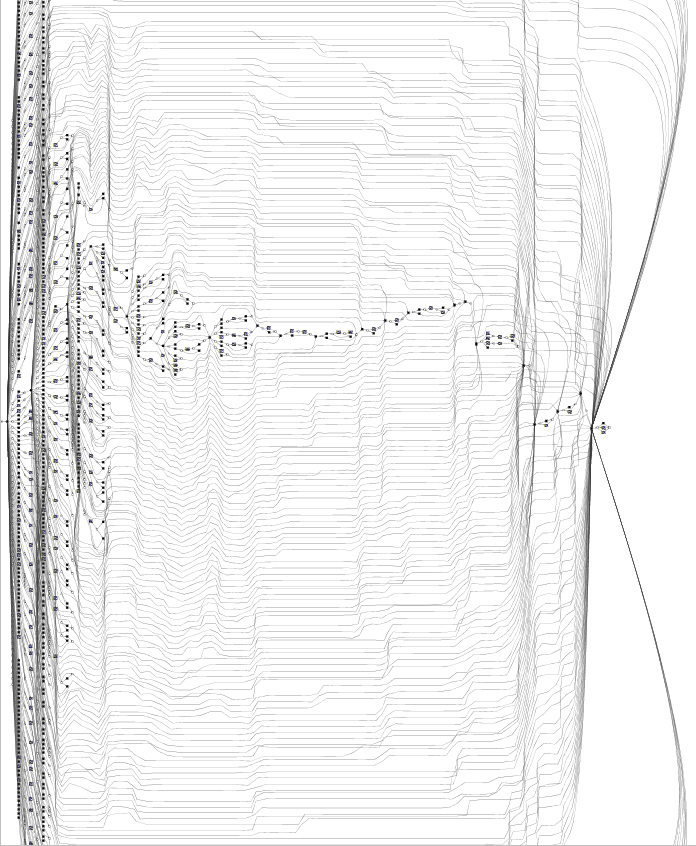}
	\caption{White Agent Petri-net generated by inductive miner algorithm (Fixed Simulation Depth, Fixed Minimax Search Depth, Iteration Times = 3000)}
	\label{fig: white_iteration_3000_inductive}
\end{figure}

\begin{itemize}
	\item Iteration Times = 1000: 
	
	\begin{itemize}
		\item Red agent: Figure~\ref{fig: red_iteration1000_iDHM} shows C-Net generated by iDHM. Figure~\ref{fig: red_iteration1000_inductive} shows Petri-net generated by inductive miner algorithm. 
		
		Model 5 and Model ~6 show the global statistics of red agent's process models generated by alpha discovery algorithm and inductive miner algorithm. 
		
		For the Petri-net generated by alpha discovery algorithm, its trace fitness is 0.10, its move-model fitness is 0.77, and its move-log fitness is 0.10. Thus, it is $non-fitting$ model because all three fitness values are below 1. 
		
		For the Petri-net generated by inductive miner algorithm, its trace fitness, move-model fitness, and move-log fitness are all equal to 1. Thus, it is a $fitting$ model because all three fitness values are perfect. 
		
		\item White agent: Figure~\ref{fig: white_iteration_1000_iDHM} shows C-Net generated by iDHM. Figure~\ref{fig: white_iteration_1000_inductive} shows Petri-net generated by inductive miner algorithm. 
		
		Model 7 and Model 8 show the global statistics of white agent's process models generated by alpha discovery algorithm and inductive miner algorithm. 
		
		For the Petri-net generated by alpha discovery algorithm, its trace fitness, move-model fitness, and move-log fitness are all equal to 1. Thus, it is a $fitting$ model because all three fitness values are perfect. 
		
		For the Petri-net generated by inductive miner algorithm, its trace fitness, move-model fitness, and move-log fitness are all equal to 1. Thus, it is a $fitting$ model because all three fitness values are perfect. 
		
	\end{itemize}
	
	\item Iteration Times = 2000: 
	
	\begin{itemize}
		\item Red agent: Figure~\ref{fig: red_iteration_2000_iDHM} shows C-Net generated by iDHM. Figure~\ref{fig: red_iteration_2000_inductive} shows Petri-net generated by inductive miner algorithm. 
		
		Model 9 and Model 10 show the global statistics of red agent's process models generated by alpha discovery algorithm and inductive miner algorithm.
		
		For the Petri-net generated by alpha discovery algorithm, its trace fitness, move-model fitness, and move-log fitness are all equal to 1. Thus, it is a $fitting$ model because all three fitness values are perfect.
		
		For the Petri-net generated by inductive miner algorithm, its trace fitness, move-model fitness, and move-log fitness are all equal to 1. Thus, it is a $fitting$ model because all three fitness values are perfect.
		
		\item White agent: Figure~\ref{fig: white_iteration_1000_iDHM} shows C-Net generated by iDHM. Figure~\ref{fig: white_iteration_1000_inductive} shows Petri-net generated by inductive miner algorithm. 
		
		Model 11 and Model 12 show the global statistics of white agent's process models generated by the alpha discovery algorithm and inductive miner algorithm.
		
		For the Petri-net generated by alpha discovery algorithm, its trace fitness, move-model fitness, and move-log fitness are all equal to 1. Thus, it is a $fitting$ model because all three fitness values are perfect.
		
		For the Petri-net generated by inductive miner algorithm, its trace fitness, move-model fitness, and move-log fitness are all equal to 1. Thus, it is a $fitting$ model because all three fitness values are perfect.
		
	\end{itemize}
	
	\item Iteration Times = 3000:
	
	\begin{itemize}
		\item Red agent: Figure~\ref{fig: red_iteration_3000_iDHM} shows C-Net generated by iDHM. Figure~\ref{fig: red_iteration_3000_inductive} shows Petri-net generated by inductive miner algorithm. 
		
		Model 13 and Model 14 show the global statistics of red agent's process models generated by alpha discovery algorithm and inductive miner algorithm.
		
		For the Petri-net generated by alpha discovery algorithm, its trace fitness, move-model fitness, and move-log fitness are all equal to 1. Thus, it is a $fitting$ model because all three fitness values are perfect
		
		For the Petri-net generated by inductive miner algorithm, its trace fitness, move-model fitness, and move-log fitness are all equal to 1. Thus, it is a $fitting$ model because all three fitness values are perfect.
		
		\item White agent: Figure~\ref{fig: white_iteration_3000_iDHM} shows C-Net generated by iDHM. Figure~\ref{fig: white_iteration_3000_inductive} shows Petri-net generated by inductive miner algorithm. 
		
		Model 15 and Model 16 show the global statistics of white agent's process models generated by alpha discovery algorithm and inductive miner algorithm.
		
		For the Petri-net generated by alpha discovery algorithm, its trace fitness is 0.12, its move-model fitness is 0.94, and its move-log fitness is 0.11. Thus, it is $non-fitting$ model because all three fitness values are below 1. 
		
		For the Petri-net generated by inductive miner algorithm, its trace fitness, move-model fitness, and move-log fitness are all equal to 1. Thus, it is a $fitting$ model because all three fitness values are perfect.
		
	\end{itemize}
	
\end{itemize}

\begin{table*}[!h]
	\centering
	\begin{tabular}{|l|p{1in}|p{1cm}|p{1cm}|p{1cm}|l|p{14mm}|p{1cm}|} \hline 
		Model&  Calculation Time (ms)&  Trace Fitness&  Move-Model Fitness&  Move-Log Fitness&  Num. States&  Approx. memory used (kb)& Trace Length\\ \hline 
		Model 5&  15.74&  0.10&  0.77&  0.10&  981.20&  105.30& 65.90\\ \hline 
		Model 6&  3243.23&  1.00&  1.00&  1.00&  200008.10&  12007.10& 65.90\\ \hline 
		Model 7&  4.33&  1.00&  1.00&  1.00&  1.00&  62.60& 65.20\\ \hline 
		Model 8&  3738.53&  1.00&  1.00&  1.00&  200010.90&  11603.90& 65.20\\ \hline 
		Model 9&  4.34&  1.00&  1.00&  1.00&  1.00&  68.50& 108.33\\ \hline 
		Model 10&  2365.69&  1.00&  1.00&  1.00&  200004.50&  11026.00& 108.33\\ \hline 
		Model 11&  4.50&  1.00&  1.00&  1.00&  1.00&  11026.00& 108.33\\ \hline 
		Model 12&  2988.04&  1.00&  1.00&  1.00&  200006.17&  10127.00& 107.83\\ \hline 
		Model 13&  15.12&  0.15&  0.91&  0.15&  1189.50&  130.90& 85.30\\ \hline
		Model 14& 2906.28& 1.00& 1.00& 1.00& 200006.70& 11703.60&85.30\\\hline
		Model 15& 12.54& 0.12& 0.94& 0.11& 816.90& 107.90&84.90\\\hline
		Model 16& 3750.52& 1.00& 1.00& 1.00& 200008.30& 12266.50&84.90\\\hline
	\end{tabular}
	\caption{Global statistics: Model 5 to Model 16}
	\label{tab:fitness-scores}
\end{table*}


\textbf{Figure~\ref{fig:Visual plot shows fitness scores across Tables 5 to 16} shows compared fitness metrics (Trace Fitness, Move-Model Fitness, and Move-Log Fitness) across Model 5 through Model 16}.


\textbf{\\ Model exhibiting perfect fitness are Model 6-12, Model 14, and Model 16.}

These models show that process models deliver a perfect fit, as evidenced by the maximum fitness values (1.0) for all three evaluation criteria: Trace Fitness, Move-Model Fitness, and Move-Log Fitness. This suggests that the generated model is capable of capturing the process from the trace perspective, which involves the sequence of activities, and move perspective entailing individual transitions, adequately.

\textbf{\\ Models with non-perfect fitness include Model 5, Model 13, and Model 15.}

{Model 5 shows the lowest fitness values across all metrics (Trace Fitness: 0.10, Move-Model Fitness: 0.77, Move-Log Fitness: 0.10). This suggests significant deviation from the actual process in both activity sequences and individual transitions. Model 13 demonstrates moderate fitness (Trace Fitness: 0.15, Move-Model Fitness: 0.91, Move-Log Fitness: 0.15). The higher Move-Model Fitness indicates that individual transitions are well-represented, but the lower Trace Fitness and Move-Log Fitness suggest issues with overall sequence representation. Model 15 exhibits high Move-Model Fitness (0.94) but lower Trace Fitness (0.12) and Move-Log Fitness (0.11). This suggests that while individual transitions are well-modeled, the overall sequence of activities and log replay don't align as closely with the actual process.}

\subsubsection {Trial 2: Fixed Iteration Times, Fixed Minimax Search Depth, Variable Simulation Depth} \label{Trial 2}

\begin{figure}[!h]
	\centering
	\includegraphics[width=\linewidth]{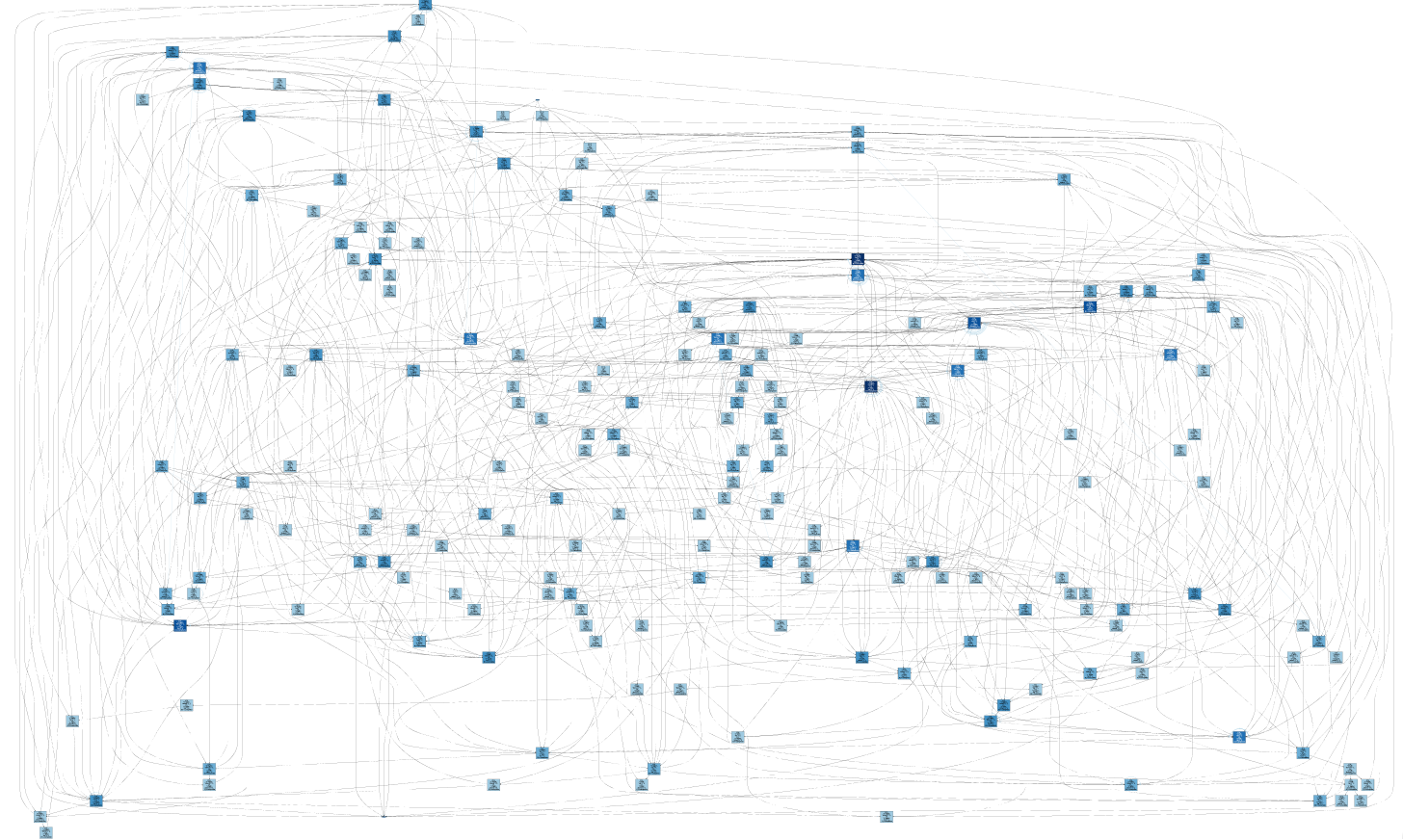}
	\caption{Red Agent C-net generated by iDHM (Fixed Iteration Times, Fixed Minimax Search Depth, Simulation Depth = 10)}
	\label{fig: red_simulation_10_iDHM}
\end{figure}

\begin{figure}[!h]
	\centering
	\includegraphics[width=\linewidth]{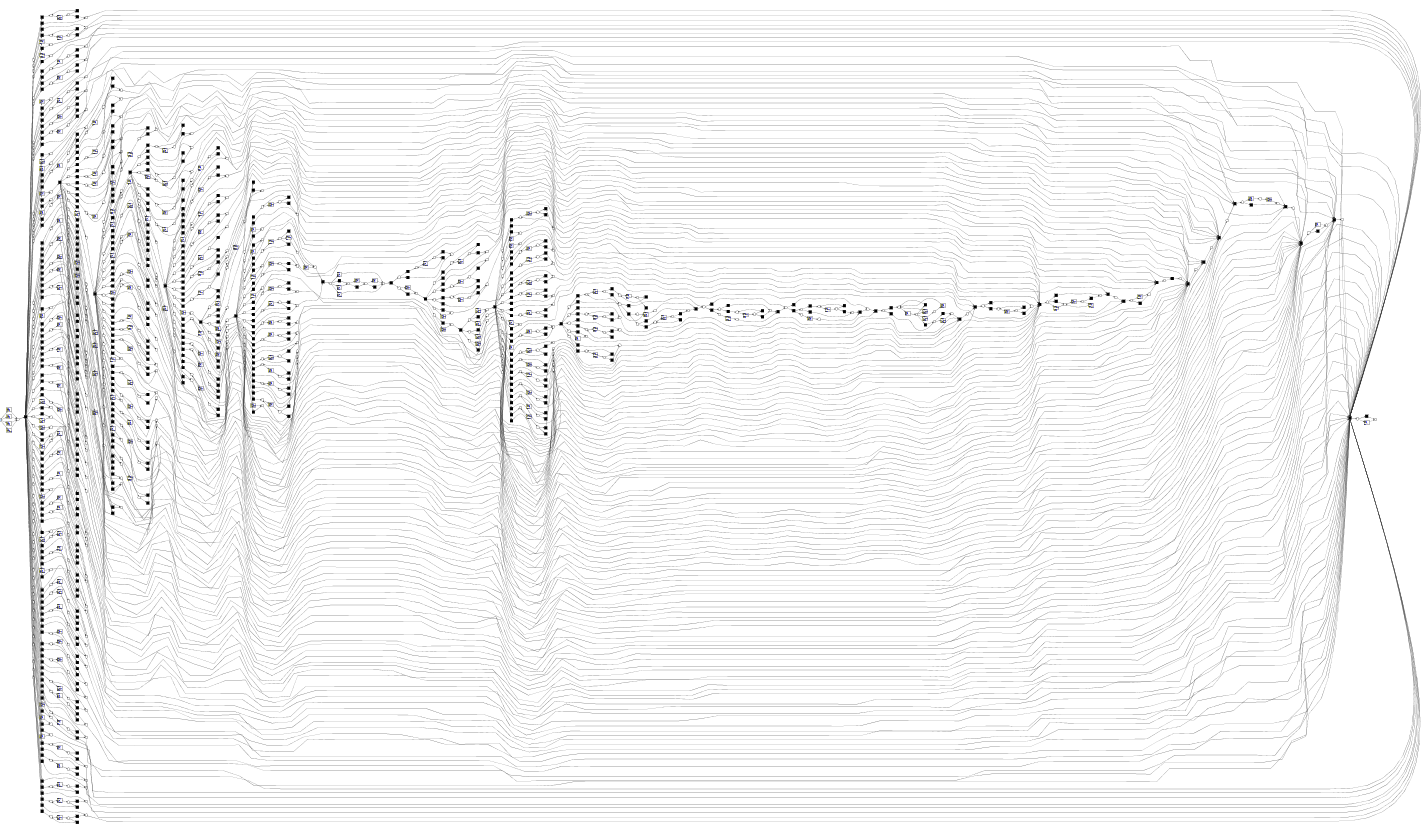}
	\caption{Red Agent Petri-net generated by inductive miner algorithm (Fixed Iteration Times, Fixed Minimax Search Depth, Simulation Depth = 10)}
	\label{fig: red_simulation_10_inductive}
\end{figure}

\begin{figure}[!h]
	\centering
	\includegraphics[width=\linewidth]{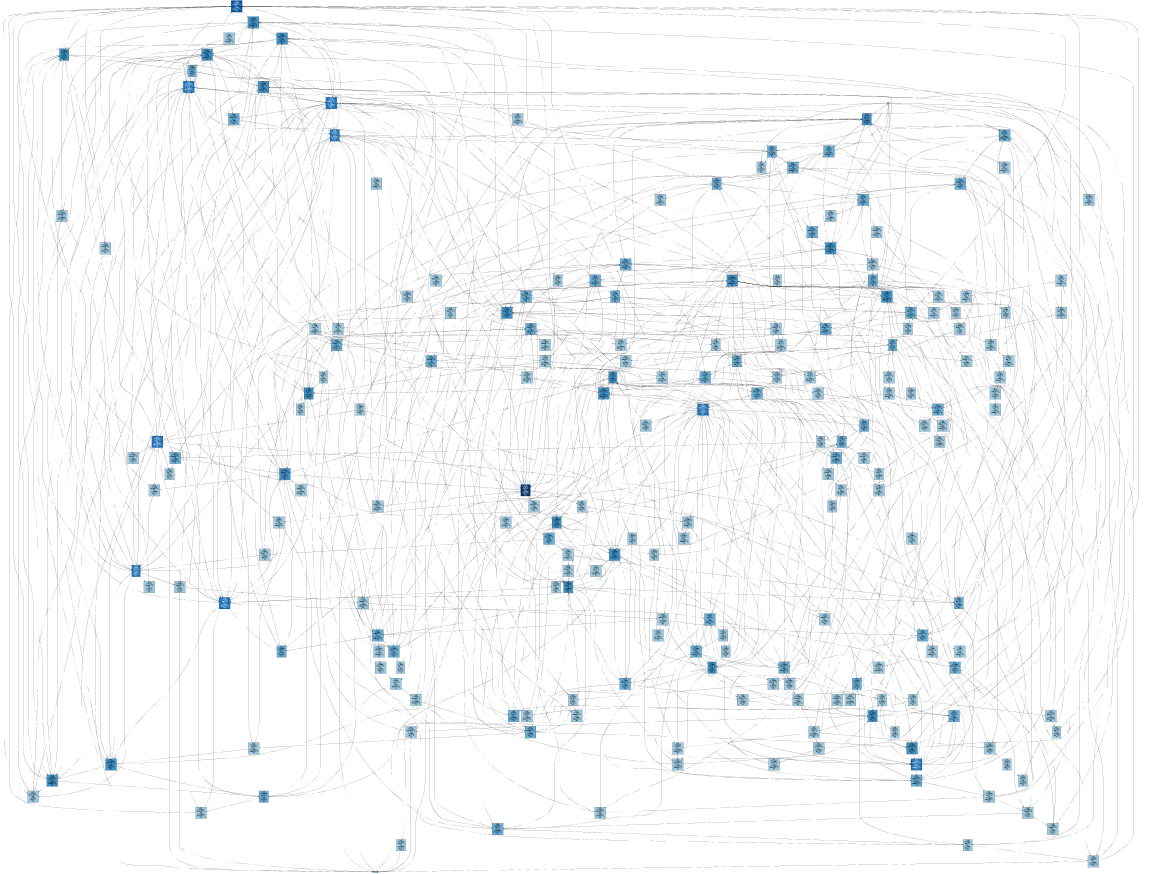}
	\caption{White Agent C-net generated by iDHM (Fixed Iteration Times, Fixed Minimax Search Depth, Simulation Depth = 10)}
	\label{fig: white_simulation_10_iDHM}
\end{figure}

\begin{figure}[!h]
	\centering
	\includegraphics[width=\linewidth]{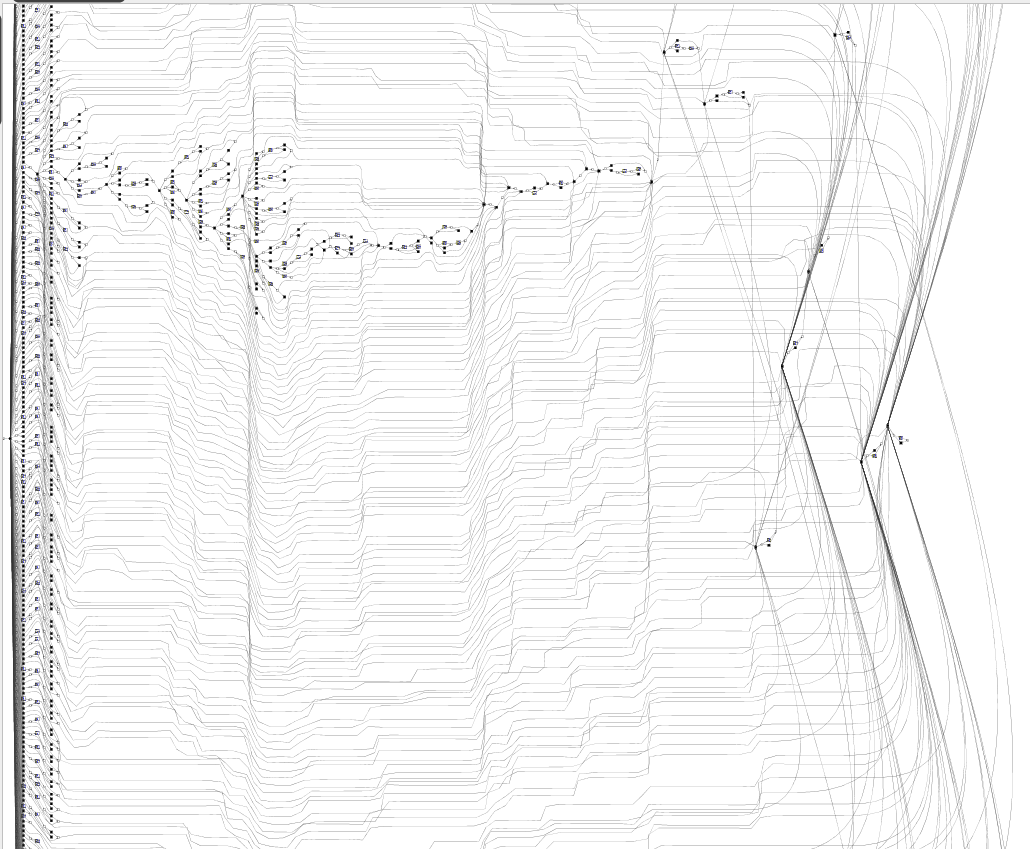}
	\caption{White Agent Petri-net generated by inductive miner algorithm (Fixed Iteration Times, Fixed Minimax Search Depth, Simulation Depth = 10)}
	\label{fig: white_simulation_10_inductive}
\end{figure}

\begin{figure}[!h]
	\centering
	\includegraphics[width=\linewidth]{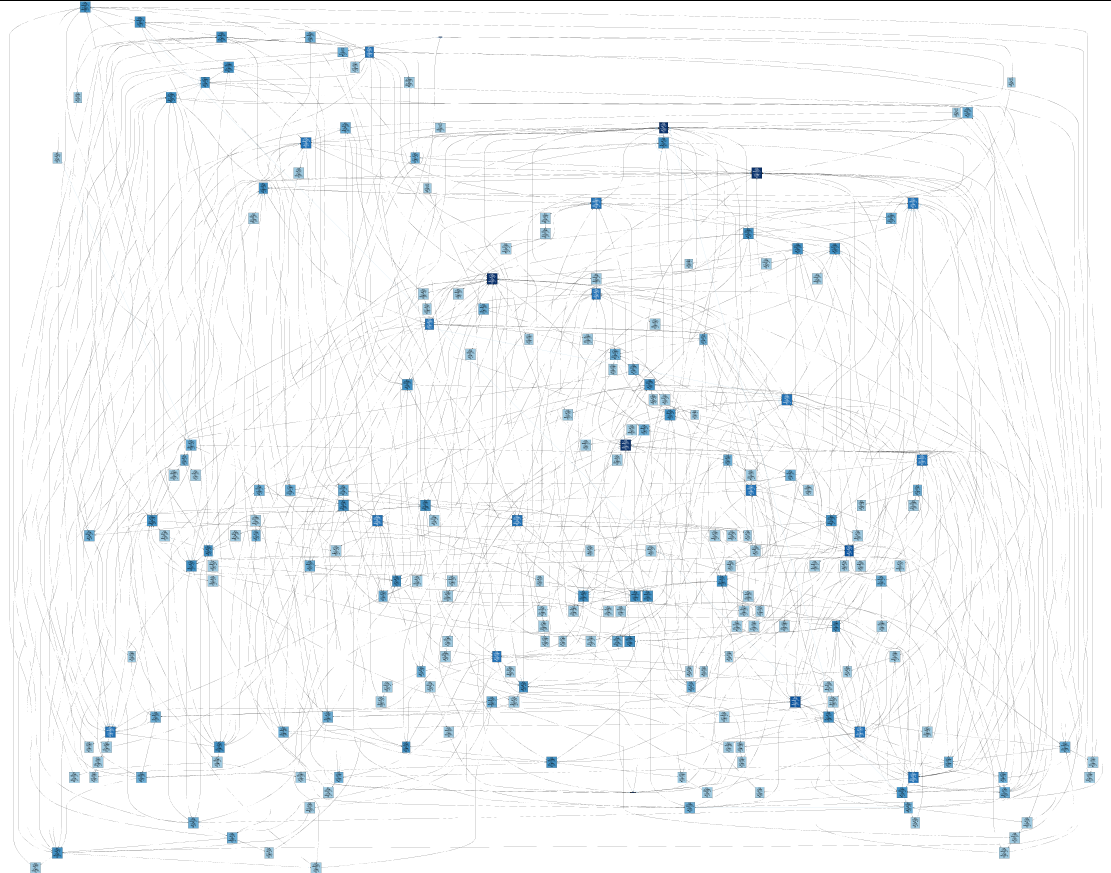}
	\caption{Red Agent C-net generated by iDHM (Fixed Iteration Times, Fixed Minimax Search Depth, Simulation Depth = 20)}
	\label{fig: red_simulation_20_iDHM}
\end{figure}

\begin{figure}[!h]
	\centering
	\includegraphics[width=\linewidth]{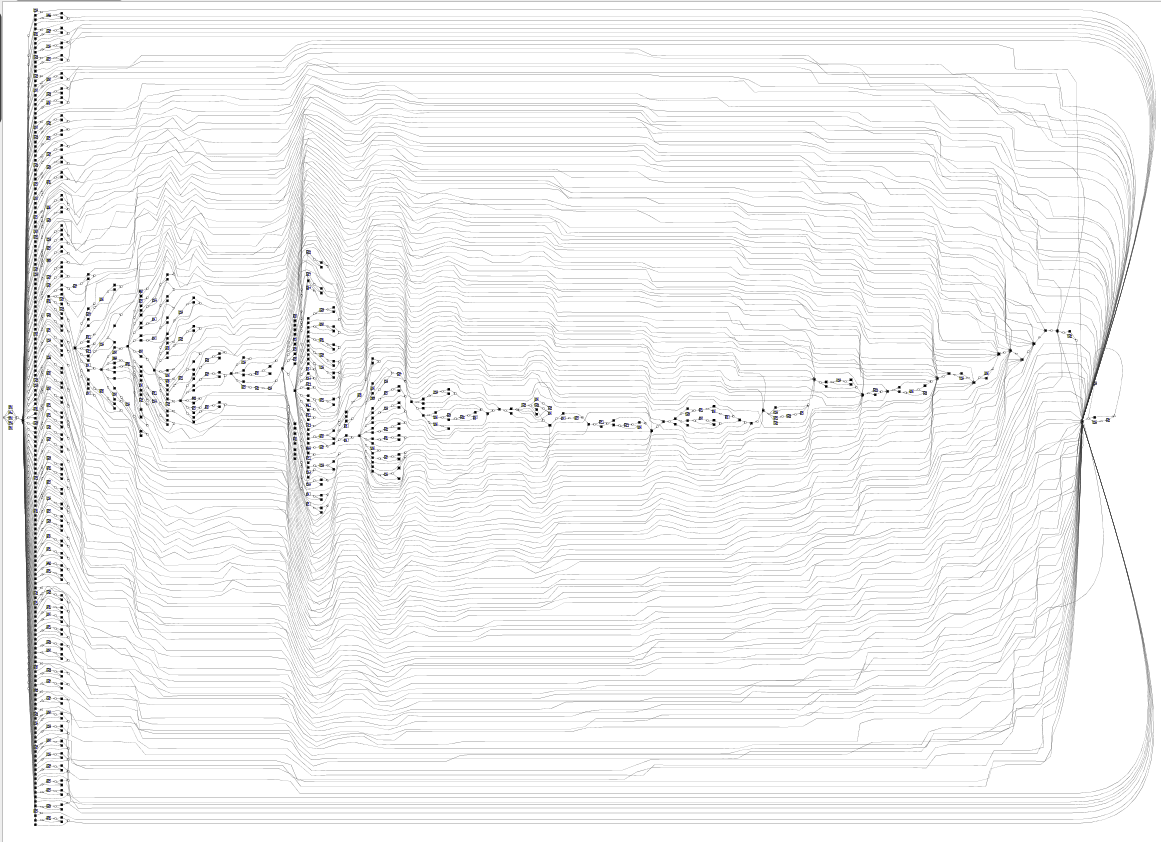}
	\caption{Red Agent Petri-net generated by inductive miner algorithm (Fixed Iteration Times, Fixed Minimax Search Depth, Simulation Depth = 20)}
	\label{fig: red_simulation_20_inductive}
\end{figure}

\begin{figure}[!h]
	\centering
	\includegraphics[width=\linewidth]{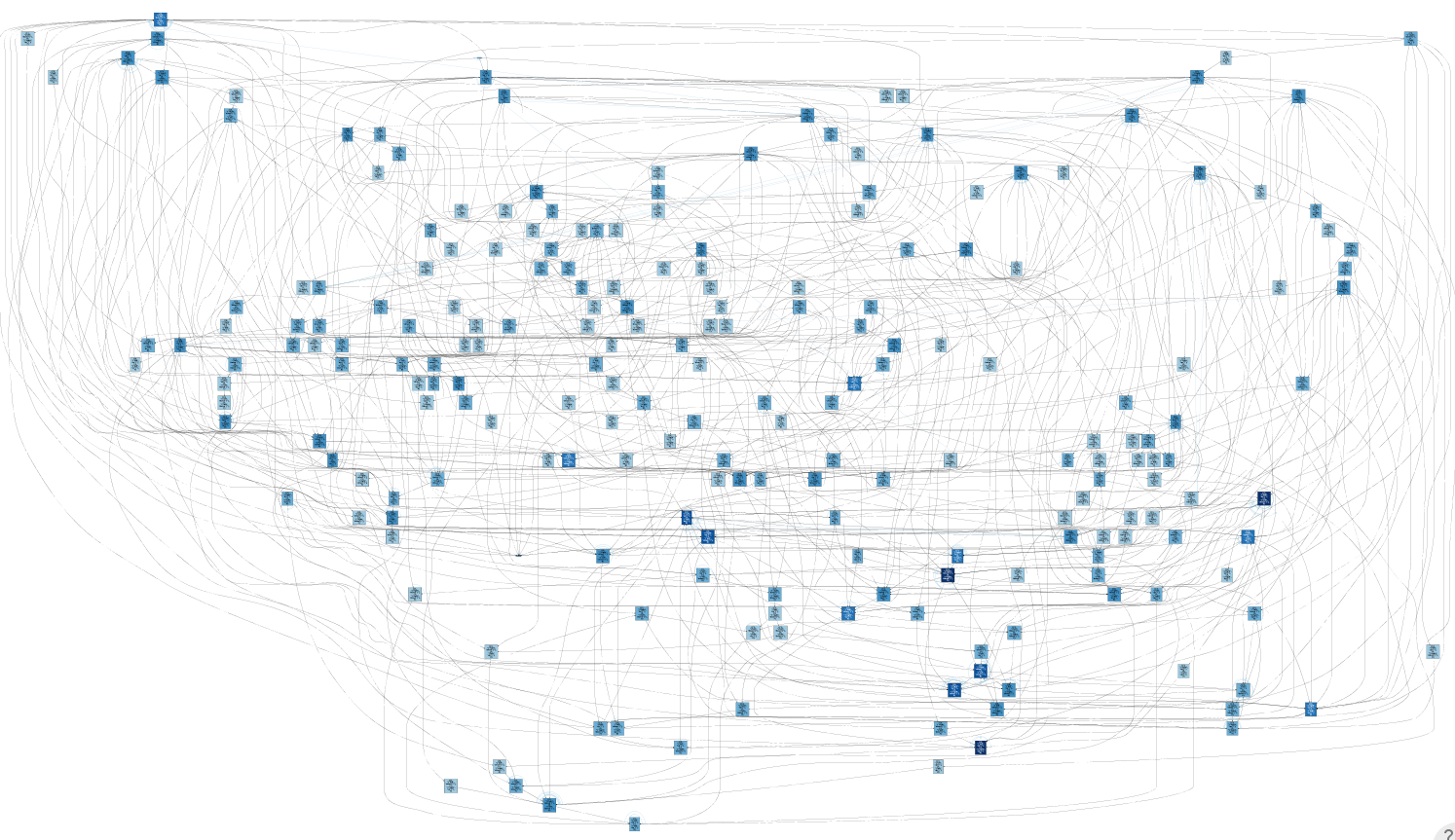}
	\caption{White Agent C-net generated by iDHM (Fixed Iteration Times, Fixed Minimax Search Depth, Simulation Depth = 20)}
	\label{fig: white_simulation_20_iDHM}
\end{figure}

\begin{figure}[!h]
	\centering
	\includegraphics[width=\linewidth]{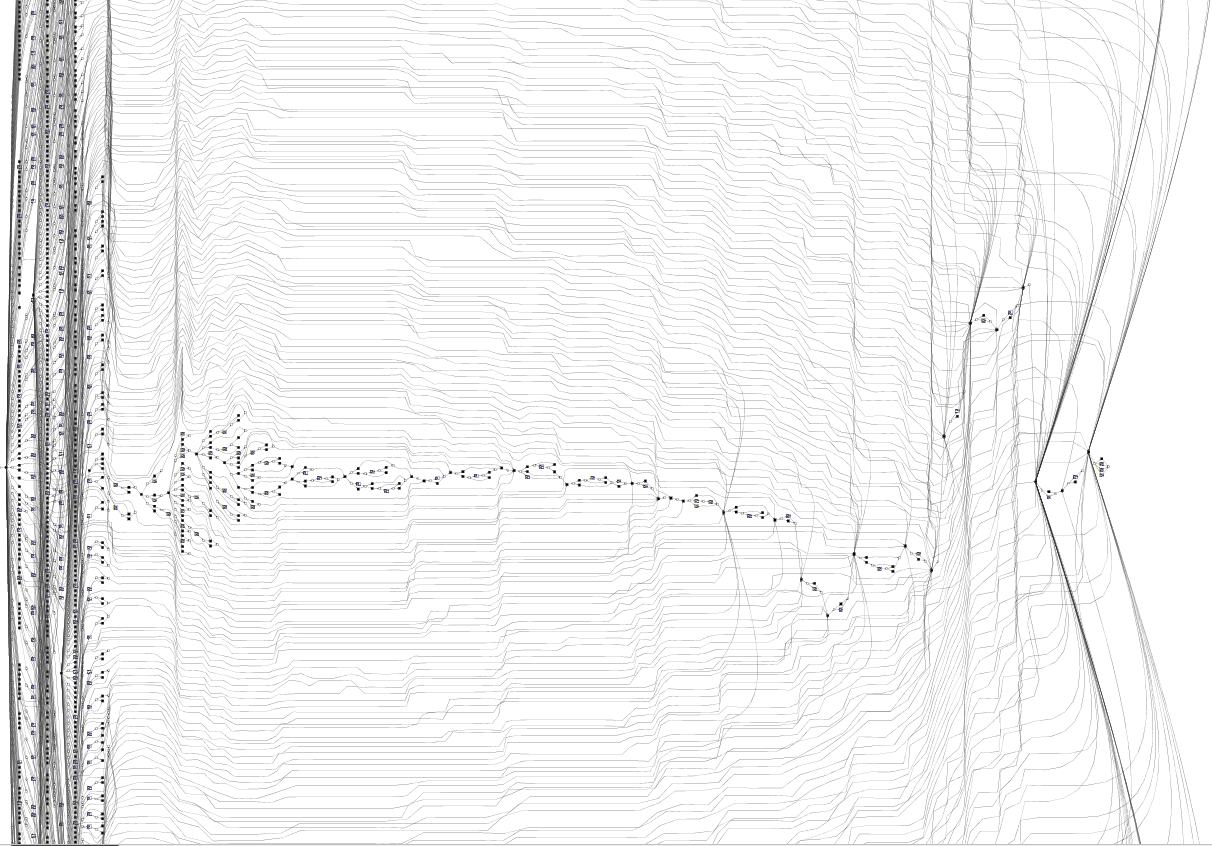}
	\caption{White Agent Petri-net generated by inductive miner algorithm (Fixed Iteration Times, Fixed Minimax Search Depth, Simulation Depth = 20)}
	\label{fig: white_simulation_20_inductive}
\end{figure}

\begin{figure}[!h]
	\centering
	\includegraphics[width=\linewidth]{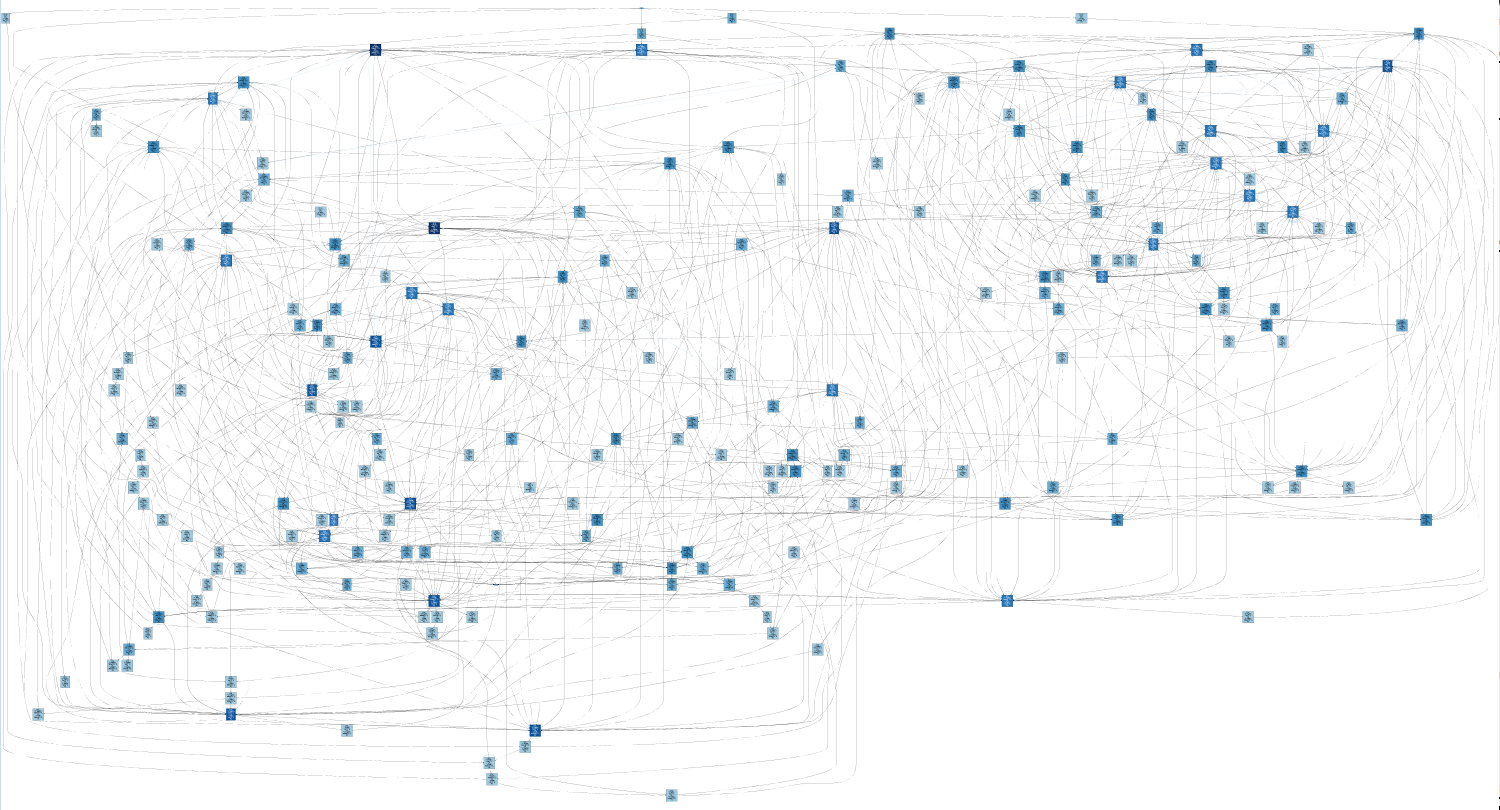}
	\caption{Red agent C-net generated by iDHM (Fixed Iteration Times, Fixed Minimax Search Depth, Simulation Depth = 30)}
	\label{fig: red_simulation_30_iDHM}
\end{figure}

\begin{figure}[!h]
	\centering
	\includegraphics[width=\linewidth]{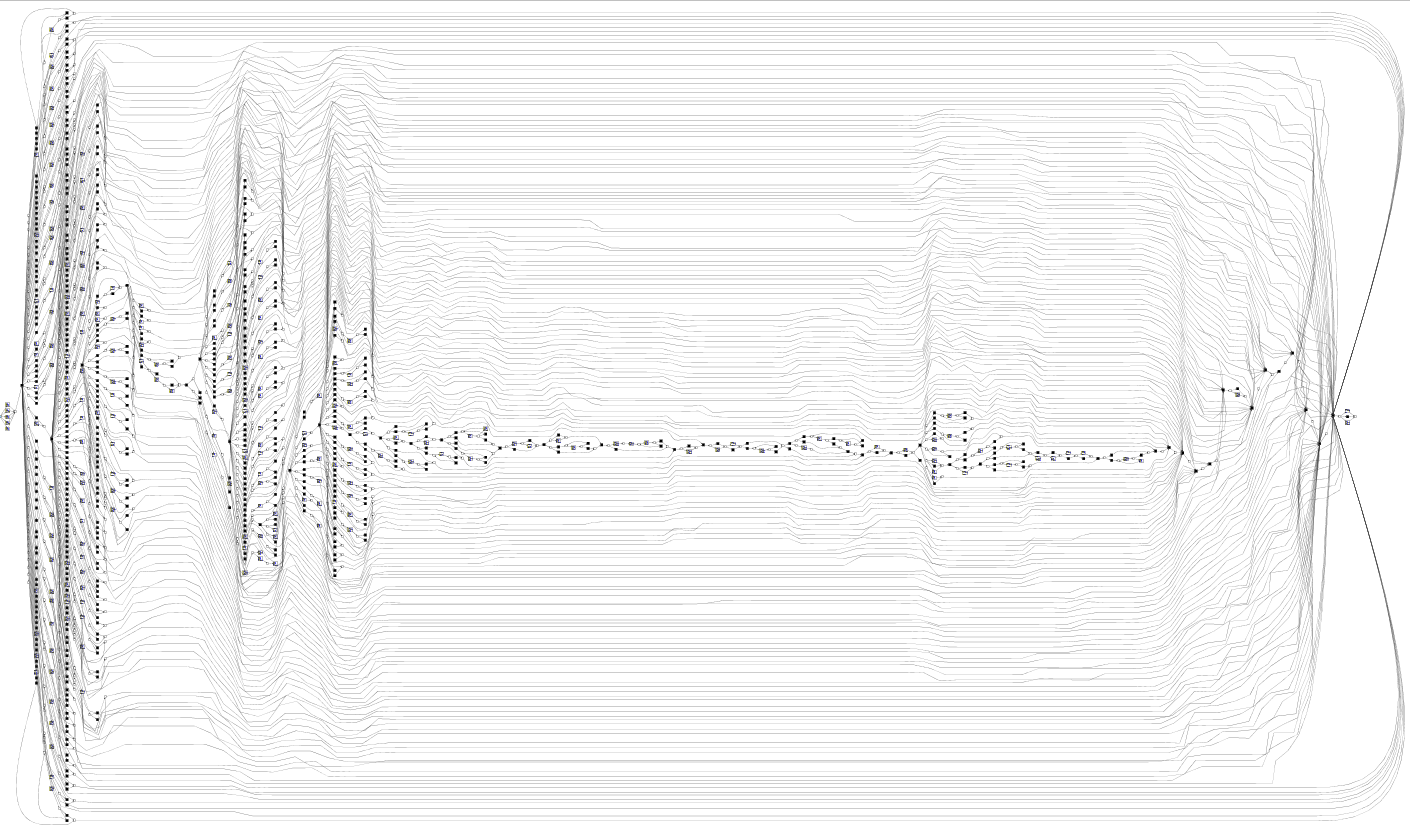}
	\caption{Red agent Petri-net generated by inductive miner algorithm (Fixed Iteration Times, Fixed Minimax Search Depth, Simulation Depth = 30)}
	\label{fig: red_simulation_30_inductive}
\end{figure}

\begin{figure}[!h]
	\centering
	\includegraphics[width=\linewidth]{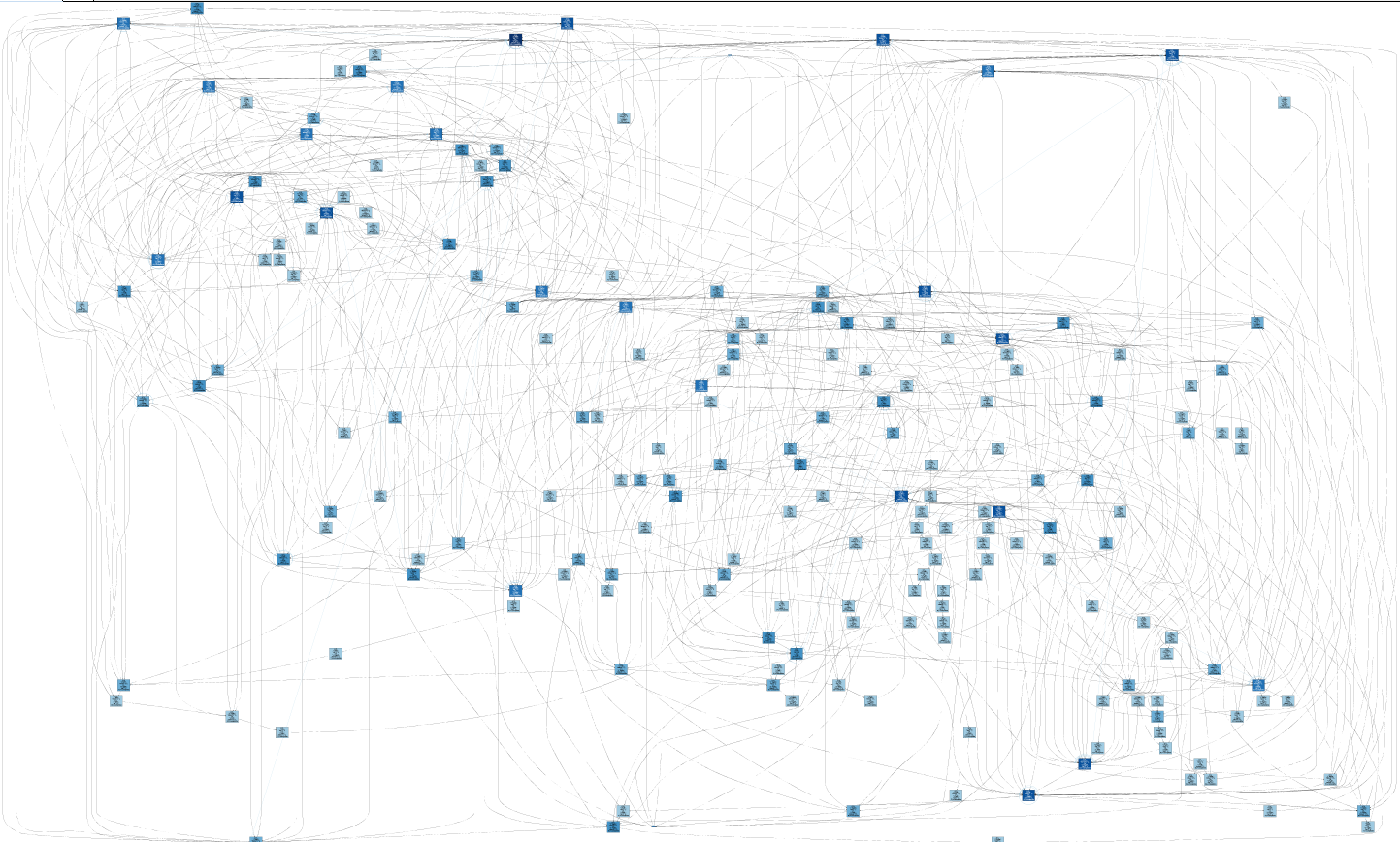}
	\caption{White agent C-net generated by iDHM (Fixed Iteration Times, Fixed Minimax Search Depth, Simulation Depth = 30)}
	\label{fig: white_simulation_30_iDHM}
\end{figure}

\begin{figure}[!h]
	\centering
	\includegraphics[width=\linewidth]{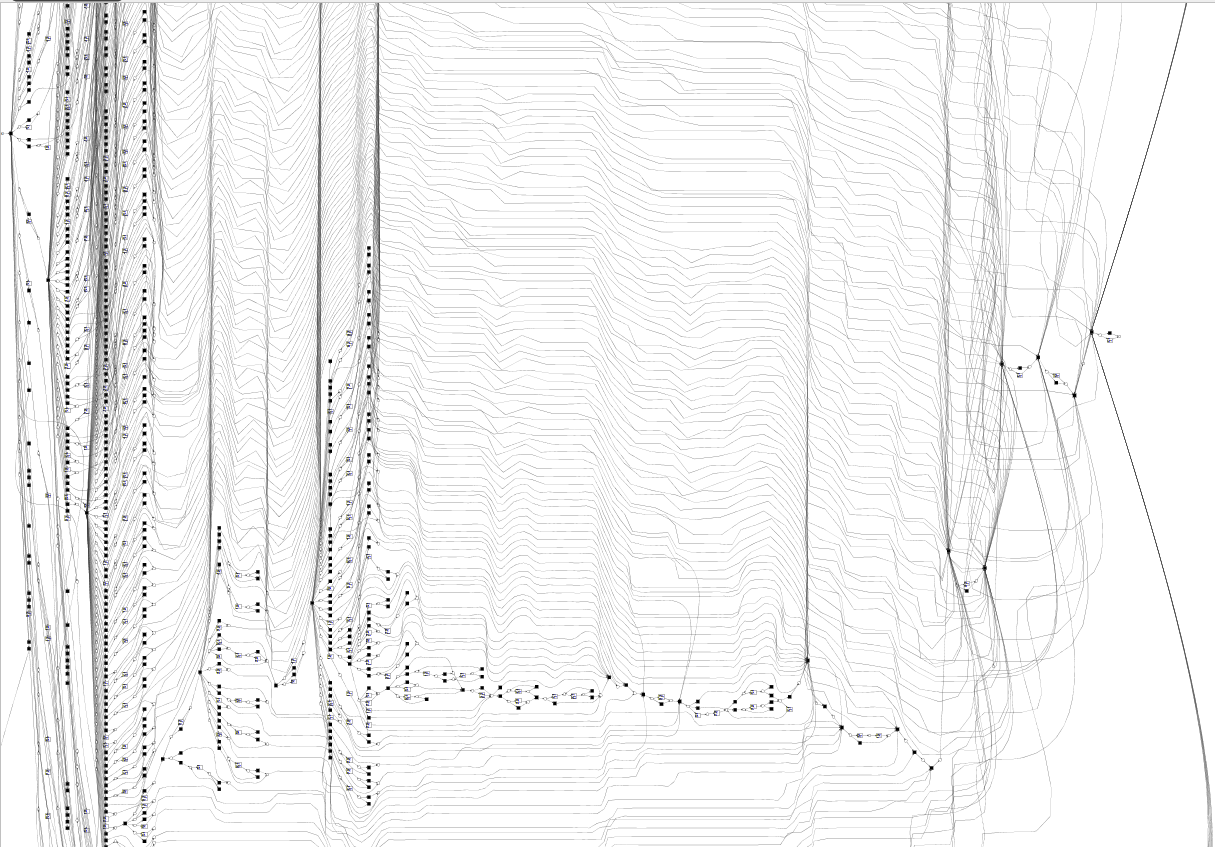}
	\caption{White agent Petri-net generated by inductive miner algorithm (Fixed Iteration Times, Fixed Minimax Search Depth, Simulation Depth = 30)}
	\label{fig: white_simulation_30_inductive}
\end{figure}

\begin{itemize}
	\item Simulation Depth = 10: 
	
	\begin{itemize}
		\item Red agent: Figure~\ref{fig: red_simulation_10_iDHM} shows C-Net generated by iDHM. Figure~\ref{fig: red_simulation_10_inductive} shows Petri-net generated by inductive miner algorithm. 
		
		Model 17 and Model 18 show the global statistics of red agent's process models generated by alpha discovery algorithm and inductive miner algorithm.
		
		For the Petri-net generated by alpha discovery algorithm, its trace fitness, move-model fitness, and move-log fitness are all equal to 1. Thus, it is a $fitting$ model because all three fitness values are perfect.
		
		For the Petri-net generated by inductive miner algorithm, its trace fitness, move-model fitness, and move-log fitness are all equal to 1. Thus, it is a $fitting$ model because all three fitness values are perfect. 
		
		\item White agent: Figure~\ref{fig: white_simulation_10_iDHM} shows C-Net generated by iDHM. Figure~\ref{fig: white_simulation_10_inductive} shows Petri-net generated by inductive miner algorithm. 
		
		Model 19 and Model 20 show the global statistics of white agent's process models generated by alpha discovery algorithm and inductive miner algorithm.
		
		For the Petri-net generated by alpha discovery algorithm, its trace fitness, move-model fitness, and move-log fitness are all equal to 1. Thus, it is a $fitting$ model because all three fitness values are perfect.
		
		For the Petri-net generated by inductive miner algorithm, its trace fitness, move-model fitness, and move-log fitness are all equal to 1. Thus, it is a $fitting$ model because all three fitness values are perfect. 
		
	\end{itemize}
	
	\item Simulation Depth = 20: 
	
	\begin{itemize}
		\item Red agent: Figure~\ref{fig: red_simulation_20_iDHM} shows C-Net generated by iDHM. Figure~\ref{fig: red_simulation_20_inductive} shows Petri-net generated by inductive miner algorithm. 
		
		Model 21 and Model 22 show the global statistics of red agent's process models generated by alpha discovery algorithm and inductive miner algorithm.
		
		For the Petri-net generated by alpha discovery algorithm, its trace fitness, move-model fitness, and move-log fitness are all equal to 1. Thus, it is a $fitting$ model because all three fitness values are perfect.''
		
		For the Petri-net generated by inductive miner algorithm, its trace fitness, move-model fitness, and move-log fitness are all equal to 1. Thus, it is a $fitting$ model because all three fitness values are perfect. 
		
		\item White agent: Figure~\ref{fig: white_simulation_20_iDHM} shows C-Net generated by iDHM. Figure~\ref{fig: white_simulation_20_inductive} shows Petri-net generated by inductive miner algorithm. 
		
		Model 23 and Model 24 show the global statistics of white agent's process models generated by alpha discovery algorithm and inductive miner algorithm.
		
		For the Petri-net generated by alpha discovery algorithm, its trace fitness is 0.14, its move-model fitness is 1.0, and its move-log fitness is 0.12. Thus, it is $non-fitting$ model because trace fitness values and move-log values are both below 1. 
		
		For the Petri-net generated by inductive miner algorithm, its trace fitness, move-model fitness, and move-log fitness are all equal to 1. Thus, it is a $fitting$ model because all three fitness values are perfect. 
		
	\end{itemize}
	
	\item Simulation Depth = 30:
	
	\begin{itemize}
		\item Red agent: Figure~\ref{fig: red_simulation_30_iDHM} shows C-Net generated by iDHM. Figure~\ref{fig: red_simulation_30_inductive} shows Petri-net generated by inductive miner algorithm. 
		
		Model 25 and Model 26 show the global statistics of red agent's process models generated by alpha discovery algorithm and inductive miner algorithm.
		
		For the Petri-net generated by alpha discovery algorithm, its trace fitness is 0.12, its move-model fitness is 0.82, and its move-log fitness is 0.12. Thus, it is $non-fitting$ model because all three fitness values are below 1. 
		
		For the Petri-net generated by inductive miner algorithm, its trace fitness, move-model fitness, and move-log fitness are all equal to 1. Thus, it is a $fitting$ model because all three fitness values are perfect. 
		
		\item White agent: Figure~\ref{fig: white_simulation_30_iDHM} shows C-Net generated by iDHM. Figure~\ref{fig: white_simulation_30_inductive} shows Petri-net generated by inductive miner algorithm. 
		
		Model 27 and Model 28 show the global statistics of white agent's process models generated by alpha discovery algorithm and inductive miner algorithm.
		
		For the Petri-net generated by alpha discovery algorithm, its trace fitness, move-model fitness, and move-log fitness are all equal to 1. Thus, it is a $fitting$ model because all three fitness values are perfect.
		
		For the Petri-net generated by inductive miner algorithm, its trace fitness, move-model fitness, and move-log fitness are all equal to 1. Thus, it is a $fitting$ model because all three fitness values are perfect.
	\end{itemize}
	
\end{itemize}

\begin{table*}[!h]
	\centering
	\begin{tabular}{|l|p{1in}|p{1cm}|p{1cm}|p{1cm}|l|p{15mm}|l|} \hline 
		Model&  Calculation Time (ms)&  Trace Fitness&  Move-Model Fitness&  Move-Log Fitness&  Num. States&  Approx. memory used (kb)& Trace Length\\ \hline 
		Model 17&  4.07&  1.00&  1.00&  1.00&  1.00&  77.40& 88.90\\ \hline 
		Model 18&  3315.54&  1.00&  1.00&  1.00&  200006.10&  12607.70& 88.90\\ \hline 
		Model 19&  15.38&  0.14&  1.00&  0.12&  742.70&  105.30& 88.60\\ \hline 
		Model 20&  4072.42&  1.00&  1.00&  1.00&  200010.60&  11789.60& 88.60\\ \hline 
		Model 21&  4.07&  1.00&  1.00&  1.00&  1.00&  77.40& 88.90\\ \hline 
		Model 22&  3315.54&  1.00&  1.00&  1.00&  200006.10&  12607.70& 88.90\\ \hline 
		Model 23&  15.38&  0.14&  1.00&  0.13&  742.70&  105.30& 88.60\\ \hline 
		Model 24&  4072.41&  1.00&  1.00&  1.00&  200010.60&  11789.60& 88.60\\ \hline 
		Model 25&  20.07&  0.12&  0.82&  0.12&  1333.40&  161.20& 100.30\\ \hline
		Model 26& 3430.38& 1.00& 1.00& 1.00& 200008.50& 12564.30&100.30\\\hline
		Model 27& 5.74& 1.00& 1.00& 1.00& 1.00& 94.50&99.90\\\hline
		Model 28& 3548.60& 1.00& 1.00& 1.00& 200007.30& 12381.70&99.90\\\hline
	\end{tabular}
	\caption{Global statistics: Model 17 to Model 28}
	\label{Table 6}
\end{table*}

\begin{figure}[!h]
	\includegraphics[width=1\linewidth]{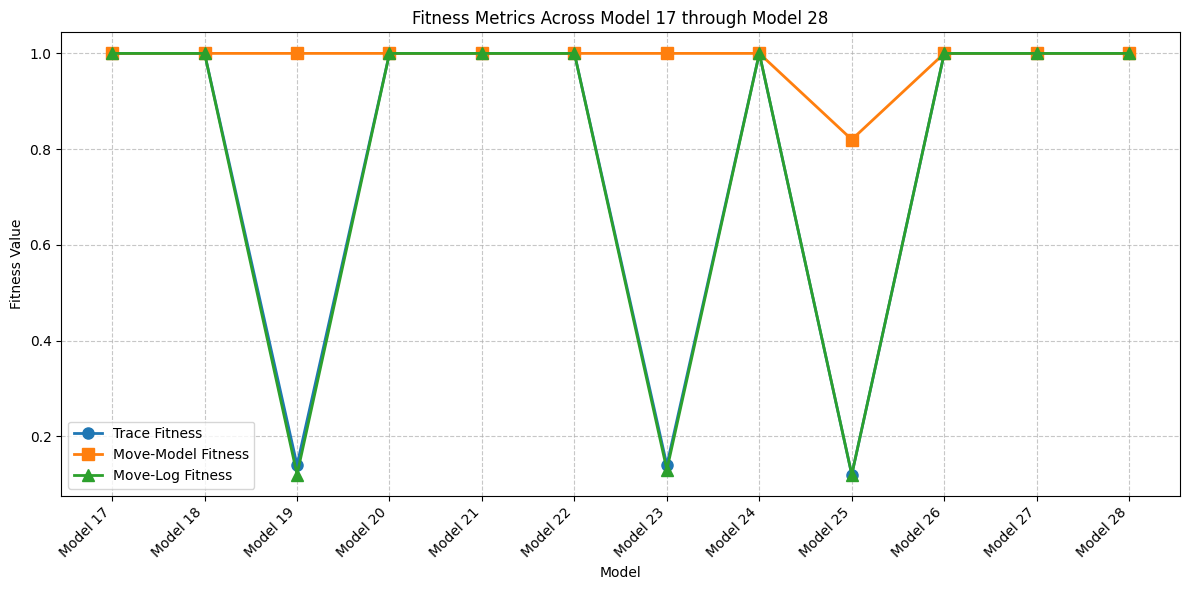}
	\caption{Visual plot shows fitness metrics across Model 17 through Model 28}
	\label{fig:Visualization Plot Table 17 to 28}
\end{figure}

\textbf{Figure~\ref{fig:Visualization Plot Table 17 to 28} shows compared fitness metrics (Trace Fitness, Move-Model Fitness and Move-Log Fitness) across Model 17 through Model 28}.


\textbf{\\ Consistent High Fitness:}

The majority of the models, including Models 17, 18, 20, 21, 22, 24, 26, 27, and 28, demonstrate perfect Trace Fitness, Move-Model Fitness, and Move-Log Fitness with a high value of 1.00. This consistency indicates a strong alignment between the model and the actual moves and logs data for these datasets, suggesting that the model accurately represents the processes in these instances.

\textbf{\\ Outliers in Fitness Metrics:}

In contrast, a few models exhibit lower fitness values, deviating significantly from the majority. Models 19 and 23 show a substantial decline in Move-Log Fitness, dropping to 0.12 and 0.13 respectively, while maintaining a perfect Move-Model Fitness of 1.00. This suggests that while the model structure remains consistent, there are discrepancies between the model and the actual log data for these specific models. Model 25 presents a unique case where the Move-Model Fitness decreases to 0.82, while the Move-Log Fitness drops to 0.12, indicating potential issues in both model representation and log alignment for this dataset. Interestingly, Trace Fitness remains consistently high at 1.00 across all models, suggesting that the tracing aspect of the model performs well regardless of other fluctuations. These variations highlight the importance of considering multiple fitness metrics to gain a comprehensive understanding of model performance across different datasets.

\begin{table*}[!h]
	\centering
	\begin{tabular}{|l|c|c|c|c|}
		\hline
		Metric & R.A-A.D 10 & R.A-I.M 10 & W.A-A.D 10 & W.A-I.M 10 \\
		\hline
		Calc. Time (ms) & 4.07 & 3315.54 & 15.38 & 4072.42 \\
		Num. States & 1.00 & 200006.10 & 742.70 & 200010.60 \\
		Trace Fitness & 1.00 & 1.00 & 0.14 & 1.00 \\
		Raw Fitness Cost & 0.00 & 0.00 & 77.10 & 0.00 \\
		Move-Model Fitness & 1.00 & 1.00 & 1.00 & 1.00 \\
		Pre-process time (ms) & 0.69 & 0.38 & 0.30 & 0.52 \\
		Move-Log Fitness & 1.00 & 1.00 & 0.12 & 1.00 \\
		Trace Length & 88.90 & 88.90 & 88.60 & 88.60 \\
		Approx. mem. used (kb) & 77.40 & 12607.70 & 105.30 & 11789.60 \\
		\hline
	\end{tabular}
	
	\vspace{0.5cm}
	
	\begin{tabular}{|l|c|c|c|c|}
		\hline
		Metric & R.A-A.D 20 & R.A-I.M 20 & W.A-A.D 20 & W.A-I.M 20 \\
		\hline
		Calc. Time (ms) & 4.07 & 3315.54 & 15.38 & 4072.41 \\
		Num. States & 1.00 & 200006.10 & 742.70 & 200010.60 \\
		Trace Fitness & 1.00 & 1.00 & 0.14 & 1.00 \\
		Raw Fitness Cost & 0.00 & 0.00 & 77.10 & 0.00 \\
		Move-Model Fitness & 1.00 & 1.00 & 1.00 & 1.00 \\
		Pre-process time (ms) & 0.69 & 0.38 & 0.30 & 0.52 \\
		Move-Log Fitness & 1.00 & 1.00 & 0.13 & 1.00 \\
		Trace Length & 88.90 & 88.90 & 88.60 & 88.60 \\
		Approx. mem. used (kb) & 77.40 & 12607.70 & 105.30 & 11789.60 \\
		\hline
	\end{tabular}
	
	\vspace{0.5cm}
	
	\begin{tabular}{|l|c|c|c|c|}
		\hline
		Metric & R.A-A.D 30 & R.A-I.M 30 & W.A-A.D 30 & W.A-I.M 30 \\
		\hline
		Calc. Time (ms) & 20.07 & 3430.38 & 5.74 & 3548.60 \\
		Num. States & 1333.40 & 200008.50 & 1.00 & 200007.30 \\
		Trace Fitness & 0.12 & 1.00 & 1.00 & 1.00 \\
		Raw Fitness Cost & 88.30 & 0.00 & 0.00 & 0.00 \\
		Move-Model Fitness & 0.82 & 1.00 & 1.00 & 1.00 \\
		Pre-process time (ms) & 0.29 & 0.48 & 0.32 & 0.57 \\
		Move-Log Fitness & 0.12 & 1.00 & 1.00 & 1.00 \\
		Trace Length & 100.30 & 100.30 & 99.90 & 99.90 \\
		Approx. mem. used (kb) & 161.20 & 12564.30 & 94.50 & 12381.70 \\
		\hline
	\end{tabular}
	\caption{Global statistics: Petri-net generated by various algorithms (W.A white agent, R.A red agent, I.M inductive miner, A.D alpha discovery) (Fixed Iteration Times, Fixed Minimax Search Depth, Simulation Depth = n)}
	\label{tab:petri-net-stats}
\end{table*}

\subsubsection {Trial 3: Fixed Simulation Depth, Fixed Iteration Times, Variable Minimax Search Depth} \label{Trial 3}

\begin{figure}[!h]
	\centering
	\includegraphics[width=\linewidth]{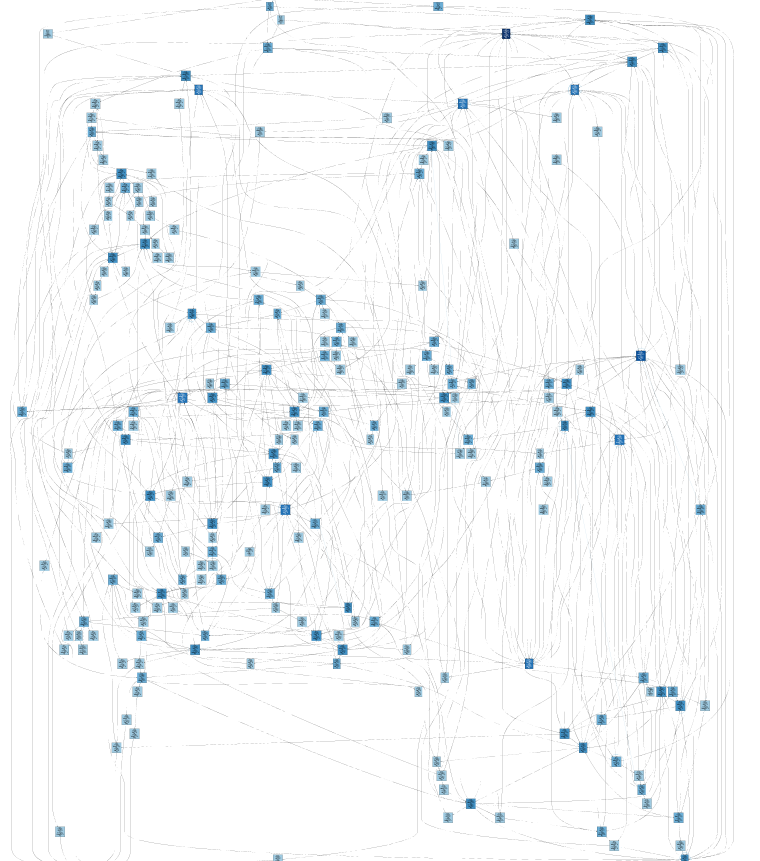}
	\caption{Red agent C-net generated by iDHM (Fixed Simulation Depth, Fixed Iteration Times, Minimax Search Depth = 1)}
	\label{fig: red_minimax_1_iDHM}
\end{figure}

\begin{figure}[!h]
	\centering
	\includegraphics[width=\linewidth]{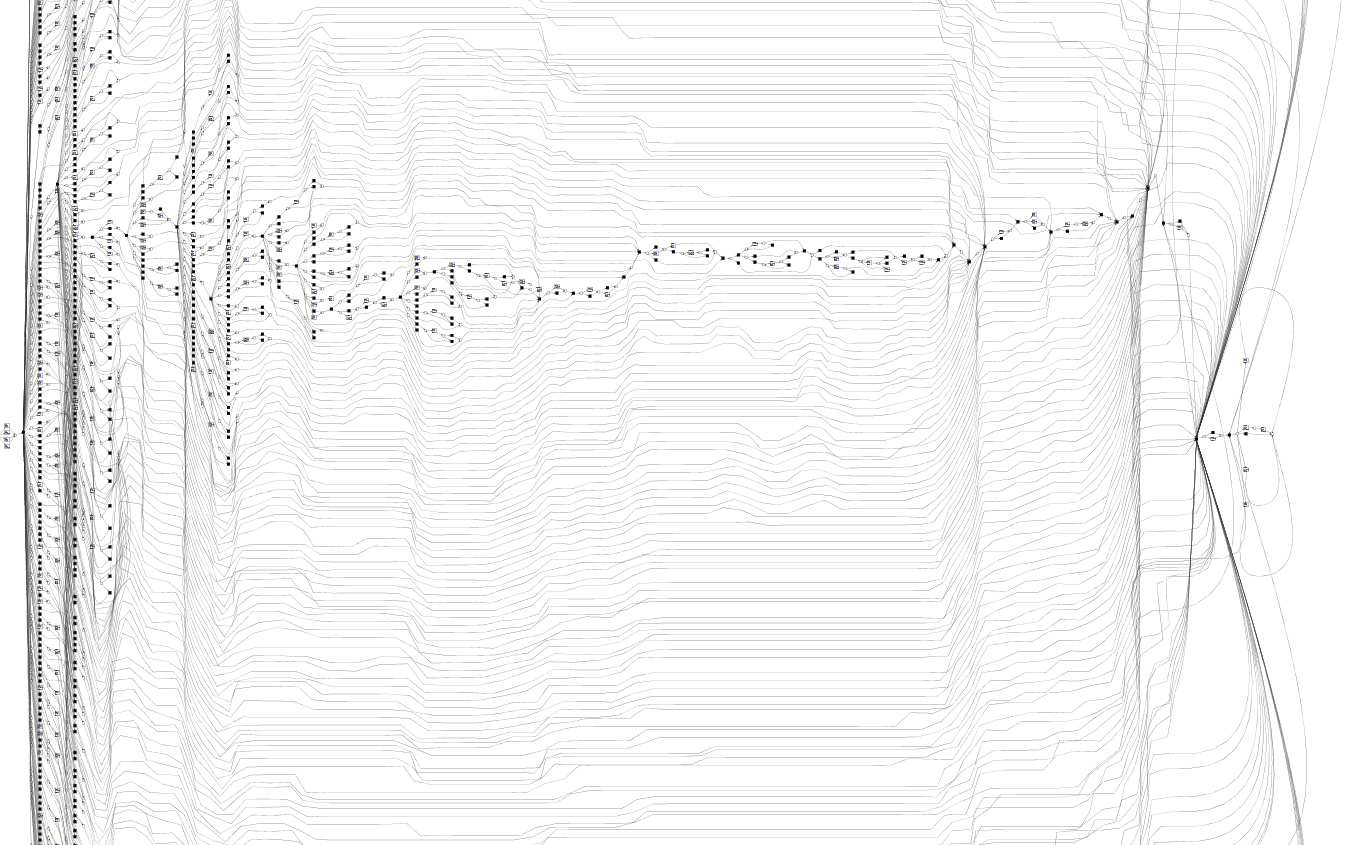}
	\caption{Red agent Petri-net generated by inductive miner algorithm (Fixed Simulation Depth, Fixed Iteration Times, Minimax Search Depth = 1)}
	\label{fig: red_minimax_1_inductive}
\end{figure}

\begin{figure}[!h]
	\centering
	\includegraphics[width=\linewidth]{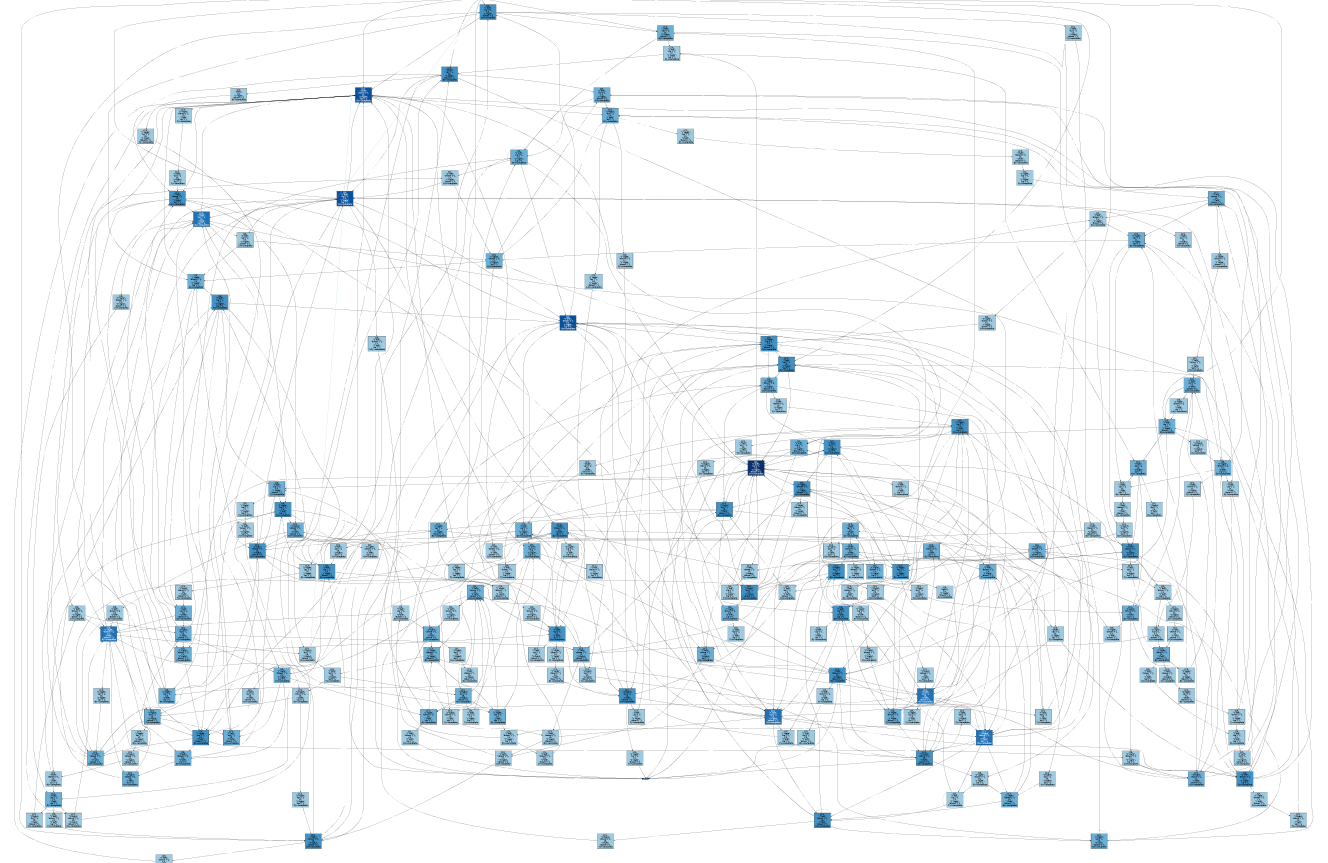}
	\caption{White agent C-net generated by iDHM (Fixed Simulation Depth, Fixed Iteration Times, Minimax Search Depth = 1)}
	\label{fig: white_minimax_1_iDHM}
\end{figure}

\begin{figure}[!h]
	\centering
	\includegraphics[width=\linewidth]{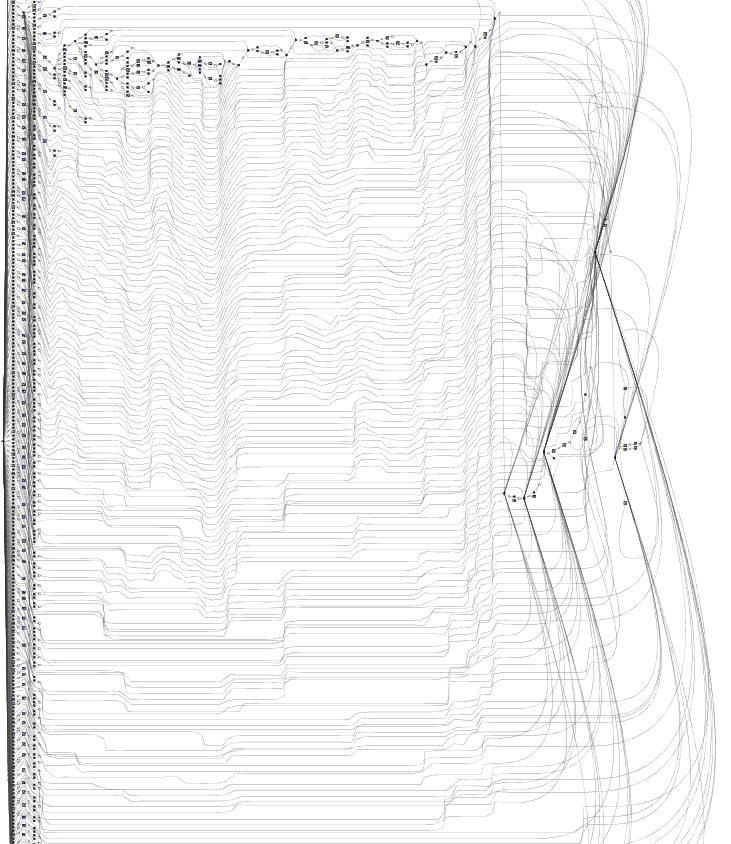}
	\caption{White agent Petri-net generated by inductive miner algorithm (Fixed Simulation Depth, Fixed Iteration Times, Minimax Search Depth = 1)}
	\label{fig: white_minimax_1_inductive}
\end{figure}

\begin{figure}[!h]
	\centering
	\includegraphics[width=\linewidth]{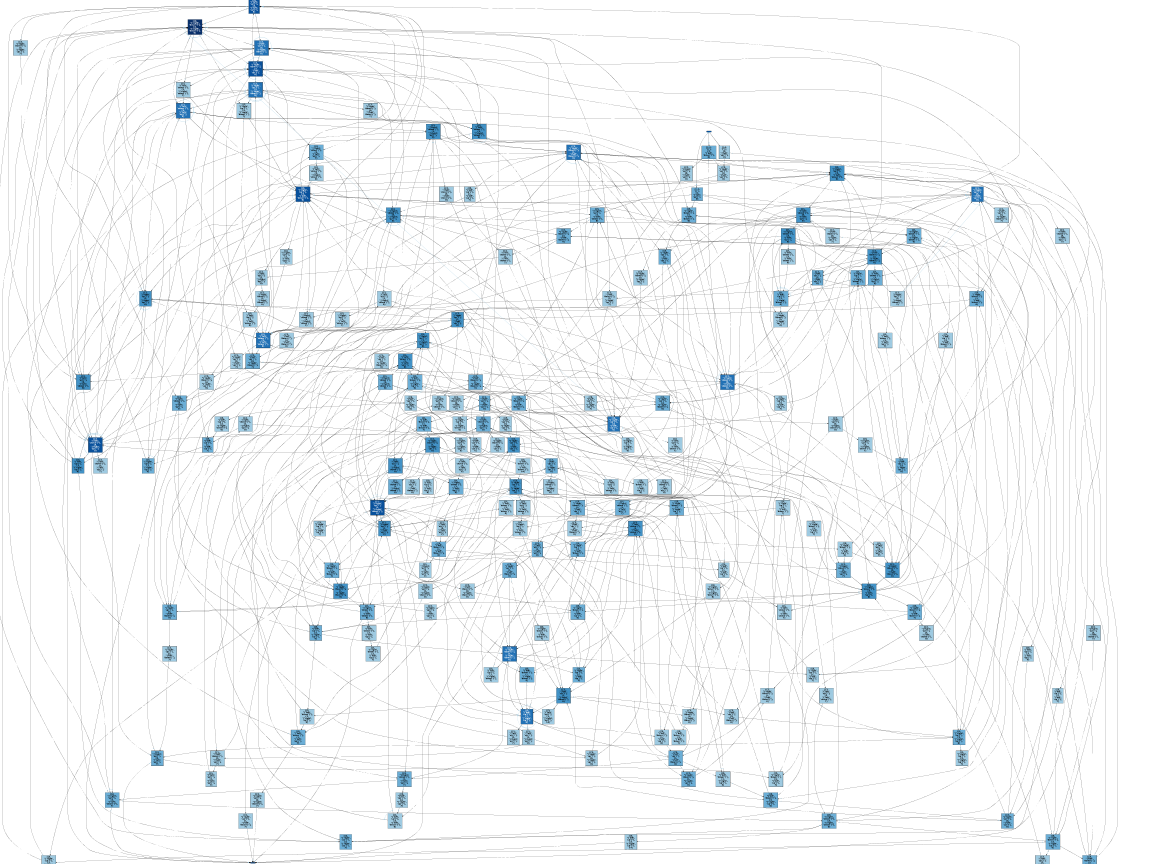}
	\caption{Red agent C-net generated by iDHM (Fixed Simulation Depth, Fixed Iteration Times, Minimax Search Depth = 2)}
	\label{fig: red_minimax_2_iDHM}
\end{figure}

\begin{figure}[!h]
	\centering
	\includegraphics[width=\linewidth]{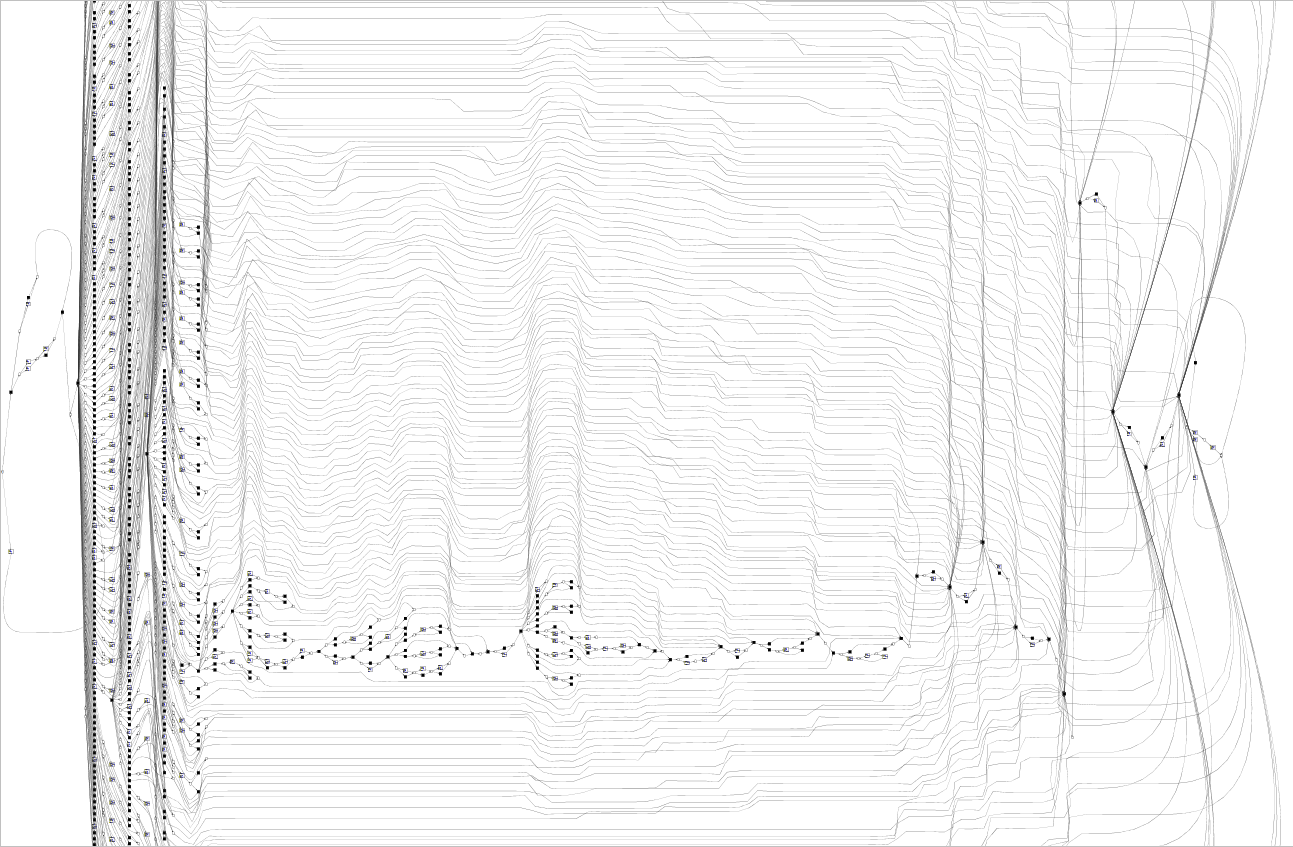}
	\caption{Red agent Petri-net generated by inductive miner algorithm (Fixed Simulation Depth, Fixed Iteration Times, Minimax Search Depth = 2)}
	\label{fig: red_minimax_2_inductive}
\end{figure}

\begin{figure}[!h]
	\centering
	\includegraphics[width=\linewidth]{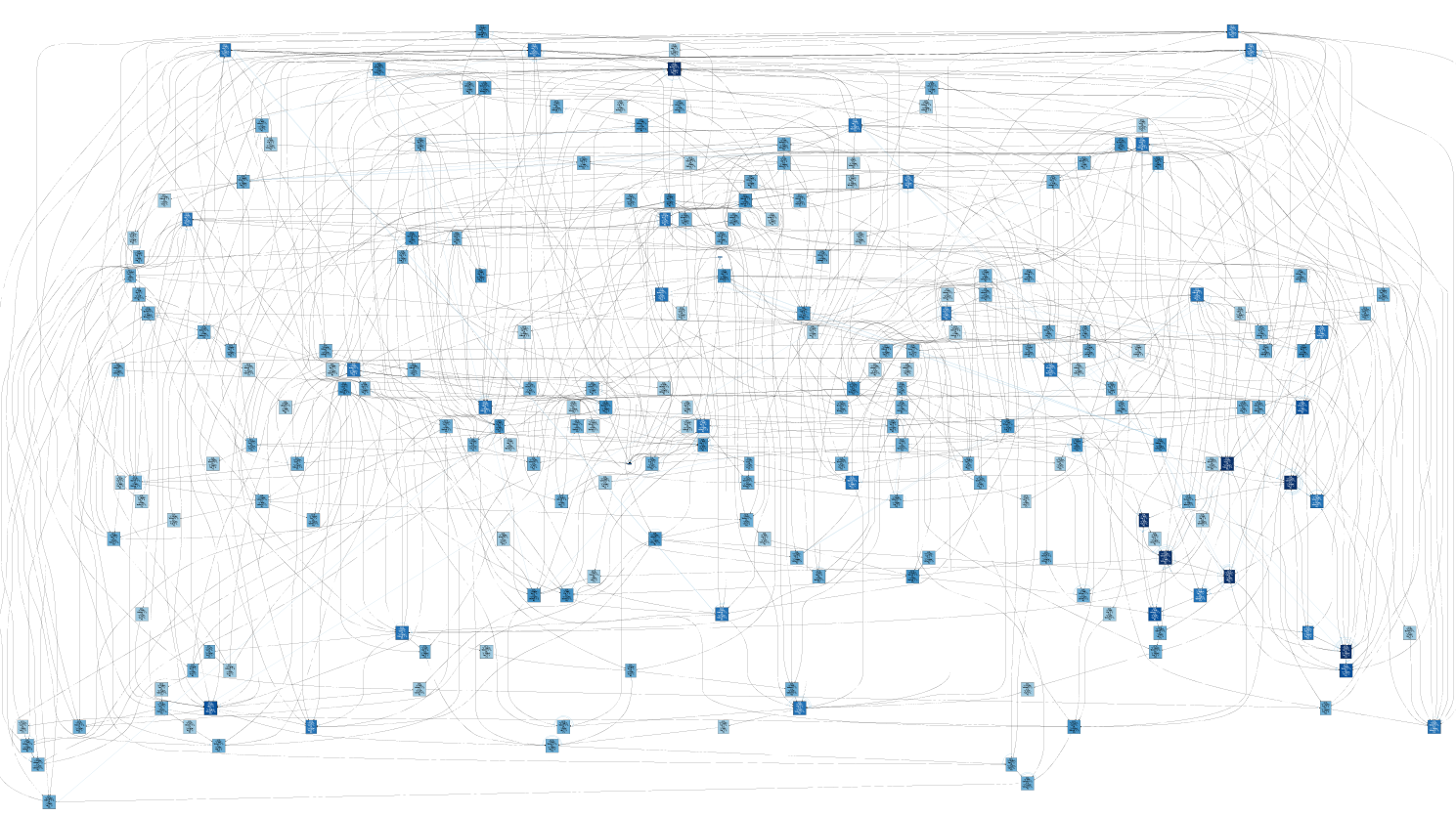}
	\caption{White agent C-net generated by iDHM (Fixed Simulation Depth, Fixed Iteration Times, Minimax Search Depth = 2)}
	\label{fig: white_minimax_2_iDHM}
\end{figure}

\begin{figure}[!h]
	\centering
	\includegraphics[width=\linewidth]{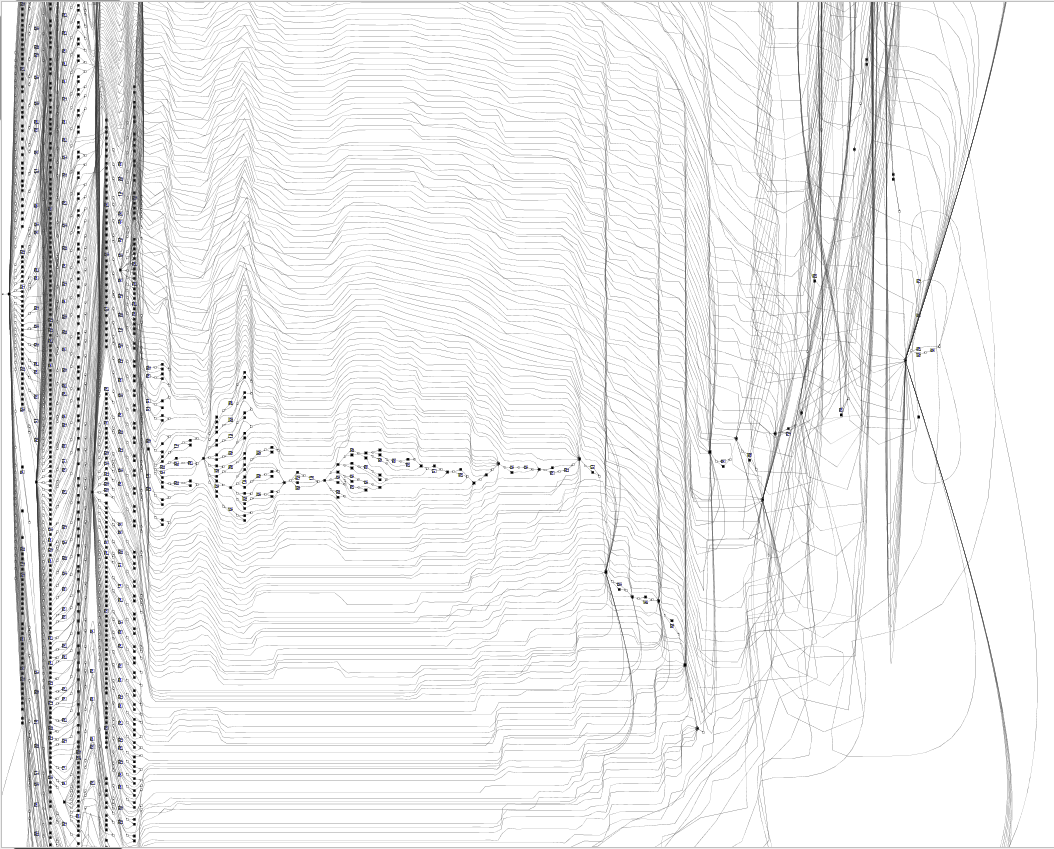}
	\caption{White agent Petri-net generated by inductive miner algorithm (Fixed Simulation Depth, Fixed Iteration Times, Minimax Search Depth = 2)}
	\label{fig: white_minimax_2_inductive}
\end{figure}

\begin{figure}[!h]
	\centering
	\includegraphics[width=\linewidth]{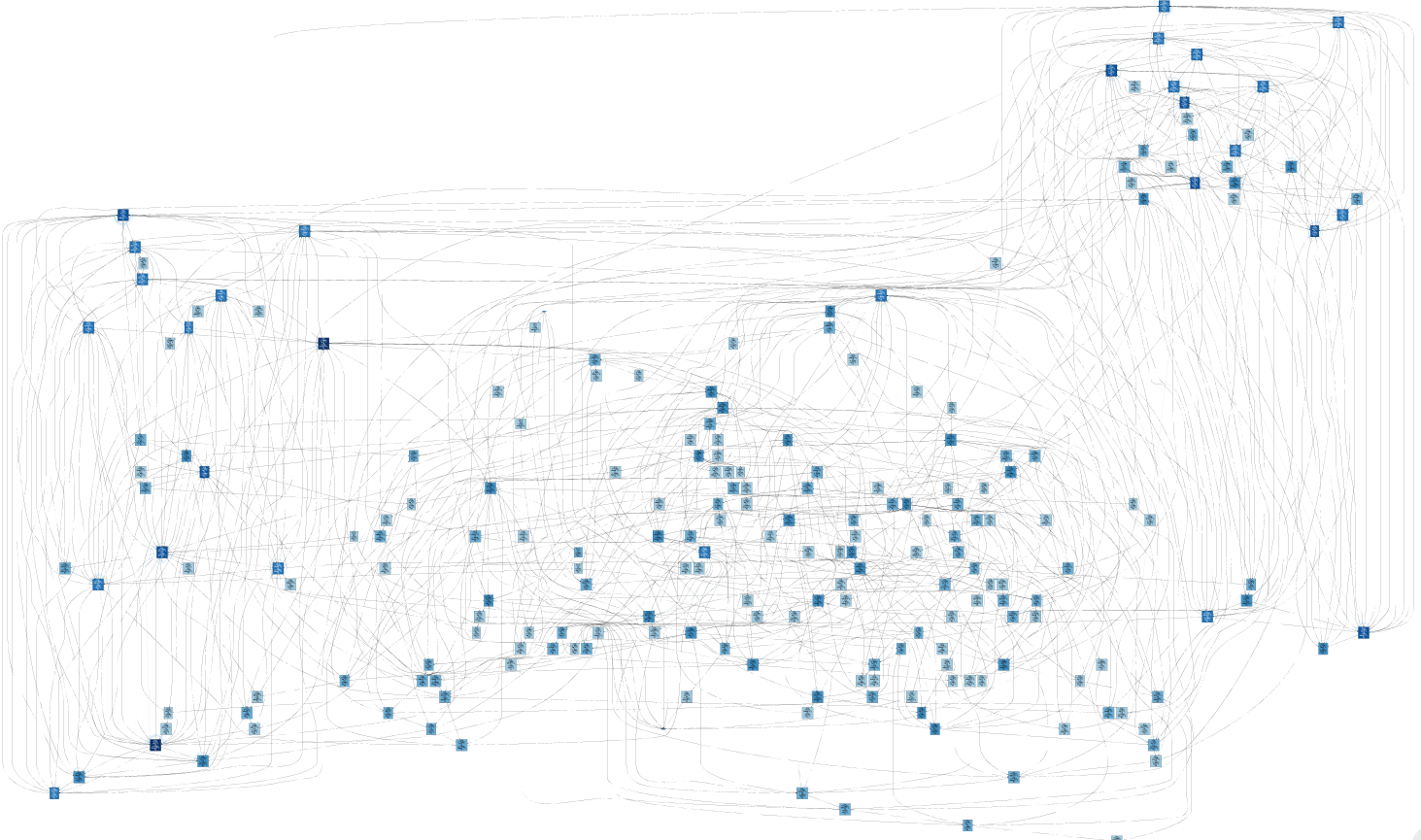}
	\caption{Red agent C-net generated by iDHM (Fixed Simulation Depth, Fixed Iteration Times, Minimax Search Depth = 3)}
	\label{fig: red_minimax_3_iDHM}
\end{figure}

\begin{figure}[!h]
	\centering
	\includegraphics[width=\linewidth]{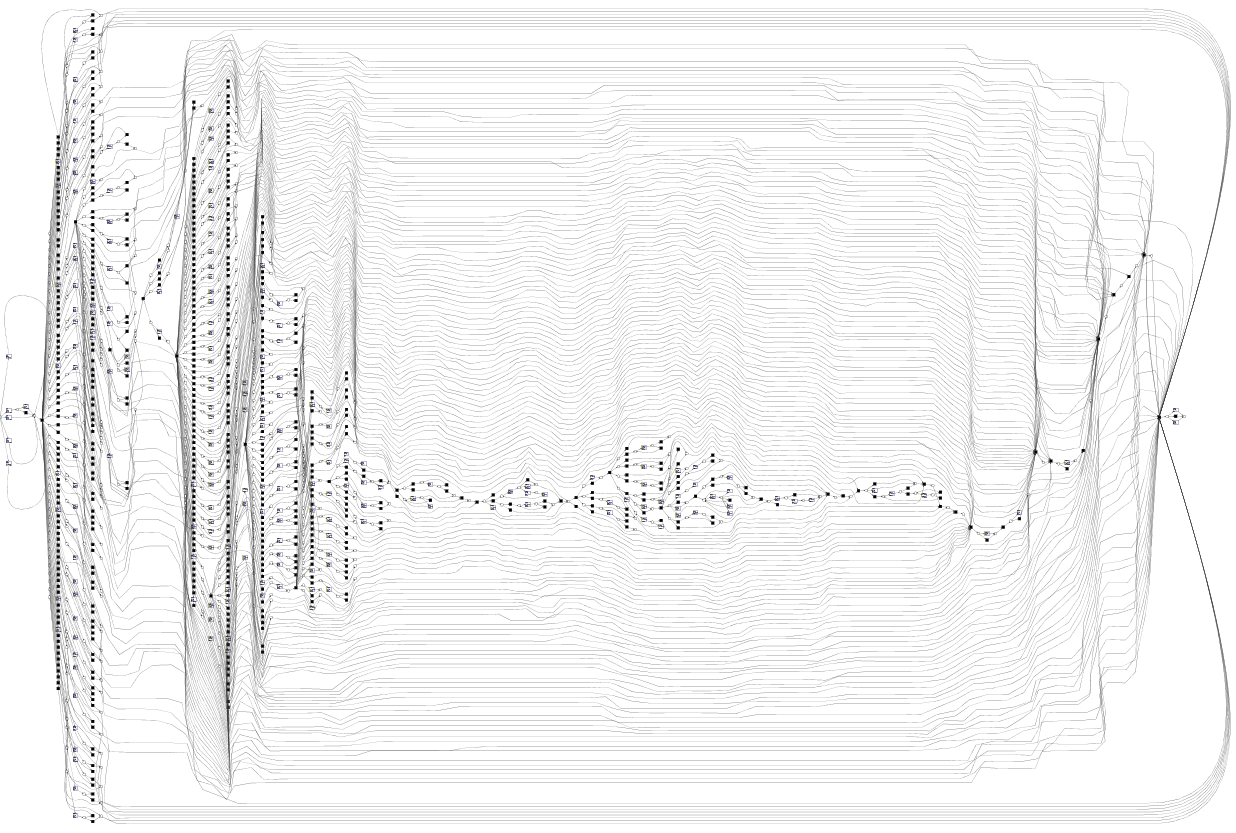}
	\caption{Red agent Petri-net generated by inductive miner algorithm (Fixed Simulation Depth, Fixed Iteration Times, Minimax Search Depth = 3)}
	\label{fig: red_minimax_3_inductive}
\end{figure}

\begin{figure}[!h]
	\centering
	\includegraphics[width=\linewidth]{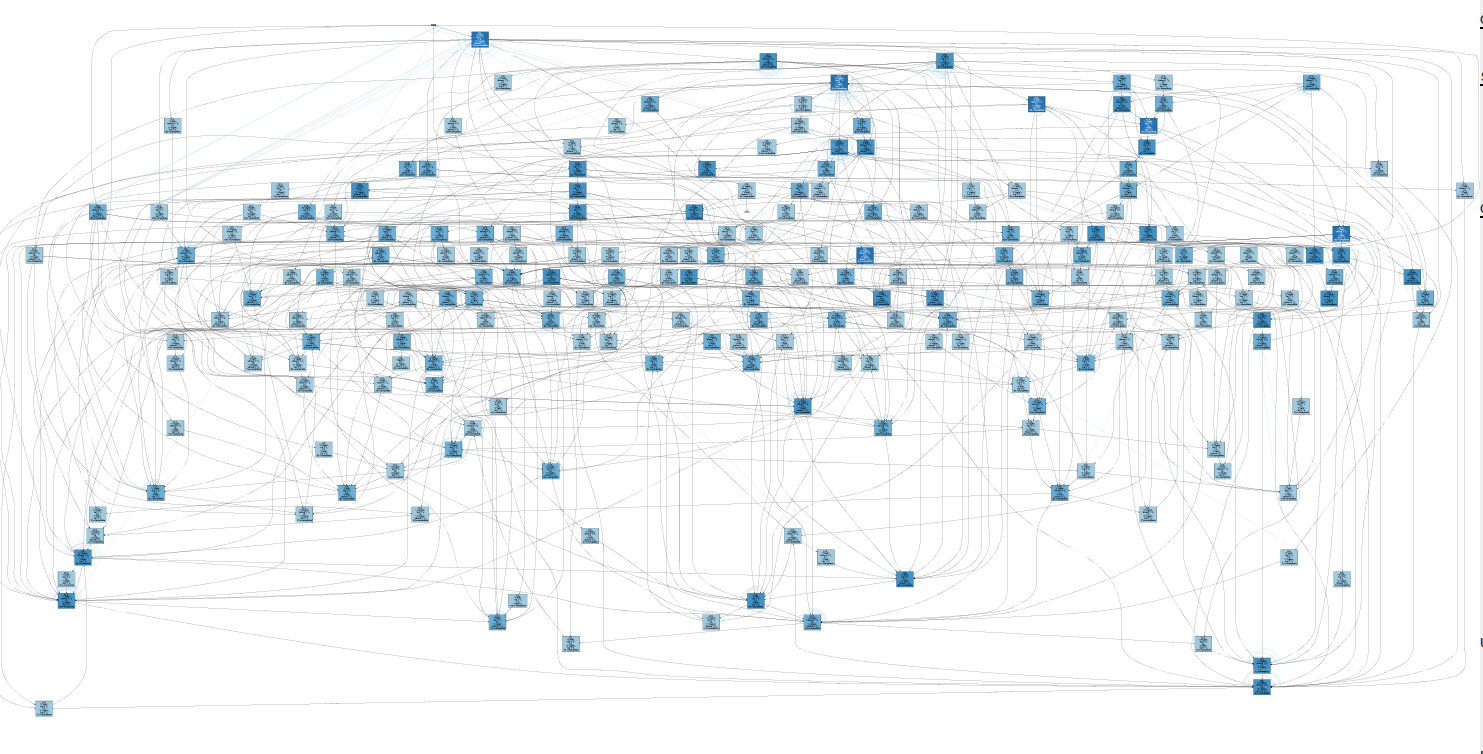}
	\caption{White agent C-net generated by iDHM (Fixed Simulation Depth, Fixed Iteration Times, Minimax Search Depth = 3)}
	\label{fig: white_minimax_3_iDHM}
\end{figure}

\begin{figure}[!h]
	\centering
	\includegraphics[width=\linewidth]{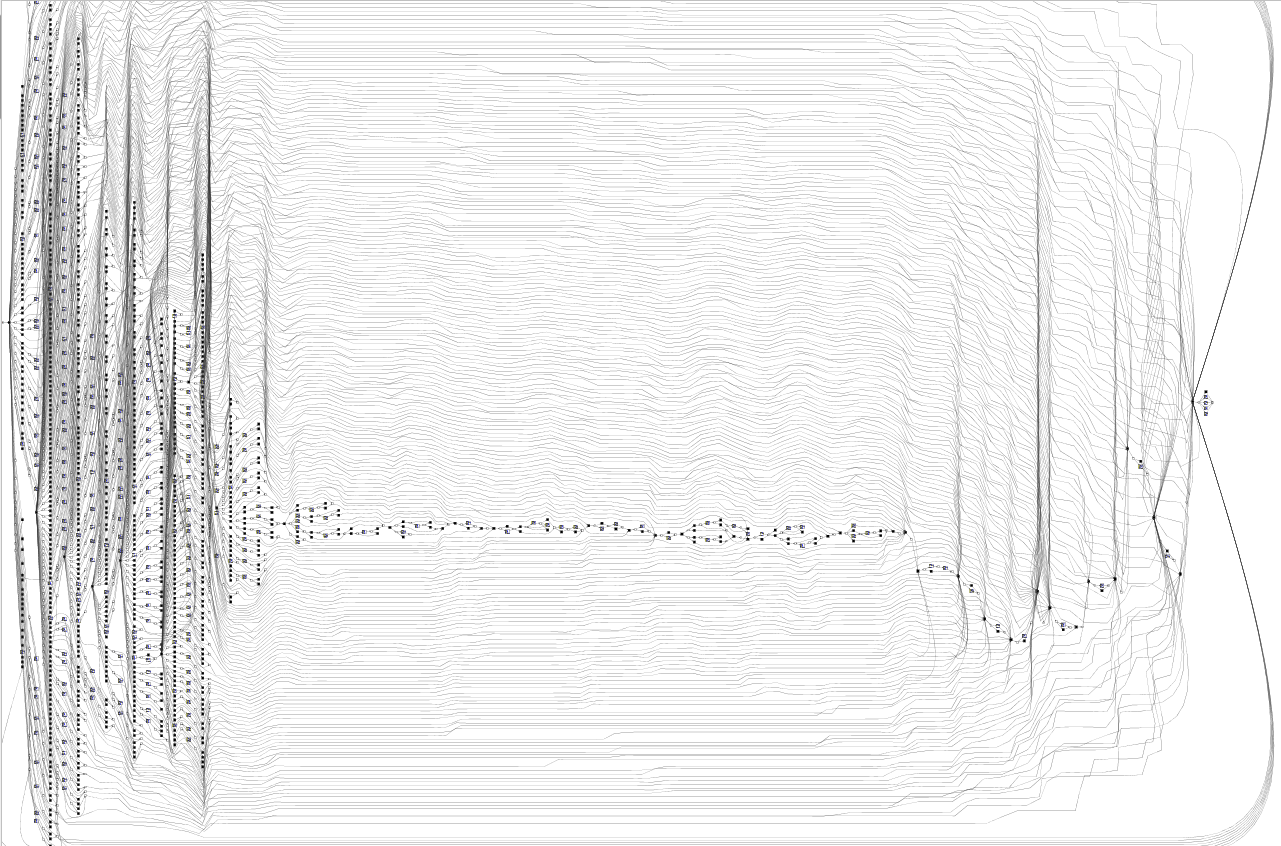}
	\caption{White agent Petri-net generated by inductive miner algorithm (Fixed Simulation Depth, Fixed Iteration Times, Minimax Search Depth = 3)}
	\label{fig: white_minimax_3_inductive}
\end{figure}

\begin{itemize}
	\item Minimax Search Depth = 1: 
	
	\begin{itemize}
		\item Red agent: Figure~\ref{fig: red_minimax_1_iDHM} shows C-Net generated by iDHM. Figure~\ref{fig: red_minimax_1_inductive} shows Petri-net generated by inductive miner algorithm. 
		
		Model 29 and Model 30 show the global statistics of red agent's process models generated by alpha discovery algorithm and inductive miner algorithm.
		
		For the Petri-net generated by alpha discovery algorithm, its trace fitness is 0.94, its move-model fitness is 0.09, and its move-log fitness is 0.09. Thus, it is $non-fitting$ model because all three fitness values are below 1. 
		
		For the Petri-net generated by inductive miner algorithm, its trace fitness, move-model fitness, and move-log fitness are all equal to 1. Thus, it is a $fitting$ model because all three fitness values are perfect. 
		
		\item White agent: Figure~\ref{fig: white_minimax_1_iDHM} shows C-Net generated by iDHM. Figure~\ref{fig: white_minimax_1_inductive} shows Petri-net generated by inductive miner algorithm. 
		
		Model 31 and Model 32 show the global statistics of white agent's process models generated by alpha discovery algorithm and inductive miner algorithm.
		
		For the Petri-net generated by alpha discovery algorithm, its trace fitness, move-model fitness, and move-log fitness are all equal to 1. Thus, it is a $fitting$ model because all three fitness values are perfect.
		
		For the Petri-net generated by inductive miner algorithm, its trace fitness, move-model fitness, and move-log fitness are all equal to 1. Thus, it is a $fitting$ model because all three fitness values are perfect. 
		
	\end{itemize}
	
	\item Minimax Search Depth = 2: 
	
	\begin{itemize}
		\item Red agent: Figure~\ref{fig: red_minimax_2_iDHM} shows C-Net generated by iDHM. Figure~\ref{fig: red_minimax_2_inductive} shows Petri-net generated by inductive miner algorithm. 
		
		Model 33 and Model 34 show the global statistics of red agent's process models generated by alpha discovery algorithm and inductive miner algorithm.
		
		For the Petri-net generated by alpha discovery algorithm, its perfect trace fitness, move-model fitness, and move-log fitness are all equal to 1. Thus, it is a $fitting$ model because all three fitness values are perfect.
		
		For the Petri-net generated by inductive miner algorithm, its perfect trace fitness, move-model fitness, and move-log fitness are all equal to 1. Thus, it is a $fitting$ model because all three fitness values are perfect.
		
		\item White agent: Figure~\ref{fig: white_minimax_2_iDHM} shows C-Net generated by iDHM. Figure~\ref{fig: white_minimax_2_inductive} shows Petri-net generated by inductive miner algorithm. 
		
		Model 35 and Model 36 show the global statistics of white agent's process models generated by alpha discovery algorithm and inductive miner algorithm.
		
		For the Petri-net generated by alpha discovery algorithm, its trace fitness is 0.15, its move-model fitness is 0.98, and its move-log fitness is 0.13. Thus, it is $non-fitting$ model because all three fitness values are below 1. 
		
		For the Petri-net generated by inductive miner algorithm, its perfect trace fitness, move-model fitness, and move-log fitness are all equal to 1. Thus, it is a $fitting$ model because all three fitness values are perfect.
		
	\end{itemize}
	
	\item Minimax Search Depth = 3:
	
	\begin{itemize}
		\item Red agent: Figure~\ref{fig: red_minimax_3_iDHM} shows C-Net generated by iDHM. Figure~\ref{fig: red_minimax_3_inductive} shows Petri-net generated by inductive miner algorithm. 
		
		Model 37 and Model 38 show the global statistics of red agent's process models generated by alpha discovery algorithm and inductive miner algorithm.
		
		For the Petri-net generated by alpha discovery algorithm, its perfect trace fitness, move-model fitness, and move-log fitness are all equal to 1. Thus, it is a $fitting$ model because all three fitness values are perfect.
		
		For the Petri-net generated by inductive miner algorithm, its perfect trace fitness, move-model fitness, and move-log fitness are all equal to 1. Thus, it is a $fitting$ model because all three fitness values are perfect. 
		
		\item White agent: Figure~\ref{fig: white_minimax_3_iDHM} shows C-Net generated by iDHM. Figure~\ref{fig: white_minimax_3_inductive} shows Petri-net generated by inductive miner algorithm. 
		
		Model 39 and Model 40 show the global statistics of white agent's process models generated by alpha discovery algorithm and inductive miner algorithm.
		
		For the Petri-net generated by alpha discovery algorithm, its trace fitness is 0.21, its move-model fitness is 0.99, and its move-log fitness is 0.19. Thus, it is $non-fitting$ model because all three fitness values are below 1.
		
		For the Petri-net generated by inductive miner algorithm, its perfect trace fitness, move-model fitness, and move-log fitness are all equal to 1. Thus, it is a $fitting$ model because all three fitness values are perfect. 
		
	\end{itemize}
	
\end{itemize}

\begin{table*}[!h]
	\centering
	\begin{tabular}{|l|p{17mm}|p{1cm}|p{1cm}|p{1cm}|c|p{1in}|c|} \hline 
		Model&  Calculation Time (ms)&  Trace Fitness&  Move-Model Fitness&  Move-Log Fitness&  Num. States&  Approx. memory used (kb)& Trace Length\\ \hline 
		Model 29&  3.44&  1.00&  1.00&  1.00&  1.00&  54.80& 64.80\\ \hline 
		Model 30&  3330.23&  1.00&  1.00&  1.00&  200011.00&  11414.50& 64.80\\ \hline 
		Model 31&  7.48&  0.09&  0.91&  0.09&  432.30&  71.80& 64.20\\ \hline 
		Model 32&  3974.60&  1.00&  1.00&  1.00&  200007.40&  11856.50& 64.20\\ \hline 
		Model 33&  2.82&  1.00&  1.00&  1.00&  1.00&  51.90& 72.10\\ \hline 
		Model 34&  3566.72&  1.00&  1.00&  1.00&  200008.40&  12427.20& 72.10\\ \hline 
		Model 35&  14.55&  0.15&  0.98&  0.13&  844.75&  128.05& 67.85\\ \hline 
		Model 36&  5050.51&  1.00&  1.00&  1.00&  200006.85&  12389.15& 67.85\\ \hline 
		Model 37&  2.79&  1.00&  1.00&  1.00&  1.00&  74.20& 101.00\\ \hline
		Model 38& 3159.70& 1.00& 1.00& 1.00& 200006.00& 11796.30&101.00\\\hline
		Model 39& 27.44& 0.21& 0.99& 0.19& 1296.27& 214.67&78.87\\\hline
		Model 40& 6170.17& 1.00& 1.00& 1.00& 200005.33& 12930.40&78.87\\\hline
	\end{tabular}
	\caption{Global statistics: Model 29 to Model 40}
	\label{tab:my_label}
\end{table*}

\begin{figure}[!h]
	\includegraphics[width=1\linewidth]{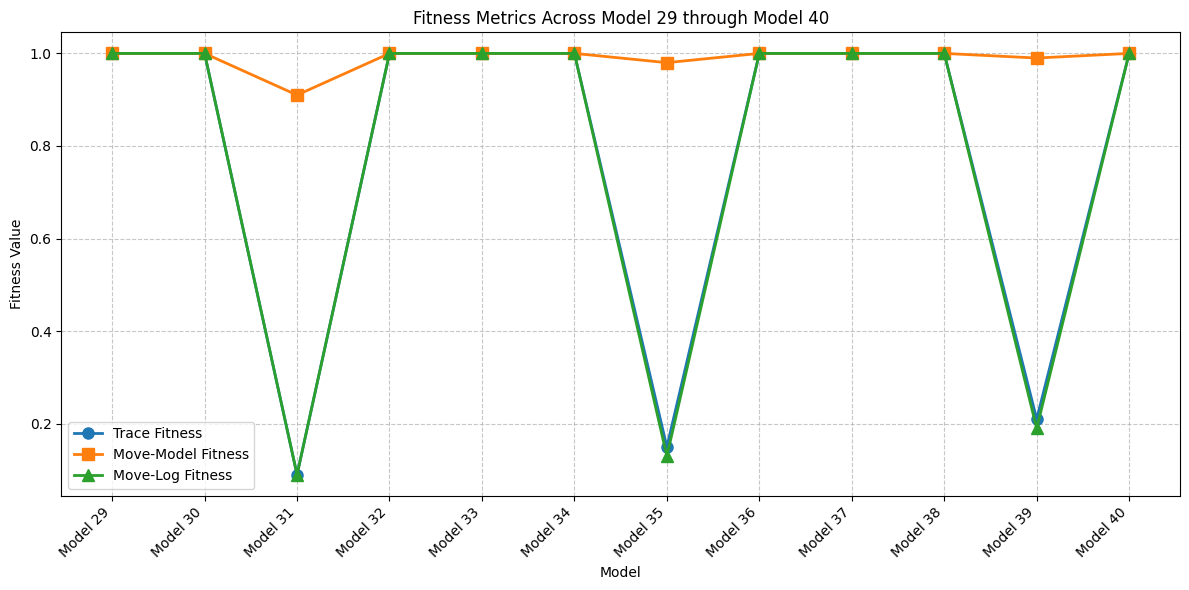}
	\caption{Visual plot shows fitness metrics across Model 29 through Model 40}
	\label{fig:Visual plot shows fitness scores across Tables 29 to 40}
\end{figure}

\textbf{Figure~\ref{fig:Visual plot shows fitness scores across Tables 29 to 40} shows compared fitness metrics (Trace Fitness, Move-Model Fitness and Move-Log Fitness) across Model 29 through Model 40}.


\textbf{\\ Perfect Model Fitness: \\}
Most datasets, including Models 29, 30, 32, 33, 34, 36, 37, 38, and 40, exhibit a high fitness value of 1.00 in the metrics Trace Fitness, Move-Model Fitness, and Move-Log Fitness, indicating consistently high fitness and that the model performs excellently for these datasets. The model is also capable of tracking the real data accurately, and the sequence is consistent with the moves of the model as well as log moves of these datasets.

\textbf{\\ Non-fitting Models: \\}
On the other hand, datasets containing low fitness degrees include Model 31, which shows a significant drop in Trace Fitness and Move-Log Fitness to 0.09. This points to the model's low accuracy in mapping real data and in matching the logs; hence, it shows a number of problems with how it captures the process in this dataset. Moreover, from Model 35, it is seen that Trace Fitness goes down to 0.15 while Move-Log Fitness comes down to 0.13, showing a critical misalignment between the model and actual data, as well as poor log tracing. Similar to previous outliers, Model 39 has a low Trace Fitness of 0.21 and Move-Log Fitness of 0.19, but a high Move-Model Fitness of 0.99, suggesting that the model's logic is reasonable, but the actual data diverges from the standard expectations. Drops in Trace Fitness and Move-Log Fitness for Models 31, 35, and 39 show consistent problems with the model.

\begin{table*}[!h]
	\centering
	\begin{tabular}{|l|c|c|c|c|}
		\hline
		Metric & R.A-A.D 1 & R.A-I.M 1 & W.A-A.D 1 & W.A-I.M 1 \\
		\hline
		Calc. Time (ms) & 3.44 & 3330.23 & 7.48 & 3974.60 \\
		Num. States & 1.00 & 200011.00 & 432.30 & 200007.40 \\
		Trace Fitness & 1.00 & 1.00 & 0.09 & 1.00 \\
		Raw Fitness Cost & 0.00 & 0.00 & 59.40 & 0.00 \\
		Move-Model Fitness & 1.00 & 1.00 & 0.91 & 1.00 \\
		Pre-process time (ms) & 0.67 & 0.53 & 0.25 & 0.31 \\
		Move-Log Fitness & 1.00 & 1.00 & 0.09 & 1.00 \\
		Trace Length & 64.80 & 64.80 & 64.20 & 64.20 \\
		Approx. mem. used (kb) & 54.80 & 11414.50 & 71.80 & 11856.50 \\
		\hline
	\end{tabular}
	
	\vspace{0.5cm}
	
	\begin{tabular}{|l|c|c|c|c|}
		\hline
		Metric & R.A-A.D 2 & R.A-I.M 2 & W.A-A.D 2 & W.A-I.M 2 \\
		\hline
		Calc. Time (ms) & 2.82 & 3566.72 & 14.55 & 5050.51 \\
		Num. States & 1.00 & 200008.40 & 844.75 & 200006.85 \\
		Trace Fitness & 1.00 & 1.00 & 0.15 & 1.00 \\
		Raw Fitness Cost & 0.00 & 0.00 & 59.30 & 0.00 \\
		Move-Model Fitness & 1.00 & 1.00 & 0.98 & 1.00 \\
		Pre-process time (ms) & 0.37 & 0.22 & 0.10 & 0.23 \\
		Move-Log Fitness & 1.00 & 1.00 & 0.13 & 1.00 \\
		Trace Length & 72.10 & 72.10 & 67.85 & 67.85 \\
		Approx. mem. used (kb) & 51.90 & 12427.20 & 128.05 & 12389.15 \\
		\hline
	\end{tabular}
	
	\vspace{0.5cm}
	
	\begin{tabular}{|l|c|c|c|c|}
		\hline
		Metric & R.A-A.D 3 & R.A-I.M 3 & W.A-A.D 3 & W.A-I.M 3 \\
		\hline
		Calc. Time (ms) & 2.79 & 3159.70 & 27.44 & 6170.17 \\
		Num. States & 1.00 & 200006.00 & 1296.27 & 200005.33 \\
		Trace Fitness & 1.00 & 1.00 & 0.21 & 1.00 \\
		Raw Fitness Cost & 0.00 & 0.00 & 63.33 & 0.00 \\
		Move-Model Fitness & 1.00 & 1.00 & 0.99 & 1.00 \\
		Pre-process time (ms) & 0.09 & 0.14 & 0.42 & 0.17 \\
		Move-Log Fitness & 1.00 & 1.00 & 0.19 & 1.00 \\
		Trace Length & 101.00 & 101.00 & 78.87 & 78.87 \\
		Approx. mem. used (kb) & 74.20 & 11796.30 & 214.67 & 12930.40 \\
		\hline
	\end{tabular}
	\caption{Global statistics: Petri-net generated by various algorithms (W.A white agent, R.A red agent, I.M inductive miner, A.D alpha discovery) (Fixed Simulation Depth, Fixed Iteration Times, Minimax Search Depth = n)}
	\label{tab:petri-net-stats}
\end{table*}

\section{Appendix D : Discovery Analysis} \label{Appendix D : Discovery Analysis}
\begin{figure}[!h]
	\centering
	\includegraphics[width=\linewidth]{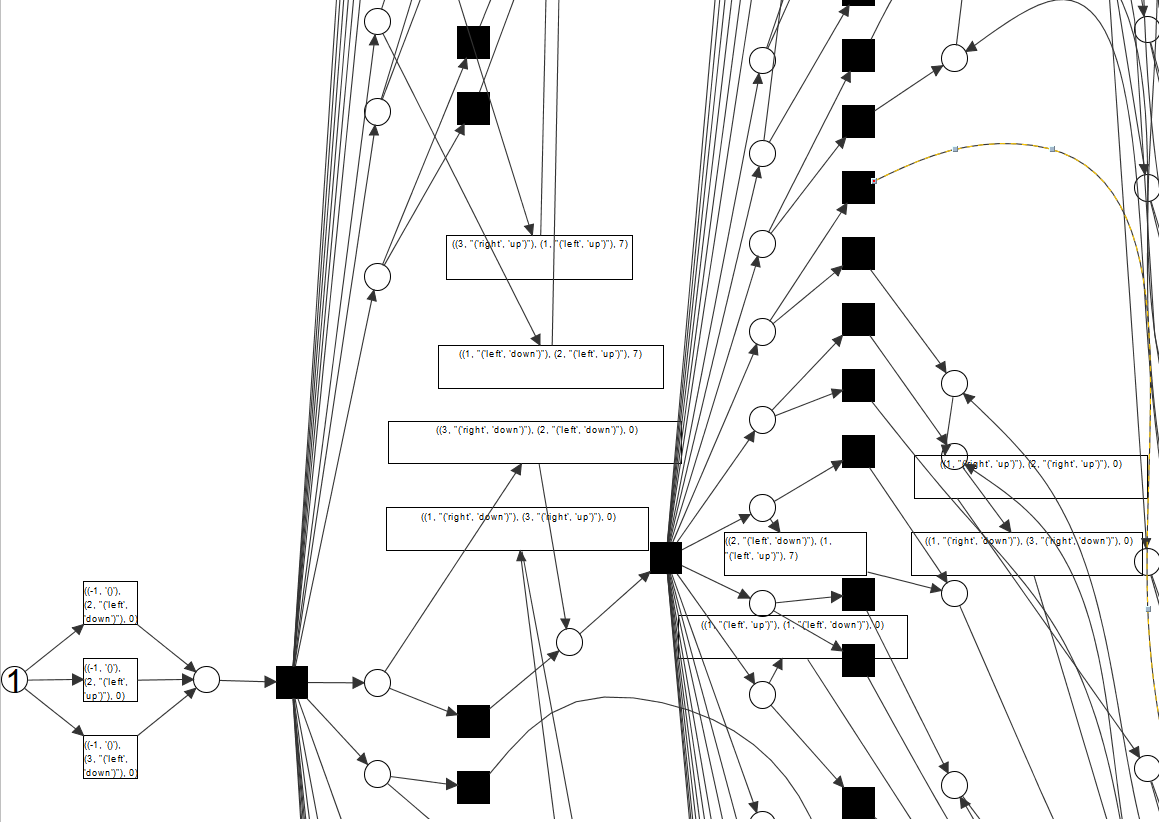}
	\caption{Simplified red agent Petri-net (10 episodes) generated by inductive miner algorithm (Fixed Simulation Depth, Fixed Minimax Search Depth, Iteration Times = 3000)}
	\label{fig: red_simplified}
\end{figure}

For red agent in Figure~\ref{fig: red_simplified} , we can see transitions in the first layer are

\begin{itemize}
	\item $((-1, ``0"), (2, (``left", ``down")), 0)$
	\item $((-1, ``0"), (2, (``left", ``up")), 0)$
	\item $((-1, ``0"), (3, (``left", ``down")), 0)$
\end{itemize}

and transitions in the second layer are 

\begin{itemize}
	\item $((3, (``right", ``up")),\allowbreak (1, (``left", ``up")), 7)$
	\item $((1, (``left", ``down")),\allowbreak (2, (``left", ``up")), 7)$
	\item $((3, (``right", ``down")),\allowbreak (2, (``left", ``down")), 0)$
	\item $((1, (``right", ``down")),\allowbreak (3, (``right", ``up")), 0)$
\end{itemize}

and transitions in the third layer are 

\begin{itemize}
	\item $((2, (``left", ``down")),\allowbreak (1, (``left", ``up")), 7)$
	\item $((1, (``left", ``up")),\allowbreak (1, (``left", ``down")), 0)$
\end{itemize}

We can treat these three transitions in the first layer as all possible actions computed by red agent in the starting turn. Since the enemy piece id in the last turn is -1, red agent is the first player. We can treat four transitions and other more transitions not shown in the Figure~\ref{fig: red_simplified} but in the second layer as all actions computed by red agent in the second turn. Thus, we can say, the red agent Petri-net generated by the inductive miner algorithm can provide players with all decisions considered by the MCTS-minimax hybrids agent in every turn of the 3v3 checkers game. We can see moving the white piece ID = 2 left and up or moving the white piece ID = 1 left and up can result in 7 reward points. Thus, we can say, when the white piece 3 moves right and up in the last turn, we recommend selecting piece 1 and moving it left and up;  when the white piece 1 moves left and down in the last turn, we recommend selecting piece 2 and moving it left and up (Why do you recommend this action?) because both actions will cause either an enemy piece to be captured or the current piece to be crowned. We do not recommend alternative actions $(2, ("left", "down"))$, $(3, ("right", "up"))$ (Why don't you recommend this alternative action?) because they won't earn any reward points for us. One of three actions in the first layer can be chosen by the red agent. If in the next turn, white piece 3 is selected to move right and up or white piece 1 is selected to move left and down, we recommend a human player holding red piece to either select piece ID = 2 and movement (left, up) or select piece ID = 1 and movement (left, up) as a provident future action because the action results in earning 7 reward points (What do you recommend in these possible futures?). If in the next turn, white piece 3 is selected to move right and down, even though selecting piece 2 and moving it left and down results in 0 reward points, we still recommend human player to choose this action. Referring to the third layer's transition $((2, (``left", ``down")), (1, (``left", ``up")), 7)$ that results in 7 reward points, selecting red piece 2 and moving it left and down can result in this transition, which brings future reward points to red.

\begin{figure}[!h]
	\centering
	\includegraphics[width=\linewidth]{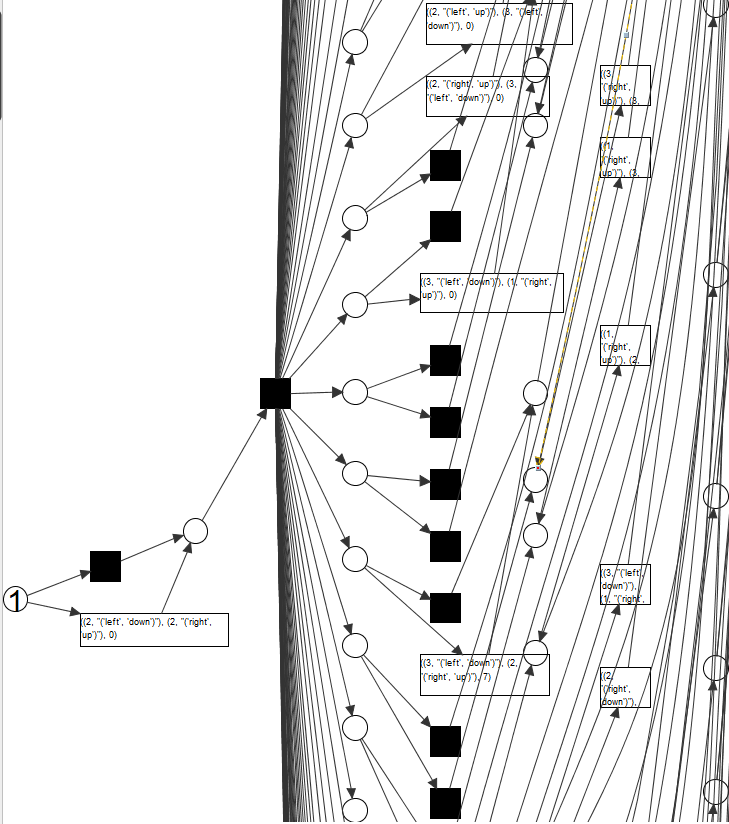}
	\caption{Simplified white agent Petri-net (10 episodes) generated by inductive miner algorithm (Fixed Simulation Depth, Fixed Minimax Search Depth, Iteration Times = 3000)}
	\label{fig: white_simplified}
\end{figure}

For white agent in Figure~\ref{fig: white_simplified}, we can see the transition in the first layer is

\begin{itemize}
	\item $((2, (``left", ``down")),\allowbreak (2, (``right", ``up")), 0)$
\end{itemize}

and transitions in the second layer are 

\begin{itemize}
	\item $((2, (``left", ``up")),\allowbreak (3, (``right", ``down")), 0)$
	\item $((2, (``right", ``up")),\allowbreak (3, (``left", ``down")), 0)$
	\item $((3, (``left", ``down")),\allowbreak (1, (``right", ``up")), 0)$
	\item $((3, (``left", ``down")),\allowbreak (2, (``right", ``up")), 7)$
\end{itemize}

We can treat these four transitions and other more transitions not shown in the Figure~\ref{fig: white_simplified} but in the second layer as all actions computed by white agent in the second turn. Thus, we can say, the white agent Petri-net generated by the inductive miner algorithm can provide players with all decisions considered by the MCTS-minimax hybrids agent in every turn of the 3v3 checkers game. We can see moving the white piece ID = 2 right and up can result in 7 reward points. Thus, we can say, when the red piece 3 moves left and down in the last turn, we recommend selecting piece 2 and moving it right and up (Why do you recommend this action?) because it will cause either an enemy piece to be captured or the current piece to be crowned. We do not recommend alternative actions $(3, (``right", ``down"))$, 
$(3, (``left", ``down"))$, $(1, (``right", ``up")), 0)$ (Why don't you recommend this alternative action?) because they won't earn any reward points for us. Considering the transition $((2, (``left", ``down")), 
(2, (``right", ``up")), 0)$ in the first layer, it is not hard to see that in the last turn, red piece 2 is selected to move left and down, and in the current turn, white piece 2 is selected to move right and up. If in the next turn, red piece 3 is selected to move left and down, we recommend a human player holding white piece to select piece ID = 2 and movement (left, up) as a provident future action because this action results in earning 7 reward points (What do you recommend in these possible futures?). If in the next turn, red piece 2 is selected to move left and up or to move right and up, we recommend a human player holding white piece to either select piece ID = 3, movement (right, down) or piece ID = 3, movement (left, down) as possible future actions. To ensure which action is the optimal, players need to refer to all transitions in the third layer, fourth layer, or further future layers to check which of these two actions can result in early reward points in future turns.

\section {Appendix E: Limitation Extended}

\begin{itemize}
	\item Lack of means to test win percentage: In this study, we use process models to provide both causal-relationship based explanations on question 1 (Why do you recommend this action?) and question 3 (Why don't you recommend this alternative action?) through observing different rewards earned by each action. We recommend a specific action due to resulting in the highest reward compared to other actions. However, such interpretation is only based on the local perspective because we are unable to gain a measurable insight on the game's win-rate through selecting a specific action. The causal-relationship based explanations need to be globally interpreted based on how much percentage the win-rate will increase for each action selected by the agent player. 
	\item Deficiency of minimax search algorithm's explainability: The MCTS and minimax are treated as a hybrid algorithm in our research. However, we only study the case when the simulation depth of MCTS is fixed with the number of iterations, how variable minimax search depth affects the quality of process models. We fail to explore the explainablility of minimax search algorithm embeded in the simulation stage of MCTS. We believe research in the decision-making strategy of minimax search algorithm needs to be conducted as well to better under how the minimax selects the most optimal action in each simulaton round of MCTS. 
	\item Adjustments on feature engineering work: To evaluate a process model's quality, we perform conformance analysis on each process model to interpret model's replay fitness. In this study, we use the Petri-net model generated by inductive miner algorithm to prove our hypothesis. We successfully prove that a process model can provide players with all possible decisions considered by the MCTS-minimax hybrids agent, as well as both causal-relationship based explanations on question 1 (Why does the MCTS-minimax hybrids agent select this movement?), question 3 (Why not MCTS-minimax hybrids select other movements?) and distal explanations on question 2 (According to the process model, what will be MCTS-minimax hybrids agent's next movement in the future?). However, we fail to use Petri-net and C-net generated by alpha process discovery algorithm and iDHM to answer our research question due to the fact that both models are too complex to be interpreted by humans. We believe more optimized feature engineering work is required to enhance the transformation on the decision-making data of MCTS-minimax hybrids algorithm model. Specifically, the content of transition data tuple in the event log needs to be transformed using advanced feature engineering technique, such as deep neural network. 
	\item Lack of generalization on sequential decision-making domains other than checkers: Our current discovery is only based on 3v3 checkers domain. We attempt to provide explanations on three fundamental types of post-hoc questions (1. Why do you recommend this action? 2. What do you recommend in these possible futures? 3. Why don't you recommend this alternative action?) by observing behaviors of MCTS-minimax hybrids agents reflected in the process model. However, current discovery is unable to be generalized to other sequential decision-making domains with different sizes (state features/number of actions), such as Go or Connect Four. 
\end{itemize}

\section{Appendix F : Future Work on Feature Engineering and Human Study} \label{Appendix E: Future Work on Feature Engineering and Human Study}
In terms of feature engineering work, in addition to using deep neural networks for feature extraction, an alternative plan is to use the change of shortest distance between all red pieces and white pieces in each piece's movement to replace the abstract movement representation. Specifically, we can treat all red pieces as an entity $Red$ and all white pieces as another entity $White$, where we store all red pieces' coordinates on the game board in an array $Red$ $Array$ and store all white pieces' coordinates in another array $White$ $Array$. A modified breadth-first search algorithm is applied here twice to search the shortest distance between red pieces in the $Red$ $Array$ and white pieces in the $White$ $Array$ before piece's movement and after pieces' movement. The change of distance $\Delta$d is treated as way to represent a piece's movement on the game board. If $\Delta$d is negative, the shortest distance between red pieces and white pieces shrink, which may imply that the selected piece is approaching enemy pieces on purpose. If $\Delta$d is positive, then the shortest distance between red pieces and white pieces increases, which may imply either an enemy piece has been captured in the movement or the selected piece moves away from enemy pieces. One advantage of this plan is that compared to the dimension size of abstract representation, it has smaller dimension and may increase the simplicity of the process model through possibly minimizing the amount of events (transition tuple data) in an even log. However, using breadth-first search algorithm may reduce the overall performance of the MCTS algorithm model since it requires up to $O$(n) ($n$ stands for the size of game board) complexity to find the shortest distance between all red pieces in $Red$ $Array$ and all white pieces in $White$ $Array$. In addition, treating the change of distance $\Delta$d as a piece's movement on the game board may miss the detection of a piece's moving direction that is originally represented in abstract representation. Listing~\ref{lst: BFS} shows how the breadth-first search algorithm is implemented to find the shortest distance between all red pieces in $Red$ $Array$ and all white pieces in $White$ $Array$. All white pieces' coordinates are push into a queue. Then, a while loop is utilized here to check whether the queue is empty. Inside the while loop, we retrieve the size of the current queue due to the fact that we only want to search all neighbor positions of every white piece in current queue. A queue follows the rule of first-in-first-out, and we pop out the position of the white piece which is firstly added to the queue. If there is a red piece on the current position, we successfully find the shortest distance between all white pieces and red pieces. Otherwise, we iterate over every possible neighbor of current position. We check whether the neighbor position is a valid position and whether the neighbor position has been visited before. If the neighbor is a valid position and it has never been visited, we push the neighbor's position in the end of the queue and record the neighbor's position in our visited set. 

Human studies are required to test whether these models are interpretable and useful. By getting them to follow a model, we can test whether they understand it. Specifically, we want human players to hold red and MCTS-minimax hybrids agent to hold white. Human players will follow decision-strategies shown in the red players's process model to play against MCTS-minimax hybrids agent. The result of each episode will be recorded and the win percentage of both human players and MCTS-minimax hybrids agent in 100 episodes will be separately computed. To provide interpretability on how the minimax search algorithm selects the most optimal action in each simulaton round of MCTS, process mining technqiue should be conducted on minimax search algorithm. For instance, if in each turn of one episode of MCTS, 20 simulations are conducted for every iteration, then for every simulation, an event log regarding the decision-making strategy of minimax search algorithm needs to be established to generate process on the purpose of helping researchers understand how the minimax chooses the most optimal action for MCTS. 

\section{Appendix G: MCTS Pruning Operation} \label{Appendix F: MCTS Pruning Operation}

\begin{figure}[!h]
	\centering
	\includegraphics[width=1\linewidth]{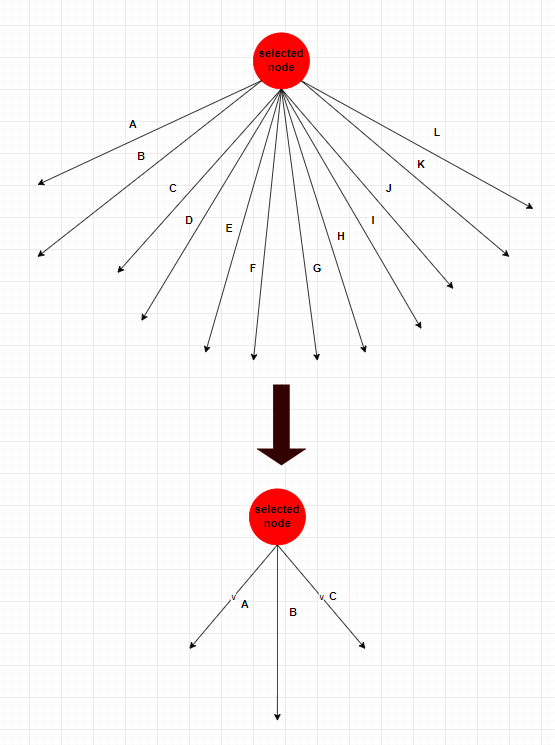}
	\caption{Pruning operation}
	\label{Pruning operation}
\end{figure}

In MCTS, a game state is stored in each node and an action is treated as a branching factor. This implies that a large number of branching factors need to be considered in MCTS' expansion stage. Pruning operations need to be applied to cut off unnecessary branching factors of each node in MCTS. Otherwise, given limited amount of time, a finite number of iterations may not be able to fully expand a selected node. If a node has not been fully expanded, before time runs out, UCT algorithm is unable to be applied to select the most optimal branch factor for this node. Therefore, a solution must be proposed to prune branching factors. We believe using a hashtable, a commonly used linear data structure, can provide an appropriate solution. For instance, in one iteration of MCTS, the algorithm completes selecting a node that has not been fully expanded. Assuming 12 actions ($A$, $B$, $C$, $D$, $E$, $F$, $G$, $H$, $I$, $J$, $K$, $L$) are retrieved as selected node's branching factors as this stage, we use an additional reward mechanism to score every action. This reward mechanism renders that actions $A$, $B$, $C$ are reward 10 points, actions $D$, $E$ are rewarded 6 points, action $F$, $G$, $H$ are rewarded 4 points, and actions $I$, $J$, $K$, $L$ are rewarded 0 points. We treat reward point as a key and a list of actions as a value in the hashtable. Figure~\ref{Pruning operation} and Table~\ref{Reward Hashtable} show our proposed methodology of this pruning operation. We prune branching factors through returning a list of actions with the highest reward points in the hashtable. One advantage of hashtable is that it supports fast insertion, fast search, and fast deletion. Since both time cost and space cost of insertion, search, and deletion is constant $O(1)$, using hashtable to prune branching factors does not affect the overall performance of MCTS algorithm model. 

\section{Appendix H: Process Models Figures} \label{Appendix G: Process Models Figures}

\begin{figure}[!h]
	\centering
	\includegraphics[width=\linewidth]{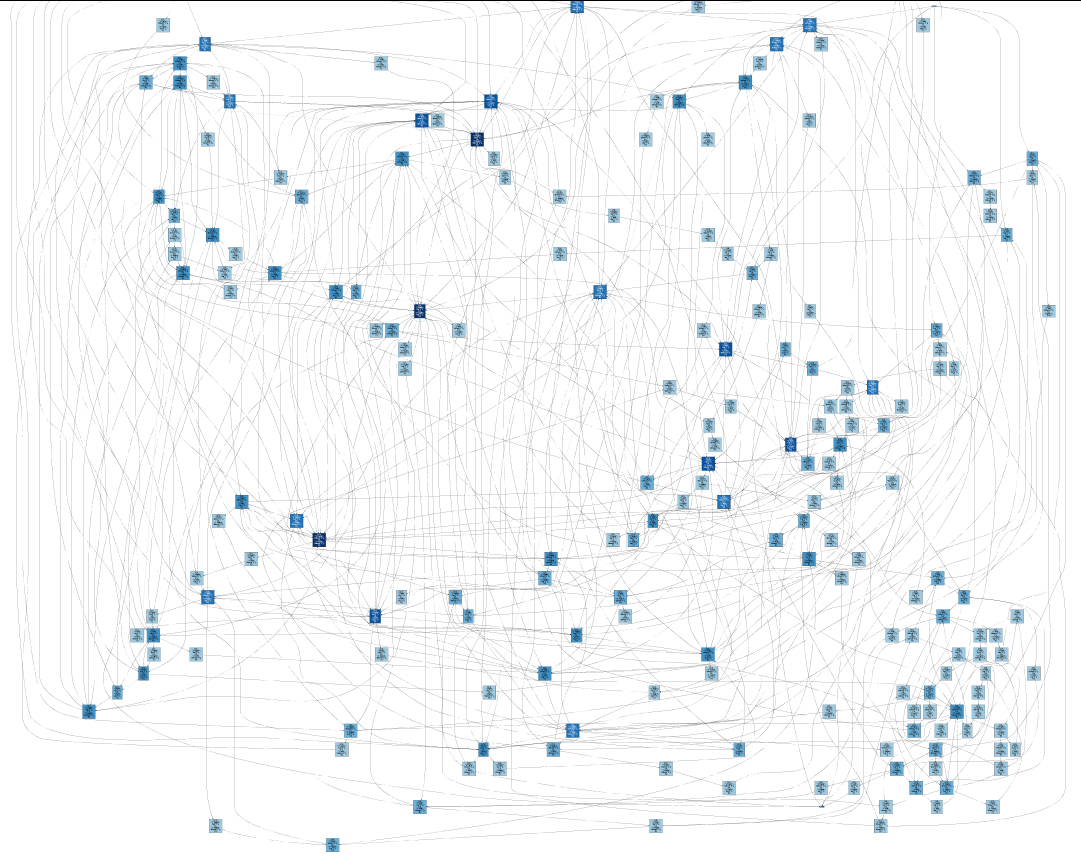}
	\caption{White agent C-net generated by iDHM (Fixed Simulation Depth, Fixed Minimax Search Depth, Iteration Times = 2000)}
	\label{fig: white_iteration_2000_iDHM}
\end{figure}

\begin{figure}[!h]
	\centering
	\includegraphics[width=\linewidth]{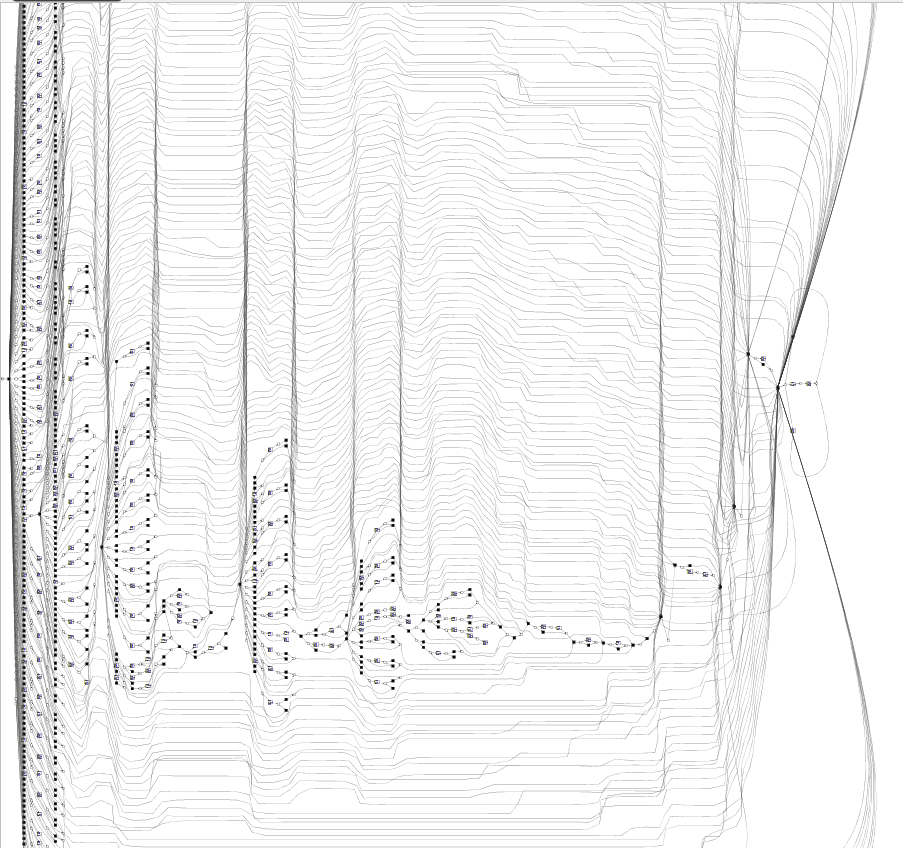}
	\caption{White agent Petri-net generated by inductive miner algorithm (Fixed Simulation Depth, Fixed Minimax Search Depth, Iteration Times = 2000)}
	\label{fig: white_iteration_2000_inductive}
\end{figure}

\clearpage

\section{Appendix I: Other Figures} \label{Appendix H: Other Figures}

\begin{figure}[!h]
	\centering
	\includegraphics[width=0.9\linewidth]{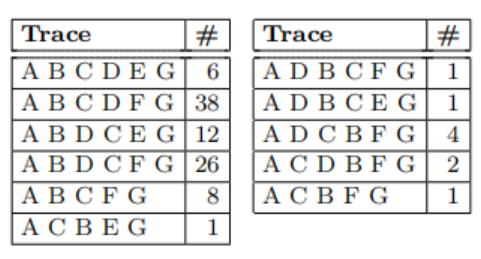}
	\caption{Sample event log II~\cite{buijs2014quality}}
	\label{fig:sample event log II}
\end{figure}

\begin{figure}[!h]
	\centering
	\includegraphics[width=0.75\linewidth]{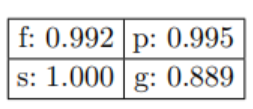}
	\caption{Sample event log II's quality dimensions~\cite{buijs2014quality}}
	\label{fig:sample event log II's Quality Dimensions}
\end{figure}

\begin{figure}[!h]
	\centering
	\includegraphics[width=1\linewidth]{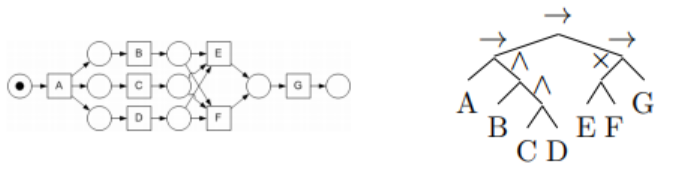}
	\caption{Petri-net and Process Tree of sample event log II~\cite{buijs2014quality}}
	\label{fig:Petri-net & Process Tree of sample event log II}
\end{figure}

\begin{figure}[!h]
	\centering
	\includegraphics[width=0.7\linewidth]{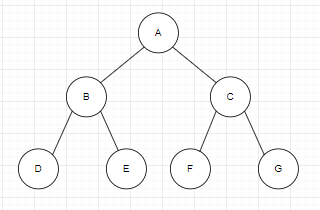}
	\caption{Binary tree}
	\label{fig: Binary Tree}
\end{figure}

\begin{figure}[!h]
	\centering
	\includegraphics[width=0.8\linewidth]{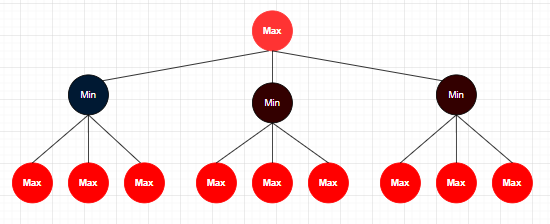}
	\caption{Game tree}
	\label{fig: Game Tree}
\end{figure}

\begin{figure}[!h]
	\centering
	\includegraphics[width=0.8\linewidth]{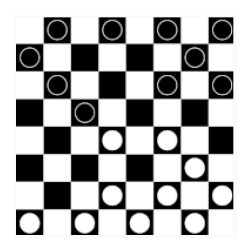}
	\caption{A typical mid-game checkers position~\cite{samuel1959some}}
	\label{fig:Checkers}
\end{figure}

\clearpage

\section{Appendix J: Python Code} \label{Appendix I: Python Code}

\begin{listing*}[!h]
	\caption{Preorder traversal in a binary tree}
	\label{lst: Preorder}
	\begin{lstlisting}[language=Python]
		class TreeNode(object):
		      def __init__(self, val=0, left=None, right=None):
		      self.val = val
		      self.left = left
		      self.right = right
		
		class Preorder(object):
    		def __init__(self):
        		self.ans = []
        		
        		def iteration(self, root):
        		
        		stack = [root]
        		
        		while stack:
        		
            		node = stack.pop()
                		if node:
                    		self.ans.append(node.val)
                    		stack.append(node.right)
                    		stack.append(node.left)
    		
    		def recursion(self, root):
    		
        		if not root:
        		      return 
    		
        		self.ans.append(root.val)
        		self.recursion(root.left)
        		self.recursion(root.right)
	\end{lstlisting}
\end{listing*}

\begin{listing*}[htbp]
	\caption{Inorder traversal in a binary tree}
	\label{lst: Inorder}
	\begin{lstlisting}[language=Python]
        class TreeNode(object):
            def __init__(self, val=0, left=None, right=None):
                self.val = val
                self.left = left
                self.right = right
        
        class Inorder(object):
            def __init__(self):
                self.ans = []
        
            def iteration(self, root):
                stack = []
                node = root
        
                while stack or node:
                    while node:
                        stack.append(node)
                        node = node.left
                    node = stack.pop()
                    self.ans.append(node.val)
                    node = node.right
        
            def recursion(self, root):
                if not root:
                    return 
                
                self.recursion(root.left)
                self.ans.append(root.val)
                self.recursion(root.right)
	\end{lstlisting}
\end{listing*}

\begin{listing*}[tb]
	\caption{Postorder traversal in a binary tree}
	\label{lst: Postorder}
	\begin{lstlisting}[language=Python]
        class TreeNode(object):
            def __init__(self, val=0, left=None, right=None):
                self.val = val
                self.left = left
                self.right = right
        
        class Postorder(object):
            def __init__(self):
                self.ans = []
            
            def iteration(self, root):
        
                stack = [root]
        
                node = root
        
                while stack:
                    node = stack.pop()
                    self.ans.append(node.val)
                    stack.append(node.left)
                    stack.append(node.right)
        
                self.ans = self.ans[::-1]
        
            def recursion(self, root):
        
                if not root:
                    return 
                
                self.recursion(root.left)
                self.recursion(root.right)
                self.ans.append(root.val)
	\end{lstlisting}
\end{listing*}

\begin{listing*}[tb]
	\caption{Tree node}
	\label{lst: TreeNode}
	\begin{lstlisting}[language=Python]
        class TreeNode(object):
            def __init__(self, board, turn, terminate, parent):
                self.board = board
                self.turn = turn
                self.terminate = terminate
                self.parent = parent
                self.children = {}
                self.visits = 0
                self.reward = [0, 0]
                self.isFullyExpanded = False
	\end{lstlisting}
\end{listing*}

\begin{listing*}[tb]
	\caption{MCTS minimax agent}
	\label{lst: MCTS_Minimax_agent}
	\begin{lstlisting}[language=Python]
        class MCTS_Minimax_agent(object):
            def __init__(self, board, agent_color, iterations, simulation_depth, minimax_depth):
                self.board = board
                self.agent_color = agent_color
                self.iterations = iterations
                self.simulation_depth = simulation_depth
                self.minimax_depth = minimax_depth
                self.exploration_constant = 1/sqrt(2)
                self.discounted_factor = 0.8
	\end{lstlisting}
\end{listing*}

\begin{listing*}[tb]
	\caption{Minimax}
	\label{lst: Minimax}
	\begin{lstlisting}[language=Python]
        def Minimax(self, node, depth, max_player):
            
            board = node.board
        
            if depth == 0 or node.board.winner() != None:
                return board.evaluate(), board
            
            if max_player:
                maxEval = float('-inf')
                best_move = None
                for move in self.get_moves(node):
                    board, reward, next_turn, terminate, movement_info = move
                    node = TreeNode(board, next_turn, terminate, node)
                    evaluation = self.Minimax(node, depth - 1, False)[0]
                    maxEval = max(maxEval, evaluation)
                    if maxEval == evaluation:
                        best_move = move
                
                return maxEval, best_move
            
            else:
                minEval = float('inf')
                best_move = None
                for move in self.get_moves(node):
                    board, reward, next_turn, terminate, movement_info = move
                    node = TreeNode(board, next_turn, terminate, node)
                    evaluation = self.Minimax(node, depth - 1, True)[0]
                    minEval = min(minEval, evaluation)
                    if minEval == evaluation:
                        best_move = move
                
                return minEval, best_move
	\end{lstlisting}
\end{listing*}

\begin{listing*}[tb]
	\caption{Simulation}
	\label{lst: Simulation}
	\begin{lstlisting}[language=Python]
        def simulation(self, node):
                
            reward = [0, 0]
            depth = 0
        
            while not node.terminate:
                maxEval, best_move = self.Minimax(node, self.minimax_depth, True)
                if maxEval == float('-inf'):
                    break
                    
                board, r, next_turn, terminate, movement_info = best_move
        
                if node.turn == WHITE:
                    reward[0] += r
                else:
                    reward[1] += r
        
                node = TreeNode(board, next_turn, terminate, node)
        
                depth += 1
                if depth == 20:
                    break
        
            return reward
	\end{lstlisting}
\end{listing*}

\begin{listing}[!h]
	\caption{Get moves}
	\label{lst: Get moves}
	\begin{lstlisting}[language=Python, linewidth=\textwidth]
        def get_moves(self, node):
            
            moves = []
        
            # iterate over every single piece in the game board
            for piece in node.board.get_all_pieces(node.turn):
                # get all possible moves based on current piece
                valid_moves = node.board.get_valid_moves(piece)
        
                for move_, skip in valid_moves.items():
        
                    temp_board = deepcopy(node.board)
                    temp_piece = temp_board.get_piece(piece.row, piece.col)
                    reward = temp_board.move(temp_piece, move_[0], move_[1])
        
                    if skip:
                        temp_board.remove(skip)
                        reward += len(skip) * 5
        
                    removed_piece_id = [p.id for p in skip]
        
                    cur_x, cur_y = piece.row, piece.col
                    target_x, target_y = move_[0], move_[1]
        
                    dx = target_x - cur_x
                    dy = target_y - cur_y
        
                    move = None
        
                    if dx > 0 and dy == 0:
                        move = ("right")
                    elif dx < 0 and dy == 0:
                        move = ("left")
                    elif dx == 0 and dy > 0:
                        move = ("up")
                    elif dx == 0 and dy < 0:
                        move = ("down")
                    elif dx > 0 and dy > 0:
                        move = ("right", "up")
                    elif dx < 0 and dy > 0:
                        move = ("left", "up")
                    elif dx < 0 and dy < 0:
                        move = ("left", "down")
                    elif dx > 0 and dy < 0:
                        move = ("right", "down")
        
                    movement_info = (temp_piece.id, move, removed_piece_id)
        
                    if node.turn == RED:
                        next_turn = WHITE
                    else:
                        next_turn = RED
        
                    if temp_board.winner():
                        terminate = True
                    else:
                        terminate = False 
        
                    moves.append((temp_board, reward, next_turn, terminate, movement_info))
        
            return moves
	\end{lstlisting}
\end{listing}

\begin{listing*}[!h]
	\caption{Main part 1}
	\label{lst: Main part 1}
	\begin{lstlisting}[language=Python, linewidth=\textwidth]
        def main(simulation_depth, minimax_depth, iterations, file_path):
        
            for i in range(1, 101):
                white_traces = { 'last_turn_enemy_piece_id': [], 'last_turn_enemy_movement': [], 'piece id': [], 'move': [], "skip": [], 'reward': [] }
                red_traces = { 'last_turn_enemy_piece_id': [], 'last_turn_enemy_movement': [], 'piece id': [], 'move': [], "skip": [], 'reward': [] }
        
                run = True
        
                game = Game(None)
                winner = None
        
                last_turn_enemy_piece_id = -1
                last_turn_enemy_movement = tuple()
        
                while run:
                    if game.turn == WHITE:
                        white_agent = MCTS_agent(game.get_board(), WHITE, iterations, simulation_depth, minimax_depth)
                        action = white_agent.get_action()
        
                        if action == None:
                            winner = RED
                            run = False
                            continue
        
                        new_board, reward, next_turn, terminate, movement_info = action
        
                        white_traces['last_turn_enemy_piece_id'].append(last_turn_enemy_piece_id)
                        white_traces['last_turn_enemy_movement'].append(last_turn_enemy_movement)
                        white_traces['piece id'].append(movement_info[0])
                        white_traces['move'].append(movement_info[1])
                        white_traces['skip'].append(movement_info[2])
                        white_traces['reward'].append(reward)
        
                        last_turn_enemy_piece_id = movement_info[0]
                        last_turn_enemy_movement = movement_info[1]
        
                    else:
                        red_agent = MCTS_agent(game.get_board(), RED, iterations, simulation_depth, minimax_depth)
                        action = red_agent.get_action()
        
                        if action == None:
                            winner = WHITE
                            run = False
                            continue
        
                        new_board, reward, next_turn, terminate, movement_info = action
        
                        red_traces['last_turn_enemy_piece_id'].append(last_turn_enemy_piece_id)
                        red_traces['last_turn_enemy_movement'].append(last_turn_enemy_movement)
                        red_traces['piece id'].append(movement_info[0])
                        red_traces['move'].append(movement_info[1])
                        red_traces['skip'].append(movement_info[2])
                        red_traces['reward'].append(reward)
        
                        last_turn_enemy_piece_id = movement_info[0]
                        last_turn_enemy_movement = movement_info[1]
        
                    game.ai_move(new_board)
                    winner = game.winner()
        
                    if winner:
                        run = False
	\end{lstlisting}
\end{listing*}

\begin{listing*}[!h]
	\caption{Main part 2}
	\label{lst: Main part 2}
	\begin{lstlisting}[language=Python, linewidth=\textwidth]
 
                print(f"episode{i} completed")
                print(f"{winner} won")
        
                white_path = file_path + f"white_episode{i}.csv"
                red_path = file_path + f"red_episode{i}.csv"
        
                df = pd.DataFrame(white_traces)
                df.to_csv(white_path, index = False)
        
                df = pd.DataFrame(red_traces)
                df.to_csv(red_path, index = False)
	\end{lstlisting}
\end{listing*}

\begin{listing*}[!h]
	\caption{Breadth-first search in checkers}
	\label{lst: BFS}
    \small
	\begin{lstlisting}[language=Python]
    from collections import deque
    
    def BFS(board, white_pieces, red_pices):
    
        row = len(board)
        col = len(board[0])
    
        def check(x, y):
            return 0 <= x < row and 0 <= y < col
    
        # BFS expansion direction
        directions = [[0, 1], [0, -1], [1, 0], [-1, 0]]
    
        # record visited coordinates
        visited = set()
    
        # record the shortest distance
        step = 0
    
        # initilize a queue
        queue = deque()
    
        # push all white pieces' positions into the queue
        for position in white_pieces:
            queue.append(position)
    
        while queue:
            length = len(queue)
            # search four directions for current pieces in the queue
            for _ in range(length):
                # get current piece's position
                x, y = queue.popleft()
                # the nearest red piece found! 
                if (x, y) in red_pices:
                    return step
                # iterate over all possible neighbor positions
                for dx, dy in directions:
                    new_x, new_y = x + dx, y + dy
                    # check whether the new position is valid 
                    if check(new_x, new_y) and (new_x, new_y) not in visited:
                        # push the new position into the queue
                        queue.append((new_x, new_y))
                        # new position will be visited
                        visited.add((new_x, new_y))
            # update step
            step += 1
    
        # cannot find the nearest red piece
        return float('inf')
	\end{lstlisting}
\end{listing*}

\end{document}